\documentclass[english]{article}
\usepackage{geometry}
\usepackage[T1]{fontenc}
\usepackage[latin9]{inputenc}
\usepackage[unicode=true,
 bookmarks=false,
 breaklinks=false,pdfborder={0 0 1},colorlinks=false]
 {hyperref}
\hypersetup{
 colorlinks,citecolor=blue,filecolor=blue,linkcolor=blue,urlcolor=blue}

\makeatletter
\usepackage{bm}
\usepackage{amsmath,mathtools}
\usepackage{amssymb}
\usepackage{amsthm}  
\usepackage{comment}
\usepackage{natbib}
\usepackage{booktabs}
\usepackage{graphicx}
\usepackage[linesnumbered,ruled,vlined]{algorithm2e}
\usepackage{algorithmic}
\usepackage{float}
\usepackage{multirow}
\usepackage{dsfont}
\usepackage{tcolorbox}
\usepackage{color}

\allowdisplaybreaks

\definecolor{yxc}{RGB}{255,0,0}
\definecolor{yjc}{RGB}{125,0,0}
\definecolor{ytw}{RGB}{255,69,0}
\definecolor{gen}{RGB}{0,0,200}
\definecolor{cc}{RGB}{231,117,0}
\definecolor{ln}{RGB}{0,128,0}

\DeclareMathOperator{\ind}{\mathds{1}}
\DeclareMathOperator{\argmax}{\mathrm{argmax}}
\DeclareMathOperator{\argmin}{\mathrm{argmin}}

\newcommand{\defn}{\coloneqq}

\newcommand{\cN}{\mathcal{N}}

\newcommand{\bE}{\mathbb{E}}
\newcommand{\bP}{\mathbb{P}}
\newcommand{\bR}{\mathbb{R}}

\newcommand{\KL}{\mathsf{KL}}

\newcommand{\diff}{\mathrm{d}}

\newcommand{\unif}{\mathsf{Unif}}
\newcommand{\veps}{\varepsilon}
\newcommand{\wh}{\widehat}
\newcommand{\wt}{\widetilde}

\newcommand{\GenPPL}{\mathrm{PPL}_{\mathrm{gen}}}

\newcommand{\ug}{\mathsf{ug}}
\newcommand{\reference}{\mathsf{ref}}
\newcommand{\data}{\mathsf{data}}
\newcommand{\uni}{\mathsf{uni}}
\newcommand{\scg}{\mathsf{scg}}
\newcommand{\iscr}{\mathsf{iscr}}

\newcommand{\mymid}{\,|\,}
\usepackage{graphicx}
\usepackage{subcaption}
\usepackage{geometry}

\usepackage{booktabs}
\usepackage{multirow}
\usepackage{subcaption}
\usepackage{tcolorbox}
\tcbset{
    samplebox/.style={
        colback=gray!4,
        colframe=gray!65,
        colbacktitle=gray!65,
        coltitle=white,
        fonttitle=\bfseries\large,
        arc=4mm,
        sharp corners=south,
        boxrule=0.8pt,
        left=4mm,
        right=4mm,
        top=3mm,
        bottom=3mm,
        toptitle=1.5mm,
        bottomtitle=1.5mm
    }
}
\tcbuselibrary{breakable}

\title{ConvergeFlow: Language Flow with Provable Convergence to Token Embeddings}
\date{\today}

\makeatother

\theoremstyle{plain} 
 
\newtheorem{proposition}{\bf Proposition}
\newtheorem{theorem}{\bf Theorem}

\theoremstyle{remark}

\author{
Na Li\footnote{The authors contributed equally. Corresponding author: Gen Li.}~\thanks{Department of Statistics and Data Science, Chinese University of Hong Kong, Hong Kong; Email: \href{mailto:nali001@cuhk.edu.hk}{\{na.li}, \href{mailto:yuchenjiao@cuhk.edu.hk}{yuchenjiao}, \href{mailto:genli@cuhk.edu.hk}{genli\}@cuhk.edu.hk}
}
\and 
Yuchen Jiao\footnotemark[1]~\footnotemark[2] 
\and 
Changxiao Cai\thanks{Department of Industrial and Operations Engineering, University of Michigan, Ann Arbor, USA; 
Email: \href{mailto:cxcai@umich.edu}{cxcai@umich.edu}.
}
\and
Gen Li\footnotemark[2]
}

\begin{document}

\maketitle

\begin{abstract}
Recent advances in continuous diffusion and flow-based language models (LMs) have achieved performance competitive with discrete LMs. However, existing continuous frameworks still rely on decoders supervised with cross entropy (CE) because the flow trajectories are not guaranteed to terminate at valid token embeddings. Motivated by this limitation, we introduce \textbf{ConvergeFlow}, an embedding-space flow-based LM, which constrains the data predictor to the convex hull of token embeddings and trains it solely with the mean squared error objective induced by flow matching. Under suitable regularity conditions, we prove that the resulting flow converges to valid token embeddings despite errors in the data predictor, enabling direct token prediction without a CE-supervised decoder. We further develop three sampling mechanisms for controlling the trade-off between the generative perplexity and entropy. Experiments on OpenWebText demonstrate that ConvergeFlow achieves performance competitive with existing continuous and discrete diffusion LMs. These findings demonstrate the potential of the flow-based paradigm for language modeling. Our code is available at \url{https://github.com/Na-Li66/ConvergeFlow}.
\end{abstract}

\tableofcontents

\section{Introduction}
\label{sec:intro}

Diffusion models \citep{sohl2015deep,song2019generative,ho2020denoising} and flow matching (FM) \citep{lipman2022flow,liu2022flow} have become the backbone for generative modeling in continuous data domains, with applications spanning image synthesis \citep{dhariwal2021diffusion,rombach2022high,esser2024scaling}, video generation \citep{ho2022video,wan2025wan}, and protein design \citep{trippe2022diffusion}.
At a high level, these models learn a transport from a simple noise distribution to the data distribution. Diffusion models construct this transformation by learning to reverse a progressive noise corruption process, while FM learns the velocity field of a prescribed probability path. Once learned, the resulting generative dynamics iteratively transform fresh noise into new samples from the data distribution.

Language modeling, a central task in modern generative modeling, has long been dominated by autoregressive (AR) language models (LMs) \citep{radford2019language,brown2020language}.
% , which factorize a sequence distribution from left to right and generate one token at a time. 
%
Despite their remarkable success in practice, AR models have two inherent drawbacks. First, the left-to-right generation order prevents earlier tokens from being revised using later context, limiting bidirectional reasoning and controllable generation. Second, one-by-one sequential generation inherently restricts parallelism and creates a fundamental bottleneck in sampling speed.

To overcome these limitations, substantial effort has recently been devoted to diffusion and flow-based language modeling \citep{li2022diffusion,sahoo2024simple}.
These models offer a fundamentally different generation principle---they iteratively refine all token positions using bidirectional context. 
This formulation permits parallel token updates and allows global planning, controllable generation, and iterative revision, offering the potential for faster and more flexible generation.

Existing diffusion language models (DLMs) can be broadly categorized into continuous and discrete approaches.
Continuous diffusion/flow-based models \citep{li2022diffusion,han2023ssd,lovelace2023latent} map discrete tokens to continuous representations and apply Gaussian diffusion in the resulting continuous space.\footnote{Because continuous diffusion models are equivalent to FM with linear Gaussian interpolation, we use the terms interchangeably throughout this paper; see Section~\ref{sec:background}.}
Discrete DLMs \citep{sahoo2024simple,shi2024simplified} tailor the diffusion framework to the discrete nature of text by leveraging discrete diffusion models \citep{hoogeboom2021argmax,austin2021structured,campbell2022continuous}, which define categorical corruption processes directly in token space.
Recent scaling efforts have shown that discrete DLMs can achieve performance competitive with AR models \citep{nie2025large,you2025llada,ye2025dream,song2025seed,labs2025mercury}.
Despite these substantial advances, operating in categorical state spaces makes it difficult to apply the extensive toolkit developed for continuous diffusion models, including classifier-free guidance (CFG) \citep{ho2022classifier}, self-conditioning \citep{chen2022analog}, few-step ODE solvers \citep{lu2022dpm,lu2022dpm++}, and distillation \citep{song2024improved,yin2024one}. 
Moreover, their reliance on discrete token states may also limit their ability to exploit the rich latent geometry underlying language. 

These considerations have motivated renewed interest in continuous flow-based LMs.
Notably, recent embedding-space flow-based LMs, including LangFlow \citep{chen2026langflow}, ELF \citep{hu2026elf}, and FLM \citep{lee2026flow}, have achieved performance competitive with discrete DLMs.
However, these continuous models still rely on token-level cross entropy (CE) supervision during training---LangFlow and FLM apply the CE objective along the flow trajectory, whereas ELF combines the FM objective at intermediate denoising steps with the CE objective at the final decoding step. 
Crucially, their learned flow trajectories are not guaranteed to terminate at valid token embeddings, because errors in the learned data predictor or velocity can leave the terminal state between vocabulary embeddings.
Consequently, these models require a CE-trained decoding mechanism to map such off-embedding states to discrete tokens.
Although effective, this reintroduced discrete supervision is inconsistent with the continuous nature of flow-based models, and may limit their full potential.

Consequently, an important question remains unresolved:
\begin{quote}
% \emph{Can a continuous flow-based LM converge to valid tokens, enabling discrete token prediction without need of CE supervised decoder?}
\emph{Can the sampling trajectories of a flow-based LM converge directly to valid token embeddings, enabling discrete token prediction without a CE-supervised decoder?}
\end{quote}

\begin{figure}[t]
    \centering
    \centering
    \includegraphics[width=0.52\textwidth]{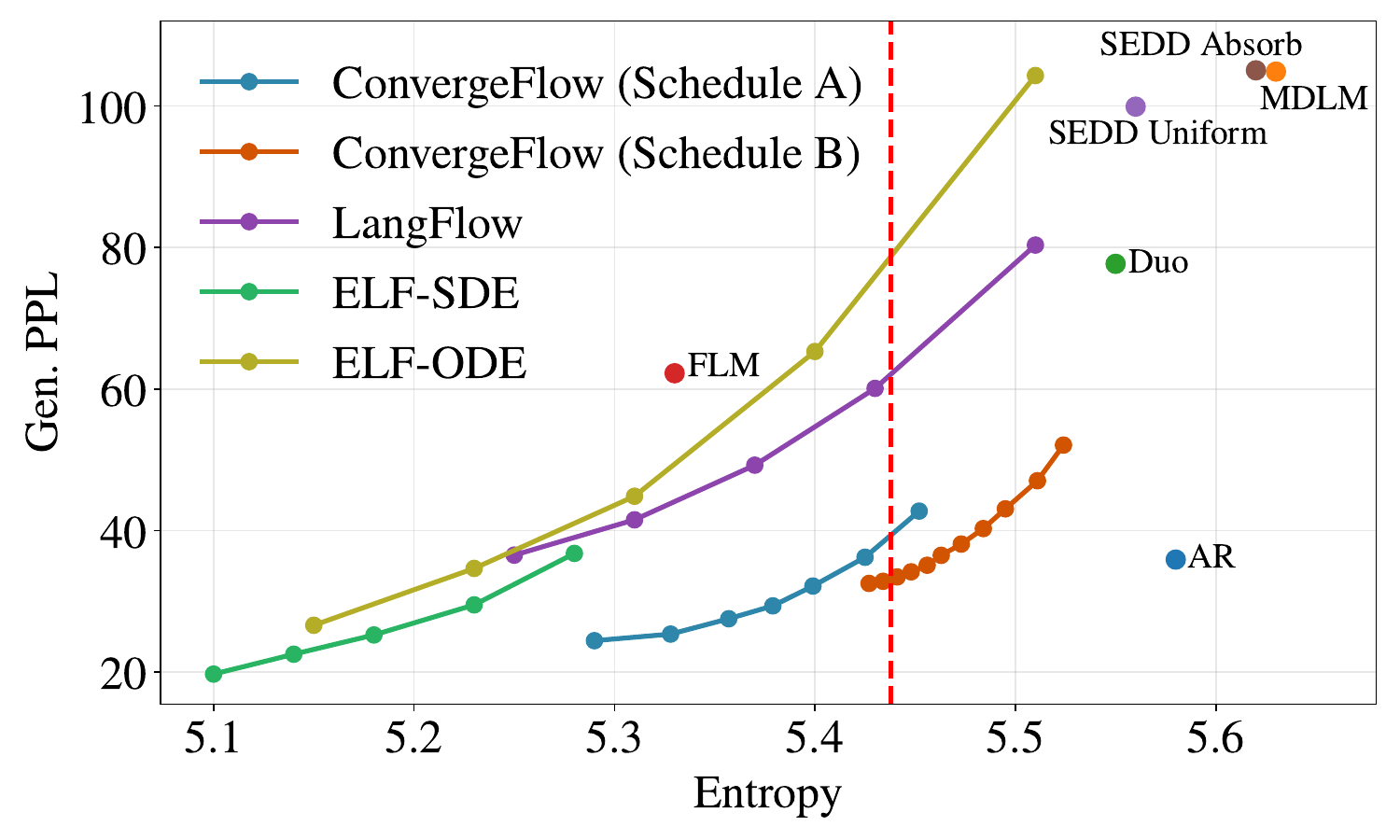}
    \caption{Gen.~PPL-entropy trade-off across different models on the OWT dataset~\citep{raffel2020exploring}. Curves show the trade-offs achieved by ConvergeFlow, LangFlow, ELF-SDE, and ELF-ODE, while individual markers indicate the reported results of the remaining baselines. Schedules A and B denote two time-adaptive guidance schedules for ConvergeFlow, defined in Section~\ref{sec:sampling}. The red dashed line marks the entropy of the OWT dataset, $5.44$.}
    \label{fig:baseline}
\end{figure}

\subsection{Contributions}

In this work, we provide an affirmative answer by introducing \textbf{ConvergeFlow}, an embedding-space flow-based LM that retains a fully continuous formulation while incorporating the discrete structure of text. 

%Motivated by the Bayes-structured factorization originally used in LangFlow, we constrain the data predictor to the convex hull of the token embeddings by expressing it as an embedding-weighted average. 
% We introduce embedding-weighted average as data predictor into flow matching, which is originally used in Plaid \citep{gulrajani2023likelihood} and LangFlow.
We parameterize the data predictor as a weighted average of vocabulary embeddings, first introduced in Plaid \citep{gulrajani2023likelihood} and later adopted by LangFlow.
Each coefficient is parameterized using a learnable weight and an exact Gaussian kernel induced by the corruption process. The resulting data predictor is trained using the mean squared error (MSE) loss induced by the FM objective.
We emphasize that LangFlow directly learns the convex-combination coefficients as token posteriors using CE supervision and subsequently maps them to a continuous data predictor. 
% Our novelty therefore lies neither in the convex-combination map nor in the Bayes factorization itself. 
Instead, ConvergeFlow uses the factorization solely as an architectural parameterization and trains the resulting continuous data predictor directly with the FM objective. 
%Consequently, the learned coefficients are not required to recover the token posterior; rather, their embedding-weighted average only needs to approximate the conditional mean of the clean data. 
% This flexibility allows the model to learn structured coefficients tailored to continuous flow matching.
% Empirically, the model achieves strong generative performance despite exhibiting a relatively high token-level CE loss, indicating that accurate recovery of the token posterior may not be necessary for effective generation.

% The corresponding coefficients are obtained by factoring the token posterior distribution into a learned weight and an exact Gaussian kernel determined by the corruption process. 
% This parametrization incorporates the token geometry into the model architecture while allowing the entire network to be trained solely with the continuous MSE flow matching loss. 

% The resulting model produces a continuous data predictor and a discrete token distribution through the same prediction, eliminating the need for token-level CE supervision or a separately terminal decoder.

Theoretically, under suitable regularity conditions, we prove that this parameterization ensures the learned flow converges to a valid token embedding despite errors in the learned data predictor.
The same data predictor can therefore drive the intermediate continuous flow updates and produce the final discrete token prediction, thereby eliminating the need for a CE-supervised token decoder. We further validate this theoretical guarantee by showing that discrete tokens can be recovered accurately and directly from the terminal flow states.

Our contributions can be summarized as twofold:

\begin{itemize}
\item \emph{Flow-based LM with provable convergence to token embeddings.}
We introduce \textbf{ConvergeFlow}, an embedding-space flow-based LM. We prove that the resulting flow provably converges to valid token embeddings, enabling direct token prediction without a CE-supervised decoder.
To our knowledge, ConvergeFlow is the first flow-based LM with provable convergence to token embeddings.

\item \emph{Quality--diversity control and strong empirical performance.}
We propose three sampling mechanisms that provide explicit control over the trade-off between generation quality, measured by generative perplexity (Gen.~PPL), and diversity, measured by entropy.
Combined with these mechanisms, ConvergeFlow achieves performance competitive with continuous baselines, including LangFlow and ELF, as well as discrete baselines such as Duo \citep{sahoo2025diffusion}.
In particular, on the OpenWebText (OWT) dataset~\citep{raffel2020exploring}, ConvergeFlow achieves a Gen.~PPL of $33.17$ while maintaining an entropy of $5.44$; see Figure~\ref{fig:baseline} and Table~\ref{tab:ppl_entropy_comparison} for details.

\begin{table}[h]
\centering
\caption[Comparison of Gen.~PPL and entropy across different models.]{Comparison of Gen.~PPL and entropy across different models. For ELF and LangFlow, we report the Gen.~PPL at the point whose entropy is closest to that of the dataset; their complete Gen.~PPL-entropy curves are shown in Figure~\ref{fig:baseline}. For ConvergeFlow, we report the Gen.~PPL at the dataset entropy using the complete results in Figure~\ref{fig:baseline} and Table~\ref{tab:guidance_nfe64}. Results and model sizes marked with $^\ddagger$ are taken from Duo. At the dataset entropy of $5.44$, our method achieves a Gen.~PPL of $33.17$, whereas the lowest Gen.~PPL among the continuous flow-based LMs is approximately $60$, even though these models are evaluated at entropies below the dataset entropy.
% \protect\footnotemark
}
\label{tab:ppl_entropy_comparison}
\small
\setlength{\tabcolsep}{10pt}
\renewcommand{\arraystretch}{1.08}

\begin{tabular}{lccc}
\toprule
Model & Gen.~PPL ($\downarrow$) & Entropy ($\uparrow$) & Model Size \\
\midrule
\textit{Autoregressive}\\
Transformer$^\ddagger$     & 35.90  & 5.58 & 170M \\
\midrule
\textit{Discrete DLMs}\\
% CANDI         & 143.13 & 5.71 & 110M \\
MDLM$^\ddagger$~\citep{sahoo2024simple}          & 104.85 & 5.63 & 170M \\
Duo$^\ddagger$~\citep{sahoo2025diffusion}           & 77.69  & 5.55 & 170M \\
SEDD Uniform$^\ddagger$~\citep{lou2023discrete}  & 99.90  & 5.56 & 170M \\
SEDD Absorb$^\ddagger$~\citep{lou2023discrete}   & 105.03 & 5.62 & 170M \\
\midrule
\textit{Continuous DLMs}\\
FLM~\citep{lee2026flow}           & 62.23  & 5.33 & 179M \\
LangFlow~\citep{chen2026langflow}      & 60.09 & 5.43 & 130M \\
ELF~\citep{hu2026elf}           & 65.30 & 5.40 & 105M \\
\textbf{ConvergeFlow}          & \textbf{33.17} & 5.44 & 130M \\
\midrule
Dataset & - & 5.44 & -\\
\bottomrule
\end{tabular}
\end{table}

\end{itemize}

\subsection{Related work}

\paragraph{Continuous DLMs.}

Continuous DLMs differ primarily in the space over which Gaussian diffusion is performed. 
At the token level, embedding-space DLMs \citep{li2022diffusion,dieleman2022continuous,gong2022diffuseq,gulrajani2023likelihood} diffuse a sequence of token-level embeddings.
Simplex-based DLMs \citep{han2023ssd,mahabadi2024tess,tae2025tess,potaptchik2026discrete,roos2026categorical} instead map each token to a point on the probability simplex over the vocabulary, while \citet{lee2026flow} adopts the one-hot encoding.
Because such token-level states may not adequately capture contextual semantics, latent DLMs perform diffusion in sequence-level latent spaces constructed from contextual representations. Earlier approaches obtain these features from the outputs of a frozen pre-trained encoder \citep{lovelace2023latent,zhang2023planner,meshchaninov2026cosmos}, whereas more recent works jointly learn the latent encoder and the diffusion model \citep{meshchaninov2026train,guo2026continuous}.

\paragraph{Discrete DLMs.}
% discrete diffusion formulated in \citep{austin2021structured} as a continuous-time Markov chain (CTMC) with categorical transition kernels 
%
Two families dominate modern discrete DLMs, distinguished by their forward corruption processes: uniform diffusion models (UDMs) and masked diffusion models (MDMs).
% The reverse-time dynamics by learning a discrete score using objectives such as concrete score matching \citep{meng2022concrete} and score entropy \citep{lou2023discrete}.
%
UDMs progressively corrupt tokens toward the uniform distribution over the vocabulary \citep{sahoo2025diffusion,sahoo2026scaling}.
MDMs instead augment the vocabulary with a special mask token as an absorbing state, and progressively replace tokens with it \citep{sahoo2024simple,shi2024simplified}. In MDMs, the discrete score \citep{meng2022concrete,lou2023discrete} is equivalent to the joint conditional distribution of the masked tokens given the unmasked context \citep{ou2024your,zheng2025masked}.
In practice, masked DLMs approximate this joint conditional by a product of token-wise conditional marginals \citep{nie2025large,arriola2025block}.
Although this enables parallel generation, it imposes a conditional independence assumption among tokens revealed in the same iteration and thus introduces an inherent factorization bias.
Consequently, the unmasking strategy---which determines the number and positions of tokens to reveal at each step---plays a critical role in the performance of masked DLMs \citep{kim2025train,wu2025fast,yu2025dimple,ben2026accelerated,fu2025bits}.

\paragraph{Theory for DLMs.}
Because masked DLMs have historically outperformed continuous DLMs and uniform DLMs, theoretical analyses for DLMs have largely focused on masked DLMs, particularly on characterizing the accuracy-speed trade-off in parallel generation. 
Early work studied fixed-size, random-ordering unmasking \citep{shi2024simplified,sahoo2024simple}, which prescribes the number of sampling steps and tokens revealed per step while selecting their positions at random.
\citet{li2025breaking} derived the first convergence guarantees for such strategies, which were subsequently sharpened using information-theoretic quantities that capture low-complexity structure in the data distribution  \citep{chen2025optimal,lavenant2025error,zhao2026adaptation,wainwright2026data,dmitriev2026efficient}.
% From a continuous-time Markov chain (CTMC) perspective, the work \citep{dmitriev2026efficient} analyzed the $\tau$-leaping sampler.
More recently, \citet{cai2026confidence} established the provable efficiency of confidence-based unmasking \citep{ben2026accelerated}, which adaptively selects both the number and positions of tokens to reveal based on the model's predictive confidence.
Parallel to these sampling convergence results, statistical generalization guarantees have been established in \citet{wakasugi2026state,zhanggeneralization}.
% , though bounds suffering from the curse of dimensionality.

\paragraph{Theory for continuous diffusion models.}
Recent years have witnessed substantial theoretical progress in continuous diffusion models \citep{lee2022convergence,chen2023probability,oko2023diffusion,cai2025minimax,wu2026diffusion}.
In particular, a line of work derives convergence guarantees for sampling from data distributions under mild assumptions, such as bounded moments, without requiring globally Lipschitz score functions \citep{chen2022sampling,chen2023improved,benton2023linear,li2024sharp,li2024d,jiao2024instance,jiao2025optimal}.
Because token-level embedding distributions have finite support and hence bounded moments, these results naturally apply to continuous DLMs and flow-based LMs.
Provably accelerated samplers based on higher-order approximations are further developed in \citep{li2024provable,li2025faster,jain2026sharp}.
Moreover, the discrete nature of text also provides additional structure where Gaussian smoothing of the discrete token-embedding distribution yields a Gaussian mixture. Exploiting this structure, nearly dimension-free convergence guarantees have been established in \citet{li2025dimension}.
\section{Background}
\label{sec:background}

\subsection{Flow matching and diffusion models}

Flow matching (FM) \citep{lipman2022flow} is a continuous-time generative modeling framework that transports a sample from a source distribution $p_0$ (typically the standard Gaussian) to a target distribution $p_1=p_\data$.

The framework consists of two steps. First, one specifies a probability path $(p_t)_{t \in (0,1)}$ interpolating between the source $p_0$ and target $p_1$. A common choice defines $p_t$ as the marginal distribution of 
\begin{align}
    \label{eq:flow_matching_path}
x_t = \alpha_t x_{\star} + \sigma_t z, \quad t\in[0,1],
\end{align}
where $x_\star \sim p_\data$ and $z\sim\cN(0,I)$ is independent Gaussian noise. The differentiable schedules $(\alpha_t,\sigma_t)_{t\in(0,1)}$ are chosen such that $\lim_{t\to 0} {\alpha_t}/{\sigma_t} = 0$ and $\lim_{t\to 1} {\alpha_t}/{\sigma_t} = \infty$. Under the standard endpoint conditions $(\alpha_0,\sigma_0)=(0,1)$ and $(\alpha_1,\sigma_1)=(1,0)$, one has $x_1 = x_\star \sim p_1$ and $x_0 = z\sim p_0$.

Second, one learns a time-dependent velocity field $v:\bR^d \times [0,1] \to \bR^d$ whose induced flow realizes the prescribed probability path. Specifically, the velocity $v(x,t)$ generates the probability path $p_t$ if the solution to the ordinary differential equation (ODE),
\begin{align}\label{eq:flow_matching_ode}
\frac{\diff x_t}{\diff t} = v(x_t,t),\quad t\in(0,1);\qquad x_0\sim p_0,
\end{align}
satisfies $x_t \sim p_t$ for all $t \in [0,1]$.

\paragraph{Training.}
A natural objective for learning the velocity $v_t$ is the \textit{flow matching loss}
\begin{align}\label{eq:flow_matching_loss}
\ell_{\mathsf{FM}}(\theta) \defn \mathbb{E}_{t, x_t}\Big[\big\|v_\theta(x_t,t) - v(x_t,t)\big\|_2^2\Big],
\end{align}
where $t\sim\unif(0,1)$ and $x_t \sim p_t$.
However, this objective cannot be evaluated directly because the velocity~$v_t$ is generally unavailable.
Fortunately, one can obtain a tractable objective by conditioning on the target $x_\star$.
%  and learning the conditional velocity field $v_t(x_t \mid x_{\star})$ that induces the conditional distribution $p_t(\cdot \mid x_{\star}) = \cN(\alpha_t x_{\star}, \sigma_t^2 I)$.
Under the prescribed probability path \eqref{eq:flow_matching_path}, consider the conditional distribution $p_t(\cdot \mid x_{\star})$ of $x_t$ given $x_{\star}$. By the path construction in \eqref{eq:flow_matching_path} and the ODE in \eqref{eq:flow_matching_ode}, the conditional velocity field is given by
\begin{align}\label{eq:conditional_velocity_field}
v(x_t,t \mid x_{\star}) &= \alpha_t' x_{\star}+\sigma_t' z   = \frac{\sigma_t'}{\sigma_t}x_t + \Big(\alpha_t'-\frac{\sigma_t'}{\sigma_t} \alpha_t \Big)x_{\star} 
= \frac{\alpha_t'}{\alpha_t}x_t + \Big(\sigma_t'-\frac{\alpha_t'}{\alpha_t} \sigma_t \Big)z.
\end{align}
% where the last two equalities follows from \eqref{eq:flow_matching_path}.
%  that $z = (x_t - \alpha_t x_{\star})/\sigma_t$. 
% A popular schedule is the linear interpolation: $\alpha_t = t$ and $\sigma_t = 1-t$, which yields $v_t(x_t \mid x_{\star}) = x_{\star}-z$.

The key observation underlying conditional flow matching (CFM) is that under mild regularity conditions,
% conditioned on $x_\star$, if a conditional velocity field $v_t(\cdot \mid x_{\star})$ can generate $p_t(\cdot \mid x_{\star})$, then 
the posterior expectation of the conditional velocity,
\begin{align}\label{eq:velocity_field}
v(x,t) = \mathbb{E}_{x_\star}[v(x_t,t \mid x_{\star})\mid x_t=x],
\end{align}
yields the marginal velocity $v_t$ that generates the probability path $p_t$ \citep{lipman2022flow}.
%
% Therefore, one can learn the velocity field $v_t^\theta$ by minimizing 
This leads to the \textit{conditional flow matching loss}:
\begin{align}\label{eq:cond_flow_matching_loss}
\ell_{\mathsf{CFM}}(\theta) \defn \mathbb{E}_{t, x_{\star}, z}\Big[\big\|v_\theta(x_t,t) - v(x_t,t \mid x_{\star})\big\|_2^2\Big],
\end{align}
where $t\sim\unif(0,1)$, $x_\star\sim p_1$, $z\sim\cN(0,I)$, and $x_t = \alpha_t x_{\star} + \sigma_t z$. 
The minimizer of the objective \eqref{eq:cond_flow_matching_loss} is given by the conditional expectation
\begin{align*}
v_{\theta^\star}(x,t) = \bE_{x_\star}\big[v(x_t,t \mid x_{\star}) \mid x_t=x\big] = v(x,t).
\end{align*}
Therefore, the CFM loss \eqref{eq:cond_flow_matching_loss} shares the same minimizer as the FM loss \eqref{eq:flow_matching_loss} and thus provides a tractable objective for learning the velocity $v_t$.
For simplicity of presentation, we will refer to the CFM loss as the FM loss in the rest of the paper.

In addition, the velocity can also be expressed via either a data predictor or a noise predictor. Define 
\begin{align}
\mu(x,t) \defn \bE[x_\star \mid x_t = x] \quad\text{and}\quad \veps(x,t) \defn \bE[z \mid x_t = x].
\end{align}
Combining \eqref{eq:conditional_velocity_field} and \eqref{eq:velocity_field}, we obtain
\begin{align}\label{eq:velocity_field_x} 
v(x,t) 
% & = \mathbb{E}_{x_\star}[\alpha_t' x_{\star}+\sigma_t' z\mid x] \\ 
& =\frac{\sigma_t'}{\sigma_t}x + \Big(\alpha_t'-\frac{\sigma_t'}{\sigma_t} \alpha_t \Big) \mu(x,t) 
= \frac{\alpha_t'}{\alpha_t}x + \Big(\sigma_t'-\frac{\alpha_t'}{\alpha_t} \sigma_t \Big) \veps(x,t) .
% \label{eq:velocity_field_noise}.
\end{align}
This identity shows that learning the velocity is equivalent to predicting the clean data or noise.
Therefore, FM is essentially the probability flow ODE in diffusion models \citep{song2020score}.
Since we use the linear Gaussian  interpolation, we use FM and diffusion models interchangeably in this paper.

Finally, for the linear schedule $\alpha_t=t$ and $\sigma_t=1-t$, the FM objective under the data prediction parameterization simplifies to
\begin{align*}
\mathbb{E}_{t, x_{\star}, z}\Big[(1-t)^{-2}\big\|x_\star - \mu_\theta(x_t,t)\big\|_2^2\Big].
\end{align*}
For general schedules, the corresponding FM objective has a schedule-dependent weighting that is not invariant across schedules at a fixed signal-to-noise ratio (SNR). To remove this dependence, we instead use the following objective:
\begin{align*}
\mathbb{E}_{t,x_\star,z}\biggl[ \Bigl(1+\frac{\alpha_t}{\sigma_t} \Bigr)^2 \bigl\|x_\star-\mu_\theta(x_{t},t)\bigr\|_2^2\biggr].
\end{align*}

\paragraph{Inference.}

After learning a velocity $v_\theta$, approximate samples from the target distribution $p_1$ can be generated by drawing $x_0\sim p_0$ from the source distribution $p_0$ and numerically solving the ODE
\begin{align}\label{eq:flow_matching_ode_inference}
\frac{\diff x_t}{\diff t} = v_\theta(x_t,t), \quad t\in[0,1].
\end{align}
In practice, one can use a forward Euler method to approximate the one-step update from $t$ to $s$:
\begin{align*}
x_s - x_t = \int_t^s v_\theta(x_\tau,\tau) \,\diff\tau \approx v_\theta(x_t,t)(s - t).
\end{align*}

The ODE in \eqref{eq:flow_matching_ode_inference} can be reparameterized using either a data predictor $\mu_\theta$ or a noise predictor $\veps_\theta$.
Replacing $\mu$ with $\mu_\theta$ in the data-prediction parameterization of the velocity in \eqref{eq:velocity_field_x} gives
\begin{align*}
\frac{\diff}{\diff t} \bigg(\frac{x_t}{\sigma_t} \bigg) = \mu_\theta(x_t,t) \frac{\diff}{\diff t} \bigg(\frac{\alpha_t}{\sigma_t} \bigg).
\end{align*}
This leads to the data prediction-based inference procedure:
\begin{align}\label{eq:data_prediction_inference}
\frac{x_s}{\sigma_s} - \frac{x_t}{\sigma_t} = \int_t^s \mu_\theta(x_\tau,\tau) \,\diff \bigg(\frac{\alpha_\tau}{\sigma_\tau} \bigg) \approx \mu_\theta(x_t,t)\bigg(\frac{\alpha_s}{\sigma_s} - \frac{\alpha_t}{\sigma_t}\bigg).
\end{align}
Similarly, the noise predictor-based inference is given by
\begin{align}\label{eq:noise_prediction_inference}
\frac{x_s}{\alpha_s} - \frac{x_t}{\alpha_t} = \int_t^s \veps_\theta(x_\tau,\tau)\,\diff\bigg(\frac{\sigma_{\tau}}{\alpha_{\tau}}\bigg) \approx \veps_\theta(x_t,t)\bigg(\frac{\sigma_s}{\alpha_s} - \frac{\sigma_t}{\alpha_t}\bigg).
\end{align}

\paragraph{Self-conditioning.}
Self-conditioning \citep{chen2022analog} is a technique that adds an additional input $c$ to the predictor.
During training, the conditional flow matching loss in \eqref{eq:cond_flow_matching_loss} is modified to
\begin{align*}
\mathbb{E}_{t, x_{\star}, z, c}\Big[\big\|v_\theta(x_t,t, c) - v(x_t,t \mid x_{\star})\big\|_2^2\Big],
\end{align*}
where the input $c$ is constructed by
\begin{align}
c = \begin{cases}
\varnothing, & \text{with probability } 1-p,\\
\mathsf{stopgrad}\big(v_\theta(x_t,t, \varnothing)\big), & \text{with probability } p.
\end{cases}
\end{align}
% \begin{align*}
% \mathbb{P}(c = \emptyset) = 1-p
% \quad\text{and}\quad
% \mathbb{P}(c = v_t(x_t, \emptyset)) = p
% \end{align*}

During training, the model makes an ordinary prediction without self-conditioning with a certain probability. Otherwise, it first produces an ordinary prediction and then uses the resulting prediction as an additional input in a second forward pass. In this way, the model learns to refine a prediction previously produced by itself.
During inference, self-conditioning is applied at every step. The self-conditioning input is initialized as empty at the first step, when no previous prediction is available, and is set to the model's prediction from the preceding step thereafter.

% \subsection{Continuous diffusion language models.}

% LangFlow \citep{chen2026langflow} does not directly learn a data predictor for the clean embedding. Instead, it first learns a probability distribution $p_\theta^{(i)}(\cdot \mid x_t,t)$ over the vocabulary for each token position in the sequence,
% \begin{align*}
% p_\theta^{(i)}(j\mid x_t,t)\approx \bP\{s^{(i)} =j \mid x_t\}, \quad i\in[L],
% \end{align*}
% via the cross-entropy (CE) loss
% \begin{align*}
% \ell_{\mathsf{CE}}(\theta) := \mathbb{E}_{t, x_\star, z}\biggl[-\sum_{i=1}^L \log p_\theta^{(i)}(s^{(i)}\mid x_t, t)\biggr].
% \end{align*}
% It then converts these distributions into a data predictor $\mu_\theta(x_t,t)$ via the embedding matrix $E$:
% \begin{align}\label{eq:data_predictor_langflow}
% \mu_\theta^{(i)}(x_t,t ) = E^\top p_\theta^{(i)}(x_t,t), \quad i\in[L].
% \end{align}

% At the final step, the model outputs token via $s^{(i)} = \argmax_j p^{(i)}_\theta(j \mid x_{t_N},t_N)$ for each position $i\in[L]$.

\subsection{Evaluation metrics for language modeling}
We briefly review the metrics commonly used to evaluate language models. Let $p_\theta$ denote the distribution induced by a trained language model.

Because the data distribution $p_\data$ is unknown, 
% we cannot directly evaluate the negative log-likelihood (NLL) of the model under the data distribution. Instead, 
the quality of the trained model is often assessed using a reference language model such as the GPT-2 Large model \citep{radford2019language}. Specifically, let $p_{\reference}$ denote the distribution of the reference model.
%  used to evaluate perplexity.
\textit{Generative perplexity} (Gen.~PPL) is defined as
\begin{align}\label{eq:generative_ppl}
\GenPPL(p_{\theta};p_{\reference}) \defn \exp\biggl(-\frac{1}{L}\mathbb{E}_{X \sim p_{\theta}}[\log p_{\reference}(X)]\biggr),
\end{align}
which satisfies the following relationship:
\begin{align}\label{eq:genppl-identiry}
\log \GenPPL(p_{\theta};p_{\reference}) 
= \frac1L \KL(p_{\theta} \,\|\, p_{\reference}) + \frac1L  H(p_\theta),
\end{align}
where $\KL(p_{\theta} \,\|\, p_{\reference})$ is the Kullback-Leibler (KL) divergence between the model distribution $p_\theta$ and the reference distribution $p_{\reference}$, and $H(p_\theta)$ denotes the entropy of $p_\theta$.

Identity \eqref{eq:genppl-identiry} reveals that low Gen.~PPL may arise either because the generated samples have high likelihood under the reference model or because the model concentrates its probability mass on a small set of likely sequences. 
Therefore, it is common to evaluate the entropy of the model distribution alongside its Gen.~PPL as a measure of diversity.

In practice, entropy is often approximated using \textit{unigram entropy}.
%  \citep{sahoo2025diffusion,hu2026elf,chen2026langflow}.
For a sequence $x=(x^{(1)},\ldots,x^{(L)})$, define its empirical unigram distribution by
\begin{align*}
\wh p_x(v) = \frac1L\sum_{i=1}^L \ind\{x^{(i)} = v\}, \quad \forall\,v,
\end{align*}
and let $H(\wh p_x)$ denote the corresponding entropy.
The unigram entropy is then defined as 
\begin{align}\label{eq:unigram_entropy}
H_{\uni}(p_{\theta}) \defn \bE_{X\sim p_{\theta}}[H(\wh p_X)].
\end{align}
% Because the unigram entropy ignores token order and cross-token dependence, it is not, in general, equal to the normalized sequence entropy $L^{-1}H(p_\theta)$ appearing in \eqref{eq:genppl-identiry}. Nevertheless, it provides a simple diagnostic of within-sequence diversity and can help detect low-diversity degeneration that may artificially improve Gen.~PPL.
The unigram entropy serves as a proxy for the normalized sequence entropy $L^{-1}H(p_\theta)$ appearing in \eqref{eq:genppl-identiry}, providing a simple diagnostic of within-sequence diversity.

In practice, the expectations defining Gen.~PPL and unigram entropy are approximated by averaging over independent samples.

% \begin{align*}
% \entropy &\approx -\mathbb{E}_{X \sim p_{\theta}}\bigg[\sum_x\widehat{p}_{\theta}(x)\log \widehat{p}_{\theta}(x)\bigg] \\
% &\approx -\sum_xp_{\theta}(x)\log p_{\theta}(x) \\
% &= \frac{1}{L}\sum_i H_{p_{\theta}}(X^{(i)}) \ge \frac{1}{L}H_{p_{\theta}}(X).
% \end{align*}

% To quantify diversity, we use the average per-token entropy
% \begin{align}
% \bar H(p_\theta)
% &\defn
% \frac{1}{L}\sum_{i=1}^L H_{p_\theta}(X^{(i)}).
% \label{eq:evaluation_entropy}
% \end{align}
% Since $\sum_i H(X^{(i)})\geq H(X)$, define the dependence gap
% \begin{align*}
% \Delta(p_\theta) \defn \bar H(p_\theta)-\frac{1}{L}H_{p_\theta}(X)
% \geq0.
% \end{align*}
% The normalized divergence from the reference language model then satisfies
% \begin{align*}
% \frac{1}{L}\mathsf{KL}(p_{\theta} \parallel p_{\reference}) = \log \PPL(p_{\theta};p_{\reference}) - \entropy(p_{\theta}) + \Delta(p_{\theta}).
% \end{align*}
\section{ConvergeFlow}
\label{sec:result}

% In this section, we introduce a novel framework for embedding-space DLMs. We first describe the , which imposes a structure on the data predictor that ensures the flow to converge to token embeddings. We then present three sampling techniques for achieving trade-offs between perplexity and entropy in generated text.

\subsection{Framework and convergence theory}

Let $s = (s^{(1)}, \ldots, s^{(L)})$ be a token sequence of length $L$ drawn from the data distribution $p_\data$, where each token $s^{(i)}$ belongs to a vocabulary of size $V$.
Without loss of generality, we assume the vocabulary is $[V]\defn\{1,\dots,V\}$.
We map tokens to continuous representations using an embedding matrix $E\in\mathbb{R}^{V\times d}$, where $d$ is the embedding dimension. For each $j\in[V]$, let $e_j^\top \defn E_{j,:} \in \bR^d$ denote the $j$-th row of the embedding matrix $E$, representing the embedding of the $j$-th token in the vocabulary. 
The target $x_\star$ is the continuous representation of the token sequence, given by
\begin{align}
x_\star = \bigl[e_{s^{(1)}}, \ldots, e_{s^{(L)}}\bigr]^\top \in \mathbb{R}^{L\times d}.
\end{align}

We consider FM with general interpolation schedules $(\alpha_t, \sigma_t)$:
\begin{align*}
x_t = \alpha_t x_\star + \sigma_t z,
\end{align*}
where $z\in\bR^{L\times d}$ is a standard Gaussian random matrix with i.i.d.~entries $z_{ij} \overset{\mathrm{i.i.d.}}{\sim} \cN(0,1)$.
Using the data-prediction parameterization, we train a data predictor $\mu_\theta:\bR^{L\times d}\times[0,1] \to \bR^{L\times d}$ using the MSE loss induced by the FM objective:
\begin{align}\label{eq:diffusion_loss}
\mathbb{E}_{t,x_\star,z}\biggl[ \Bigl(1+\frac{\alpha_t}{\sigma_t} \Bigr)^2 \bigl\|x_\star-\mu_\theta(x_{t},t)\bigr\|_{\mathrm{F}}^2\biggr].
\end{align}
This objective, also used by ELF, provides purely continuous supervision and does not involve a token-level CE loss.

An important caveat is that the FM objective alone does not sufficiently supervise the joint learning of the embedding matrix and the data predictor. Because the target $x_\star$ itself is defined by the embedding matrix, the FM objective admits degenerate embedding-collapse solutions. We therefore use the pre-trained embedding matrix from LangFlow and keep it fixed throughout training and inference.

We note that for language data, each row of the target $x_\star$ is supported on a finite collection of token embeddings rather than an unrestricted Euclidean space. The MSE loss in \eqref{eq:diffusion_loss}, however, treats the data predictor $\mu_\theta(x_t, t)$ as an unconstrained regressor and therefore fails to exploit this discrete support. 
This observation motivates the structured parameterization introduced next.

\paragraph{Embedding-weighted data predictor.}

Fix a token position $i\in[L]$.
The clean embedding of the token at position $i$ is $x_\star^{(i)} = e_{s^{(i)}}.$
Observe that its conditional expectation given $x_t$ is a weighted average of all token embeddings according to the posterior token distribution:
\begin{align}\label{eq:posterior_token_distribution}
\mathbb{E}[x_{\star}^{(i)}\mid x_t] = \sum_{j=1}^V \bP\{s^{(i)} = j \mid x_t\} \,e_j.
\end{align}
Consequently, the Bayes-optimal data predictor under the MSE loss \eqref{eq:diffusion_loss} lies in the convex hull of the vocabulary embeddings.
Moreover, the following proposition reveals a useful multiplicative structure of the posterior distribution: it can be factored into a context-only posterior and an exact Gaussian kernel. The proof is deferred to Appendix~\ref{sec:proof_posterior_token_distribution_bayes}.
\begin{proposition}\label{lem:posterior_token_distribution_bayes}
For $x=[x^{(1)},\ldots,x^{(L)}]^\top\in\bR^{L\times d}$, denote
\begin{align}
x^{(-i)} \defn
\bigl[x^{(1)},\ldots,x^{(i-1)},x^{(i+1)},\ldots,x^{(L)}\bigr]^\top \in\bR^{(L-1)\times d}.
\end{align}
The posterior token distribution satisfies
\begin{align}\label{eq:posterior_token_distribution_bayes2}
    \bP\{s^{(i)} = j \mid x_t\} 
    \propto \bP\{s^{(i)} = j \mid x_t^{(-i)}\} \exp\bigl(-\|x_t^{(i)} - \alpha_t e_j\|_2^2/(2\sigma_t^2)\bigr).
\end{align} 
\end{proposition}

% Therefore, any estimator $w_\theta^{(i)}:\bR^d \times [0,1] \mapsto \Delta([V])$ of the posterior token distribution naturally induces a data predictor in the continuous embedding space:
% \begin{align}\label{eq:data_predictor_from_token_weights}
% \mu_\theta^{(i)}(x_t, t) = \sum_{j=1}^V w_\theta^{(i)}(j\mid x_t, t)e_j = E^\top w_\theta^{(i)}(\cdot\mid x_t, t).
% \end{align}
% On the other hand, we notice that since the embedding dimension $d$ is typically smaller than the vocabulary size $V$, the convex combination is not unique. Hence, \eqref{eq:posterior_token_distribution} shows the existence of a data predictor $\mu_\theta^{(i)}$ that is a convex combination of token embeddings, but only necessarily implies we need to strictly learn the posterior token distribution $w_\theta^{(i)}$.
%

The identity in \eqref{eq:posterior_token_distribution} establishes an existence result---the posterior probabilities constitute one set of convex weights whose embedding-space barycenter equals the conditional mean of the clean embedding.
However, these weights are generally not unique. Because the vocabulary size $V$ is typically much larger than the embedding dimension $d$, distinct convex weights can produce exactly the same data prediction. Nevertheless, the convex structure in \eqref{eq:posterior_token_distribution} suggests a useful parameterization.
% Consequently, identifying the posterior probabilities may be ill-posed.
% Nevertheless, the convex structure in \eqref{eq:posterior_token_distribution} suggests the useful parameterization. 

Consequently, our goal is not to learn the posterior distribution itself, but to learn a valid set of convex weights whose embedding-space barycenter accurately predicts the conditional mean. Inspired by the multiplicative form in Proposition~\ref{lem:posterior_token_distribution_bayes}, we parameterize these convex coefficients using a learned base weight function and the known Gaussian corruption kernel. Specifically, we learn a base weight function $f_\theta^{(i)}:\bR^{L\times d} \times [0,1] \mapsto \Delta([V])$, which is optimized using the MSE loss in \eqref{eq:diffusion_loss}. Importantly, $f_\theta^{(i)}$ is neither supervised nor interpreted as a token posterior. In particular, it is not intended to estimate the distribution $\bP\{s^{(i)} = \cdot \mid x_t^{(-i)}\}$ appearing in Proposition~\ref{lem:posterior_token_distribution_bayes}. Rather, the proposition motivates only the form of the parameterization.
We then define the convex weights
\begin{align}
w_\theta^{(i)}(j \mid x_t, t) = \frac{f_\theta^{(i)}(j \mid x_t, t)\exp\bigl(-\|x_t^{(i)} - \alpha_t e_j\|_2^2/(2\sigma_t^2)\bigr)}{\sum_{j'\in[V]} f_\theta^{(i)}(j' \mid x_t, t)\exp\bigl(-\|x_t^{(i)} - \alpha_t e_{j'}\|_2^2/(2\sigma_t^2)\bigr)}, \quad j\in[V].
\end{align}
The resulting data predictor for the token at position $i$ is given by
\begin{align}
\mu_\theta^{(i)}(x_t, t) = \sum_{j=1}^V w_\theta^{(i)}(j\mid x_t, t)e_j = E^\top w_\theta^{(i)}(\cdot\mid x_t, t). \label{eq:posterior_structured_data_predictor}
\end{align}
Applying \eqref{eq:posterior_structured_data_predictor} to each token position $i$ and stacking the outputs yields the full data predictor $\mu_\theta(x_t, t)$.

In summary, our proposed parameterization preserves the target of unconstrained MSE data prediction while explicitly incorporating both the discrete token structure and the known Gaussian corruption.

\paragraph{Provable convergence to token embeddings.}
Notably, our proposed parameterization for the data predictor guarantees convergence of the sampling trajectory to a valid token embedding. 
This is formalized below, with the proof deferred to Appendix~\ref{sec:proof_continuous_flow}.
\begin{theorem}[Flow convergence to token embeddings]\label{thm:continuous_flow}

Assume that, for every token position $i\in[L]$, the learned base weight function satisfies $f_\theta^{(i)}(j\mid x_t,t) > 0$ for any $j\in[V]$, state $x_t$, and time $t\in(0,1)$. Moreover, assume that the log-weight is Lipschitz continuous along the sampling trajectory: there exists a constant $\widetilde{L}$ such that
\begin{align*}
\max_{j\in[V]}\big|\log f_\theta^{(i)}(j\mid x_{t}, t) - \log f_\theta^{(i)}(j\mid x_{\tau}, \tau)\big|
\le \widetilde{L}\,|t - \tau|, \quad \forall\,  0<t,\tau<1.
\end{align*}
Finally, assume that the time grid $0 = t_0 < t_1 < \ldots < t_N < 1$ satisfies $t_N \to 1$ as $N \to \infty$ and
\begin{align*}
\max_{0\leq k < N}\frac{t_{k+1} - t_k}{(1 - t_{k+1})^3} < \delta
\end{align*}
for a sufficiently small $\delta > 0$.
Then, for each token position $i\in[L]$, there exists some $j_{i}\in[V]$ such that
\begin{align}\label{eq:thm-convex-combine-mu}
x_{t_N}^{(i)} \to e_{j_i}~~\text{in probability~~~~as}~~N \to \infty.
\end{align}
\end{theorem}

Theorem~\ref{thm:continuous_flow} shows that every token-level state converges to a valid token embedding. Consequently, provided that the vocabulary embeddings are distinct, nearest-neighbor decoding naturally yields the token prediction. In particular, no separately trained terminal decoder is required. We empirically validate this convergence behavior in Section~\ref{sec:experiment}.

We next explain why data prediction accuracy alone does not guarantee convergence to token embeddings, and why additional structure, such as our convex-structured parameterization, is necessary. In particular, an unconstrained data predictor may be asymptotically accurate along any corruption path while its induced flow fails to converge to any token embedding.

The following proposition provides a concrete counterexample. Its proof is deferred to Appendix~\ref{sec:proof_counterexample}.
\begin{proposition}
\label{prop:counterexample}
There exists a smooth, unconstrained data predictor $\mu_\theta$ such that
\begin{align*}
\mu_\theta(x_t, t) \to x_\star~~\text{in probability~~~~as}~~t \to 1,
\end{align*}
where $x_t = \alpha_t x_\star + \sigma_t z$ and $z_{ij} \overset{\mathsf{i.i.d.}}{\sim} \mathcal{N}(0, 1)$.
Let $x_{t_N}$ denote the flow output induced by this data predictor on a time grid $0 = t_0 < t_1 < \ldots < t_N < 1$ with $t_N \to 1$ as $N \to \infty$.
Then there exists a constant \(c_{\mathsf{lb}}>0\), independent of \(N\), such that, for every token position \(i\in[L]\) and all sufficiently large \(N\),
\begin{align}
\mathbb{P}\left\{
\min_{j\in[V]}
\bigl\|
\alpha_{t_N}^{-1}x_{t_N}^{(i)}-e_j
\bigr\|_2
\geq 1
\right\}
\geq
c_{\mathsf{lb}}.
\label{eq:counterexample}
\end{align}
\end{proposition}

% Concretely, we show that there exists a sufficiently smooth data predictor $\mu_{\theta}(x_t, t)$ obeying $\mu_{\theta}(x_t, t) \to x_\star$ as $t \to 1$. the final output $x_{t_N}$ may not converge to any token embedding with constant probability. 
% Concretely, there exists some constant $c_{\mathsf{lb}} > 0$, independent of $N$, such that for any $i\in[d]$ and all sufficiently large $N$,
% \begin{align}\label{eq:counterexample}
% \mathbb{P} \biggl\{\min_{j\in[V]}\|\alpha_{t_N}^{-1}x_{t_N}^{(i)} - e_j\|_2 \ge 1\biggr\} \ge c_{\mathsf{lb}}
% \end{align}

This counterexample demonstrates that a smooth, asymptotically accurate data predictor does not by itself ensure convergence to the discrete vocabulary. The convex-structured parameterization provides sufficient structure to guarantee convergence to a valid token embedding.

%In addition, we found that the variance during training can be decreased significantly through ...
% Moreover, since $\mu_{\theta}^{(i)}(x_t, t) = \sum_{j=1}^V w_{\theta}^{(i)}(j\mymid x_t, t)e_j$ with $w_{\theta}^{(i)}(j\mymid x_t, t) \ge 0$ and $\sum_{j=1}^V w_{\theta}^{(i)}(j\mymid x_t, t)=1$, the following proposition holds, with the proof deferred to Appendix \ref{app:proof-prop-2}.
% It allows us to select the $j$-th embedding with $j=\arg\max_{j} w_{\theta}^{(i)}(j\mymid x_t, t)$, even though we didn't use CE loss to learn $w_{\theta}^{(i)}(j\mymid x_t, t)$.
Moreover, recall that the data predictor is a convex combination of the token embeddings, with weights given by $w_\theta^{(i)}$. The following proposition shows that if the data predictor converges to a token embedding, then the corresponding weight vectors converge to a one-hot vector. The proof is deferred to Appendix~\ref{app:proof-prop-2}. 
\begin{proposition}\label{prop:dis-weight-converge}
Assume that the token embeddings $\{e_j\}_{j\in[V]}$ have the same norm and are pairwise separated, namely, $\max_{j\neq j'}\big|\frac{\langle e_j, e_j' \rangle}{\|e_j\|_2\|e_j'\|_2}\big| \le 1-\rho$ for some constant $\rho>0$.
If there exists some $j\in[V]$ such that $\mu_{\theta}^{(i)}(x_t, t) \to e_j$ as $t \to 1$, then one has
\begin{align}
w_{\theta}^{(i)}(\cdot\mymid x_t, t) \to \delta_j \quad \text{as}~~t \to 1,
\end{align}
where $\delta_{j}\in\mathbb{R}^V$ denotes the one-hot vector associated with token $j$.
\end{proposition}

\paragraph{Comparison with embedding-space flow-based LMs.}
\begin{itemize}
\item \textit{Comparison with ELF \citep{hu2026elf}.}
Both ELF and our framework use the MSE objective to train a data predictor. 
ELF directly learns the data predictor as an unconstrained regressor.
At the final step, it invokes a distinct trained decoder.
In contrast, we impose additional structure on the data predictor through \eqref{eq:posterior_structured_data_predictor}.
This guarantees that the flow automatically converges to a token embedding, so the final decoding does not require a separate decoder.

    \item \textit{Comparison with LangFlow \citep{chen2026langflow}.}
Both LangFlow and our framework produce a data predictor through a convex combination of vocabulary embeddings. LangFlow directly  learns the posterior distribution and trains it using the discrete CE objective. In contrast, we parameterize each convex weight using a learned base weight and an exact Gaussian likelihood, and train the resulting data predictor using the continuous MSE objective \eqref{eq:diffusion_loss}. 
% Moreover, compared to the CE objective, we found that the important thing is the network structure for the predictor instead of the objective.
% The diffusion loss can produce a better result, as evidenced by ...
\end{itemize}

\subsection{Sampling}
\label{sec:sampling}
Given a trained data predictor $\mu_\theta$, we generate samples by solving the data prediction-based ODE in \eqref{eq:data_prediction_inference}, initialized with a standard Gaussian random matrix $x_{t_0}$ with i.i.d.~$\cN(0,1)$ entries.

Given $N$ sampling steps and a time grid $t_0 < t_1 < \ldots < t_N$, the ODE can be solved using the first-order Euler method, yielding the following update rule:
\begin{align}\label{eq:data_prediction_update}
\frac{x_{t_{i+1}}}{\sigma_{t_{i+1}}} - \frac{x_{t_i}}{\sigma_{t_i}} = \mu_{t_i}\bigg(\frac{\alpha_{t_{i+1}}}{\sigma_{t_{i+1}}} - \frac{\alpha_{t_i}}{\sigma_{t_i}}\bigg),
\end{align}
where $\mu_{t_i}= \mu_{\theta}(x_{t_i}, t_i)$ denotes the data prediction used at step $i$. 
As we will see momentarily, the data prediction $\mu_{t_i}$ can be constructed in various ways, leading to different trade-offs between Gen.~PPL and entropy.

After the last step, we convert the generated embedding $x_{t_N}$ into a token sequence $\wh s = (\wh s^{(1)}, \ldots, \wh s^{(L)})$ by taking the nearest neighbor in the embedding space for each token position:
\begin{align}\label{eq:nearest_neighbor_decoding}
\wh s^{(i)} = \argmin_{j\in[V]} \|x_{t_N}^{(i)} - e_j\|_2, \quad i\in[L].
\end{align}
Alternatively, we can use the trained weights $w_\theta(x_{t_N},t_N)$ as the token distribution for decoding, i.e.,
\begin{align}\label{eq:weight_decoding}
\wh s^{(i)} = \argmax_{j\in[V]} w_\theta^{(i)}(j \mid x_{t_N}, t_N), \quad i\in[L].
\end{align}
Notably, both token prediction rules are parameter-free and require no separately trained terminal decoder.

Next, we describe sampling with self-conditioning.
%
% For a fixed state $x_t$, define the self-conditioning map
% \begin{align}\label{eq:data_prediction_map}
% F_t(c) \defn \mu_\theta(x_t, t, c).
% \end{align}
%
The data prediction in the ideal two-pass implementation of self-conditioning is given by
\begin{align*}
    % \label{eq:self_conditioning_ideal}
% \frac{x_{t_{i+1}}}{\sigma_{t_{i+1}}} - \frac{x_{t_i}}{\sigma_{t_i}} = \mu_\theta\bigl(x_{t_i}, t_i, \mu_\theta(x_{t_i}, t_i, \varnothing)\bigr)\bigg(\frac{\alpha_{t_{i+1}}}{\sigma_{t_{i+1}}} - \frac{\alpha_{t_i}}{\sigma_{t_i}}\bigg).
\mu_{t_i} = \mu_\theta\bigl(x_{t_i}, t_i, \mu_\theta(x_{t_i}, t_i, \varnothing)\bigr).
\end{align*}
% where
% \begin{align*}
% \mu_{t_i} = F_{t_i}(c)
% \quad\text{with}\quad
% c = \mu_\theta(x_{t_i}, t_i, \varnothing).
% \end{align*}
To reduce the computational burden, it is common to construct the data prediction $\mu_{t_{i}}$ at step $i$ using that from the preceding step $i-1$ as the self-conditioning input, in place of the same-step unconditional prediction $\mu_\theta(x_{t_i}, t_i, \varnothing)$, i.e.,
\begin{align}\label{eq:self_conditioning_practical}
\mu_{t_0} = \mu_\theta\bigl(x_{t_0}, t_0, \varnothing\bigr) \quad\text{and}\quad \mu_{t_i} = \mu_\theta\bigl(x_{t_i}, t_i, \mu_{t_{i-1}}\bigr),\,\,\,i\geq 1.
\end{align}
The data prediction $\mu_{t_i}$ is then used in the sampling update \eqref{eq:data_prediction_update}.

One can expect that $\mu_{t_{i-1}}\approx \mu_\theta(x_{t_i}, t_i, \varnothing)$ because $t_{i-1}$ and $t_{i}$, as well as $x_{t_{i-1}}$ and $x_{t_{i}}$ are close when the solver uses sufficiently many steps.
However, we observe that such a computational shortcut also introduces a deeper self-conditioning recursion, which will be elaborated later.

\paragraph{Controlling Gen.~PPL-entropy trade-off.}
We introduce three inference mechanisms for controlling the trade-off between Gen.~PPL and entropy; see Table \ref{tab:guidance_comparison} for a summary.
\begin{itemize}
\item \emph{self-conditioning guidance.}
Motivated by classifier-free guidance (CFG) \citep{ho2022classifier}, we introduce a CFG-type guidance for self-conditioning.
At each solver step $i$, we form the guided data prediction via the unconditional and one-step self-conditioned predictions:
\begin{align}\label{eq:self_conditioning_guided_sampler}
\mu_{t_i}^{\scg} = \mu_\theta(x_{t_i}, t_i, \varnothing) + w_{\scg}\bigl(\mu_\theta(x_{t_i}, t_i, c_{t_i}) - \mu_\theta(x_{t_i}, t_i, \varnothing) \bigr) \quad \text{with}\quad c_{t_i} = \mu_\theta(x_{t_i}, t_i, \varnothing).
\end{align}
When $w_{\scg} = 0$, this reduces to sampling without self-conditioning; when $w_{\scg} = 1$, it recovers the ordinary sampling with self-conditioning. Values of $w_{\scg} > 1$ extrapolate beyond the self-conditioned prediction and amplify the refinement induced by self-conditioning.

Although the form in \eqref{eq:self_conditioning_guided_sampler} resembles CFG, the condition here is generated by the model itself rather than supplied externally.

% Finally, this CFG-type extension can be combined with the unconditional guidance in \eqref{eq:unconditional_guided_sampler}.

\item \emph{Iterative self-conditioning refinement.}
Recall the computational shortcut for self-conditioning in \eqref{eq:self_conditioning_practical}, which reuses the data prediction from the previous step.
Unrolling this recursion over $K$ steps shows that the data prediction $\mu_{t_i}$ at time $t_i$ satisfies
\begin{align*}
\mu_{t_{i-j}} = \mu_\theta \bigl(x_{t_{i-j}},t_{i-j},\mu_{t_{i-j-1}}\bigr), \quad j = 0, \ldots, K-1,
\end{align*}
or equivalently,
\begin{align*}
\mu_{t_i} = \mu_\theta \Bigl(x_{t_i},t_i,\mu_\theta \bigl(x_{t_{i-1}},t_{i-1},\dots,\mu_\theta(x_{t_{i-K+1}},t_{i-K+1},\mu_{t_{i-K}})\dots\bigr)\Bigr).
\end{align*}
% \begin{align*}
% \mu_{t_i} = F_{t_i} \circ F_{t_{i-1}} \circ \cdots \circ F_{t_{i-n+1}}(\mu_{i-n}).
% \end{align*}
If the time grid is sufficiently fine, then the state and time vary little over these $K$ steps, and $\mu_\theta(x_{t_{i-j}},t_{i-j},c) \approx \mu_\theta(x_{t_{i}},t_{i},c)$ for $j = 0, \ldots, K-1$. Consequently, the data prediction $\mu_{t_i}$ is approximately given by
\begin{align*}
\mu_{t_i} \approx \mu_\theta \Bigl(x_{t_i},t_i,\mu_\theta \bigl(x_{t_i},t_i,\dots,\mu_\theta(x_{t_i},t_i,\mu_{t_{i-K}})\dots\bigr)\Bigr),
\end{align*}
where $K$ recursive evaluations are all applied to the current state $x_{t_i}$ and time $t_i$.
% In words, 
% Thus, applying self-conditioning once at each solver step approximately resembles repeatedly applying the same self-conditioning map $\mu_\theta (x_{t_{i}},t_{i},\cdot)$ at a fixed state and time.
Thus, reusing the previous data prediction in self-conditioning implicitly produces a recursive refinement whose effective depth depends on the number and spacing of solver steps.

Motivated by this observation, we make the self-conditioning refinement explicit. At each sampling step $i$, we define
\begin{align}\label{eq:explicit_iterative_self_conditioning}
u^0_{t_i} \defn \mu_\theta(x_{t_i}, t_i, \varnothing), \qquad u^k_{t_i} \defn \mu_\theta(x_{t_i}, t_i, u^{k-1}_{t_i}), \quad k=1,\dots,K,
\end{align}
and use $u^K_{t_i}$ as the data prediction in the solver update.
This construction makes the recursion depth $K$ an explicit hyperparameter, thereby decoupling it from the number of solver steps.

% Let's define
% \begin{align*}
% \mu_{t_i}^{k} = \mu_{t_i}(x_{t_i}, \mu_{t_i}^{k-1})
% \quad\text{with }\mu_{t_i}^{0} = \mu_\theta(x_{t_i}, t_i, \varnothing),
% \end{align*}
% which is closely related to the current implication of self-condition in the following sense.

% Recall that in practice, $c = \mu_{t_{i-1}}$ when calculating $\mu_{t_i}(x_{t_i}, c)$.
% Then for $t_{i - n}$ and $t_i$ close enough, we have
% \begin{align*}
% \mu_{t_{i-j}} = \mu_{t_{i-j}}(x_{t_{i-j}}, \mu_{t_{i-j-1}}) \approx \mu_{t_i}(x_{t_i}, \mu_{t_{i-j-1}}),
% \end{align*}
% and then
% \begin{align*}
% \mu_{t_i} \approx \mu_{t_i}^{n},
% \end{align*}
% where
% \begin{align*}
% \mu_{t_i}^{k} = \mu_{t_i}(x_{t_i}, \mu_{t_i}^{k-1})
% \quad\text{with }\mu_{t_i}^{0} = \mu_{t_{i-n}}.
% \end{align*}

% We test the performance of the sampler with iterative self-conditioning refinement, as well as the standard self-conditioning that reuses previous data predictor \eqref{eq:self_conditioning_practical}.
Empirically, we observe that iterative self-conditioning refinement is less sensitive to the solver-step count than the standard self-conditioning shortcut in \eqref{eq:self_conditioning_practical}.
%
% One plausible explanation is that the iterative self-conditioning refinement controls the recursion depth directly and may therefore be less sensitive to the solver-step count. In contrast, reusing the previous prediction in the standard self-conditioning produces a recursion whose effective depth changes with the number of solver steps.
Moreover, varying the depth $K$ provides an effective means of controlling the trade-off between Gen.~PPL and entropy.

\item \emph{Unconditional guidance.}
We note that improving PPL is equivalent to increasing $\log p(x)$, so the most efficient way is to move in the direction of $\nabla \log p(x)$.
By Tweedie's formula, we have
\begin{align*}
\nabla \log p_t(x) = -\frac{\veps(x,t)}{\sigma_t},
\quad \text{with} \quad
\veps(x,t) = \frac{x - \alpha_t\mu(x,t)}{\sigma_t}.
\end{align*}
Recall the standard noise prediction-based update rule from \eqref{eq:noise_prediction_inference}:
\begin{align*}
\frac{x_{t_{i+1}}}{\alpha_{t_{i+1}}} - \frac{x_{t_i}}{\alpha_{t_i}} = \bigg(\frac{\sigma_{t_{i+1}}}{\alpha_{t_{i+1}}} - \frac{\sigma_{t_i}}{\alpha_{t_i}}\bigg)\veps_\theta(x_{t_i},{t_i}).
\end{align*}
As $\sigma_t/\alpha_t$ decreases along the sampling process, the coefficient on the right-hand side is negative and the update therefore moves in the direction of $-\veps_\theta$. 
To strengthen this motion, we multiply the update by a factor of $1 + w_{\ug}$, resulting in the following sampler:
\begin{align}\label{eq:unconditional_guided_sampler}
\frac{x_{t_{i+1}}}{\alpha_{t_{i+1}}} - \frac{x_{t_i}}{\alpha_{t_i}} = (1 + w_{\ug})\bigg(\frac{\sigma_{t_{i+1}}}{\alpha_{t_{i+1}}} - \frac{\sigma_{t_i}}{\alpha_{t_i}}\bigg)\veps_\theta(x_{t_i},{t_i}).
\end{align}
% Empirically, we find it beneficial to apply weaker guidance in the high-noise regime and stronger guidance near the data endpoint. Hence, we use a time-dependent guidance coefficient
% \begin{align}
% w_{\ug}(t) = \frac{w_{\ug}}{1+\sigma_{t}/\alpha_{t}}.
% \end{align}

\end{itemize}

\begin{table}[htbp]
\centering
\caption{Summary of sampling techniques for quality-diversity trade-offs.}
\label{tab:guidance_comparison}
\begin{tabular}{p{0.25\linewidth} p{0.18\linewidth} p{0.39\linewidth}}
\toprule
Technique
&
Control parameter
&
Intended effect
\\
\midrule
Self-conditioning guidance
&
Coefficient $w_{\scg}$
&
Amplify refinement from self-conditioning
\\
Iterative self-conditioning refinement
&
Iteration count $K$
&
Refine data prediction
\\
Unconditional guidance
&
Coefficient $w_{\mathsf{ug}}$
&
Strengthen movement toward gradient of PPL
\\
\bottomrule
\end{tabular}
\end{table}

\section{Experiments}
\label{sec:experiment}

% Notice that when all token embeddings $x_{\star}$ are the same, the diffusion loss will be zero.
% Hence, training embeddings with solely diffusion loss is impossible, which needs extra design, such as CE loss.
% Here, we use the pretrained token embeddings from LangFlow, and fix them during the training with diffusion loss.
% Moreover, to save time, we make use of the weight of LangFlow to initialize our model except for the test of convergence.
% Particularly, we run our model with $4\times$A100 40GB.
% Hence, we set a little smaller batch size $480$ to make GPUs more efficient, and choose a smaller learning rate $1E-5$.
% Below, we consider the sampler~\eqref{eq:sampler} without using any extra techniques as baseline.

\paragraph{Dataset.}
We follow the experimental setup used in the literature on DLMs~\citep{chen2026langflow,hu2026elf,sahoo2025diffusion}.
We conduct all experiments on the OpenWebText (OWT) dataset~\citep{raffel2020exploring}, which contains approximately 9B tokens, and pack the text into sequences of length $L=1024$.

\paragraph{Training.}
We follow the architecture and setup of LangFlow~\citep{chen2026langflow}.
We use the same DiT-style Transformer architecture \citep{peebles2023scalable} as LangFlow, which consists of $12$ layers, a hidden dimension of $768$, and $12$ attention heads, totaling approximately $130$M parameters. 
Self-conditioning is applied during training with probability $0.25$.
% Notably, optimizing the token embeddings jointly with the diffusion loss admits a degenerate solution in which all token embeddings $x_{\star}$ collapse to the same vector. We therefore use the pretrained LangFlow token embeddings and keep them fixed during training with the diffusion loss. 

Because jointly learning the token embeddings and data predictor under the MSE objective admits degenerate embedding-collapse solutions, we use the embedding matrix from the LangFlow checkpoint and keep it fixed throughout training. 
For all experiments except the convergence study, the remaining trainable parameters are also initialized from the same checkpoint.

% Starting from the checkpoint of LangFlow, we continue to train our model with the MSE loss and LangFlow with the CE loss for 200K iterations on four or eight NVIDIA A100 $40$GB GPUs depending on availability.
% We use the AdamW optimizer with a global batch size of $480$ and a learning rate of $10^{-5}$.
For a controlled comparison, we continue training our model using the MSE objective and the LangFlow baseline using its token-level CE objective for additional 200K steps. Both models are trained using AdamW \citep{loshchilov2017decoupled} with a global batch size of $480$ and a learning rate of $10^{-5}$. Training is distributed across four or eight NVIDIA A100 40 GB GPUs, depending on availability.

\paragraph{Evaluation.}
For each sampling configuration, we generate $1024$ samples with length $L=1024$.
We measure generation quality using Gen.~PPL, evaluated by GPT-2 Large \citep{radford2019language}, and quantify diversity using unigram entropy. We compare sampling methods based on their Gen.~PPL--entropy trade-off, where lower Gen.~PPL indicates higher quality and higher entropy indicates greater diversity.

We observe that the sampling grid used by LangFlow is highly nonuniform near the two endpoints. Therefore, we use the uniform grid $t_i=(i+0.5)/N$ for $i=0,1,\ldots,N-1$, where $N$ denotes the number of sampling steps. 
This grid slightly outperforms the original one in LangFlow; see Figure~\ref{fig:schedule} in Appendix~\ref{sec:appendix_exp}. 
% {\footnote{For example, with NFE $N=32$, the SNR jumps from $1.08\times10^{-7}$ to $2.61\times10^{-4}$ over the first interval and from $2.31\times10^{-2}$ to $7.42\times10^{-2}$ over the final interval, whereas it changes much more smoothly across the intermediate steps. These abrupt boundary changes can introduce additional discretization error.}}

Unless otherwise specified, our default sampling configuration is the standard sampler with one-step self-conditioning, as defined in \eqref{eq:data_prediction_update} and \eqref{eq:self_conditioning_practical}, without additional techniques introduced in Section~\ref{sec:sampling}; see Appendix~\ref{subsec:samples} for generated examples at an entropy of 5.44.
This corresponds to $w_{\scg}=1$, $w_{\mathsf{ug}}=0$, and $K=1$. 
We compare all sampling methods under the same number of function evaluations (NFEs). Every evaluation of the data predictor is counted, including additional evaluations introduced by self-conditioning guidance or iterative self-conditioning refinement.

% The results reported below are obtained after 175K iterations of MSE-based training initialized from the pre-trained LangFlow checkpoint.
Unless otherwise stated, the results reported below use the checkpoint obtained after 175K additional training steps, which achieved the best performance among the evaluated checkpoints.

\subsection{Empirical flow convergence to token embeddings}

\begin{figure}[t]
    \centering
    \begin{subfigure}[b]{0.48\textwidth}
        \centering
        \includegraphics[width=\textwidth]{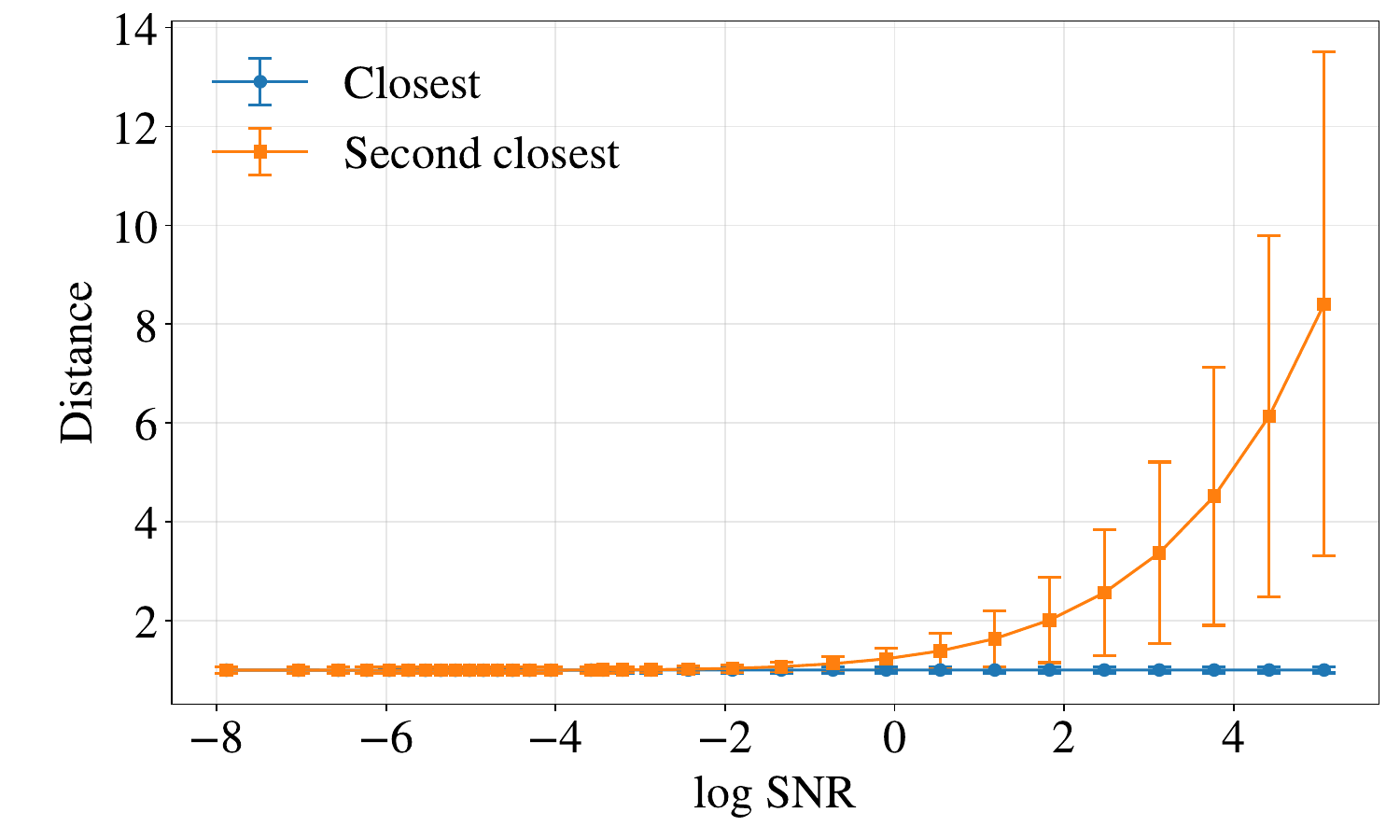}
        \caption{Embedding-weighted data predictor}
        \label{fig:distance_dist_logits}
    \end{subfigure}
    \hfill
    \begin{subfigure}[b]{0.48\textwidth}
        \centering
        \includegraphics[width=\textwidth]{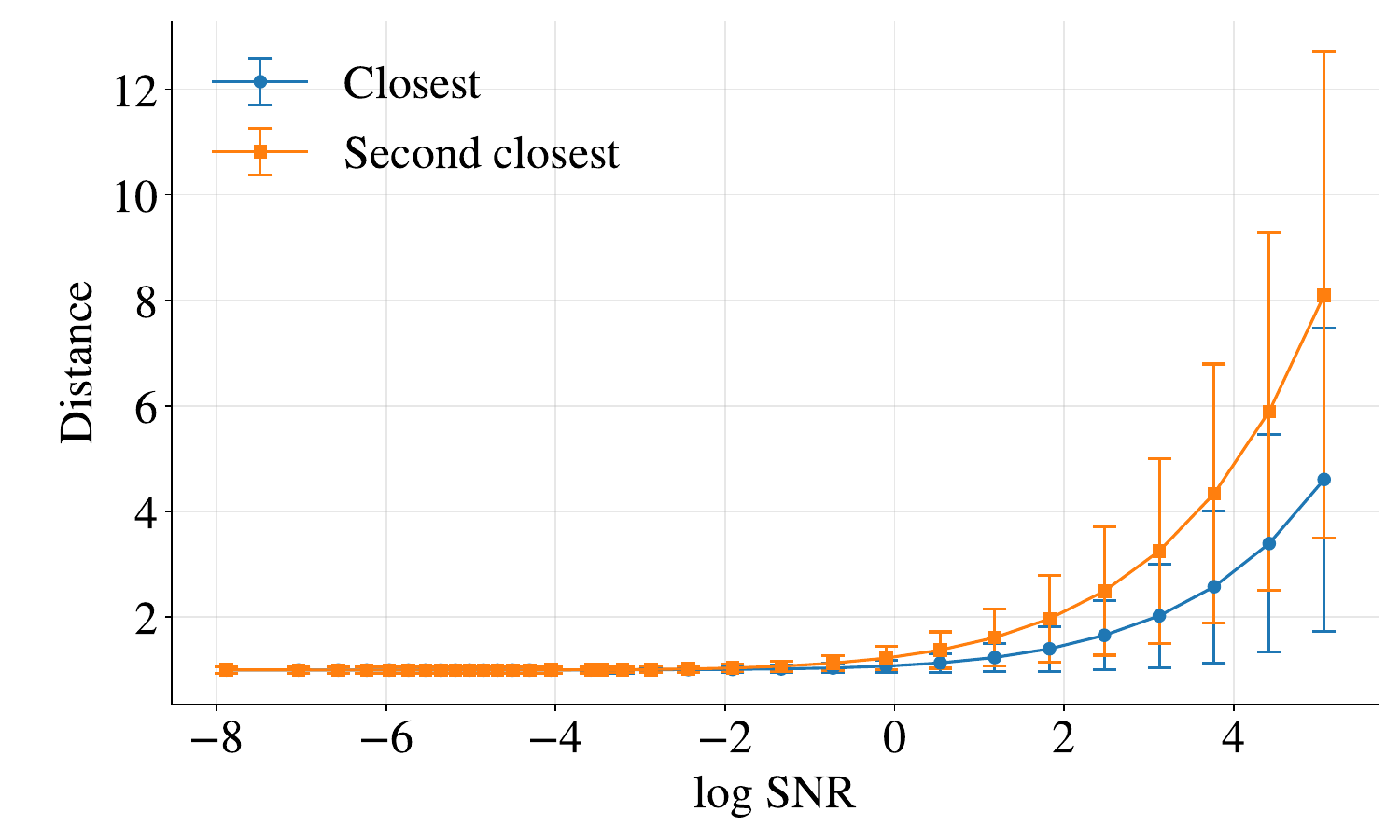}
        \caption{Unconstrained direct data predictor}
        \label{fig:distance_dist_x0pred}
    \end{subfigure}
    
    \caption{Comparison of flow convergence between the embedding-weighted and direct data predictors. The two models use the same fixed LangFlow embeddings, architecture, and training configuration, differing only in their output parameterization. We plot the smallest and second-smallest normalized distances $\|x_t^{(i)}-\alpha_t e_j\|_2/(\sigma_t\sqrt{d})$ from the flow state associated with each token position \(x_t^{(i)}\) to the scaled token embeddings as functions of log-SNR.}
    \label{fig:distance_dist}
\end{figure}

We empirically examine how the parameterization of the data predictor affects the convergence of the learned flow. 
We train two models from random initialization using the same pre-trained LangFlow embedding matrix, network architecture, and training configuration.
The models differ only in their output parameterization. 
The embedding-weighted data predictor expresses its output as a weighted combination of token embeddings, as defined in \eqref{eq:posterior_structured_data_predictor}, whereas the direct data predictor outputs an unconstrained vector in the embedding space.
This controlled comparison isolates the effect of the structured parameterization: Theorem~\ref{thm:continuous_flow} guarantees the learned flow driven by the embedding-weighted data predictor converges to valid token embeddings, while Proposition~\ref{prop:counterexample} shows that such a convergence guarantee may not hold for an unconstrained data predictor.
% This controlled comparison contrasts the convergence guarantee in Theorem~\ref{thm:continuous_flow} with the lack of such a guarantee for unconstrained predictors demonstrated by Proposition~\ref{prop:counterexample}.
% Specifically, for each $j\in[V]$, we compute the normalized distance $\|x_t-\alpha_t e_j\|_2/(\sigma_t\sqrt{d})$.
%  and evaluated across $\log \mathrm{SNR}$.
% We consider the empirical mean and standard deviation over 16,384 samples, for both the logits and direct predictors.
% \cc{is the plot of the mean of all tokens or a specific token? make it concise}\LN{Mean of all tokens, modified the last sentance of this paragraph}

For each token position $i\in[L]$, we compute the normalized distance between its continuous state $x_t^{(i)}$ and every scaled token embedding $\alpha_t e_j$:
\begin{align*}
\frac{\|x_t^{(i)}-\alpha_t e_j\|_2}{\sigma_t\sqrt d}, \qquad j\in[V].
\end{align*}
% For each log-SNR value, we compute the smallest and second-smallest distances at every token position and report the midpoint of each empirical 1st--99th percentile range, with error bars spanning the corresponding range.
% The statistics are computed over all $16 \times 1024 = 16{,}384$ token positions from 16 generated sequences of length $L=1024$ for both predictors.
For each log-SNR value, we aggregate the smallest and second-smallest distances over all $16\times1024=16{,}384$ token positions from 16 generated sequences of length $L=1024$, separately for each data predictor. For each distance statistic, we plot the midpoint of its empirical 1st--99th percentile interval, with error bars spanning the full interval.

% For each log-SNR value, we report the empirical mean and two times standard deviation of the smallest and second-smallest distances over all $16\times1024=16{,}384$ token positions from 16 generated sequences of length $L=1024$, for both the embedding-weighted and direct data predictors.

% We also plot $\max w(x_{\star}\mymid x_t, t)$.

As shown in Figure~\ref{fig:distance_dist}, the closest and second-closest distances are nearly indistinguishable in the low-SNR regime for both data predictors, but their behaviors diverge as the SNR increases. 
For the embedding-weighted data predictor in Figure~\ref{fig:distance_dist_logits}, the closest normalized distance approaches one and remains tightly concentrated, whereas the second-closest distance increases rapidly, producing a clear separation. 
In contrast, for the unconstrained direct data predictor in Figure~\ref{fig:distance_dist_x0pred}, both distances increase, exhibit substantially greater variation, and remain poorly separated.
These empirical results validate the contrasting convergence behaviors characterized by Theorem~\ref{thm:continuous_flow} and Proposition~\ref{prop:counterexample}.
% As the SNR increases, the nearest distance of the logits predictor in Figure~\ref{fig:distance_dist_logits} remains near one with little variation while the second-nearest distance grows rapidly.
% In contrast, both distances for the direct predictor increase and exhibit greater variation, although the nearest embedding remains consistently nearer, which is illustrated in Figure~\ref{fig:distance_dist_x0pred}.

Having empirically examined convergence toward token embeddings, we next examine the consistency of the weight-based and distance-based token prediction rules.
Theorem~\ref{thm:continuous_flow} and Proposition~\ref{prop:dis-weight-converge} together imply their asymptotic equivalence: the flow state converges to a token embedding, while the corresponding weights converge to its one-hot representation.
As the number of sampling steps $N$ varies from $32$ to $512$, the two rules agree at $99.16\%$--$99.82\%$ of token positions, with the agreement increasing as the sampling discretization becomes finer. This near-perfect agreement indicates that the two rules are effectively equivalent in practice, with the remaining discrepancies diminishing as the sampling time discretization becomes finer.

% \begin{figure}[!htbp]
%     \centering
%     \includegraphics[width=0.52\textwidth]{figure/token_diff.pdf}
%     \caption{Token-level agreement between the logits-based and distance-based output strategies as a function of the number of sampling steps $N$. The matching probability is the fraction of token positions at which the two strategies select the same token.}
%     \label{fig:token_match}
% \end{figure}

% \subsection{Influence of sampler schedule}

\subsection{Effect of the training objective: MSE versus CE}
Starting from the pre-trained LangFlow checkpoint, we continue training two models for 200K iterations under identical configurations, using the MSE and CE objectives, respectively.
We evaluate Gen.~PPL and entropy every 25K steps. 

The entropy remains above $5.5$ for both objectives throughout training.
As for Gen.~PPL, Figure~\ref{fig:ce_mse_from_langflow} shows that training with the CE objective provides no consistent improvement over the initial checkpoint and its Gen.~PPL remains near its initial  value. In contrast, training with the MSE objective steadily reduces Gen.~PPL and maintains a clear advantage throughout training, despite some fluctuations across checkpoints.

% To further evaluate the effect of the training objectives, we switch the objectives at 100k iterations. Specifically, the training checkpoint of CE objective is subsequently optimized with MSE objective, while the better-performing checkpoint by MSE objective is optimized with CE objective.
To further isolate the effect of the training objective, we conduct a crossover experiment using the checkpoints obtained after 100K steps. Specifically, we continue the CE-trained checkpoint using the MSE objective and, conversely, continue the MSE-trained checkpoint using the CE objective.
% As shown in Figure~\ref{fig:ce_mse_switch}, switching the objective from CE to MSE substantially reduces Gen.~PPL, whereas switching it from MSE to CE degrades performance despite starting from the better checkpoint. This crossover experiment confirms that the improvement is attributable to the MSE objective and that training with CE objective can undo gains previously obtained with MSE.
As shown in Figure~\ref{fig:ce_mse_switch}, despite starting from the worse-performing CE-trained checkpoint, switching to MSE reduces Gen.~PPL consistently. Conversely, switching the better-performing MSE-trained checkpoint to CE degrades Gen.~PPL. 
This crossover experiment confirms that the improvement is attributable to the MSE objective rather than favorable initialization. It further demonstrates that subsequent CE-based training can reverse gains previously obtained through MSE-based training.

\begin{figure}[t]
    \centering
    \begin{subfigure}[b]{0.48\textwidth}
        \centering
        \includegraphics[width=\textwidth]{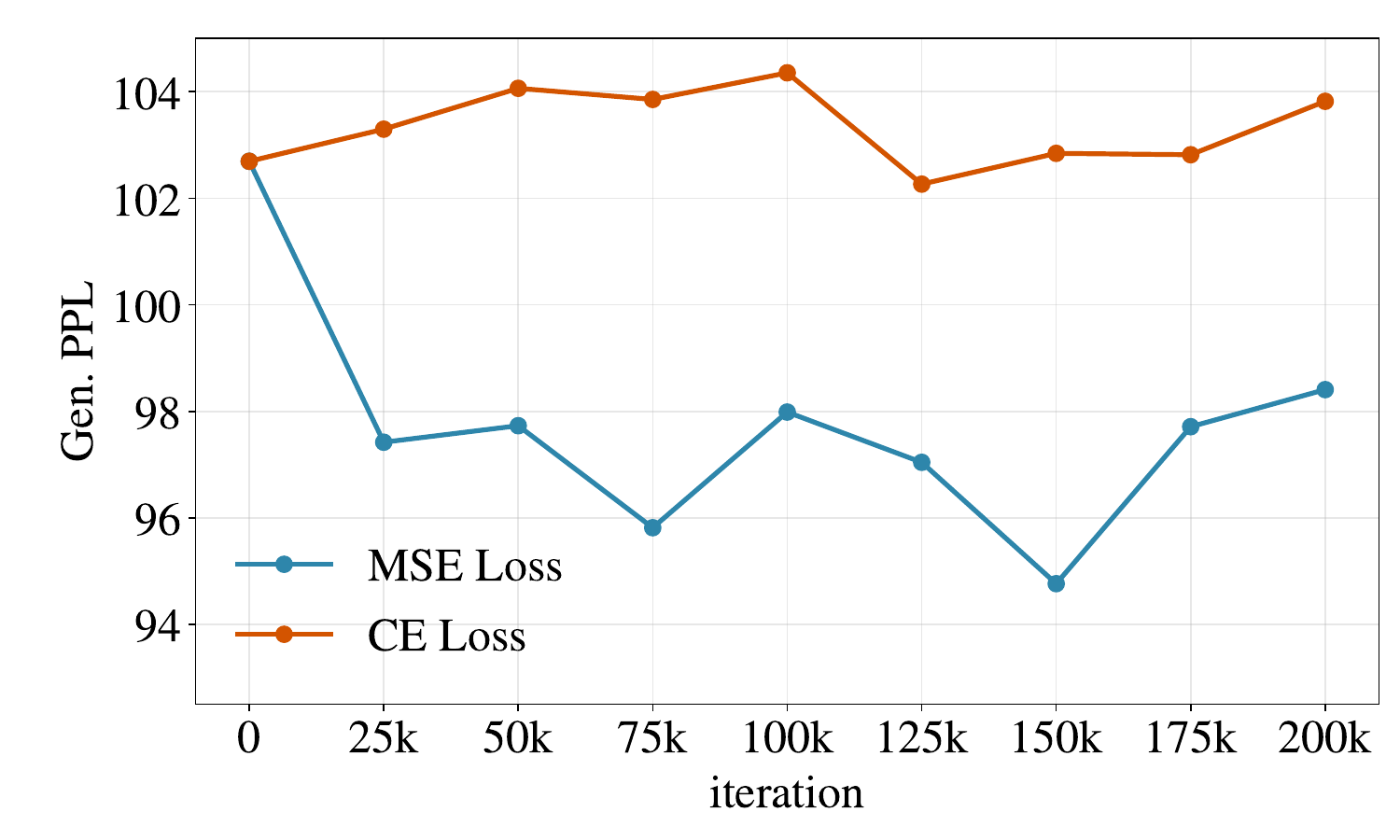}
        \caption{Continued training from same LangFlow checkpoint}
        \label{fig:ce_mse_from_langflow}
    \end{subfigure}
    \hfill
    \begin{subfigure}[b]{0.48\textwidth}
        \centering
        \includegraphics[width=\textwidth]{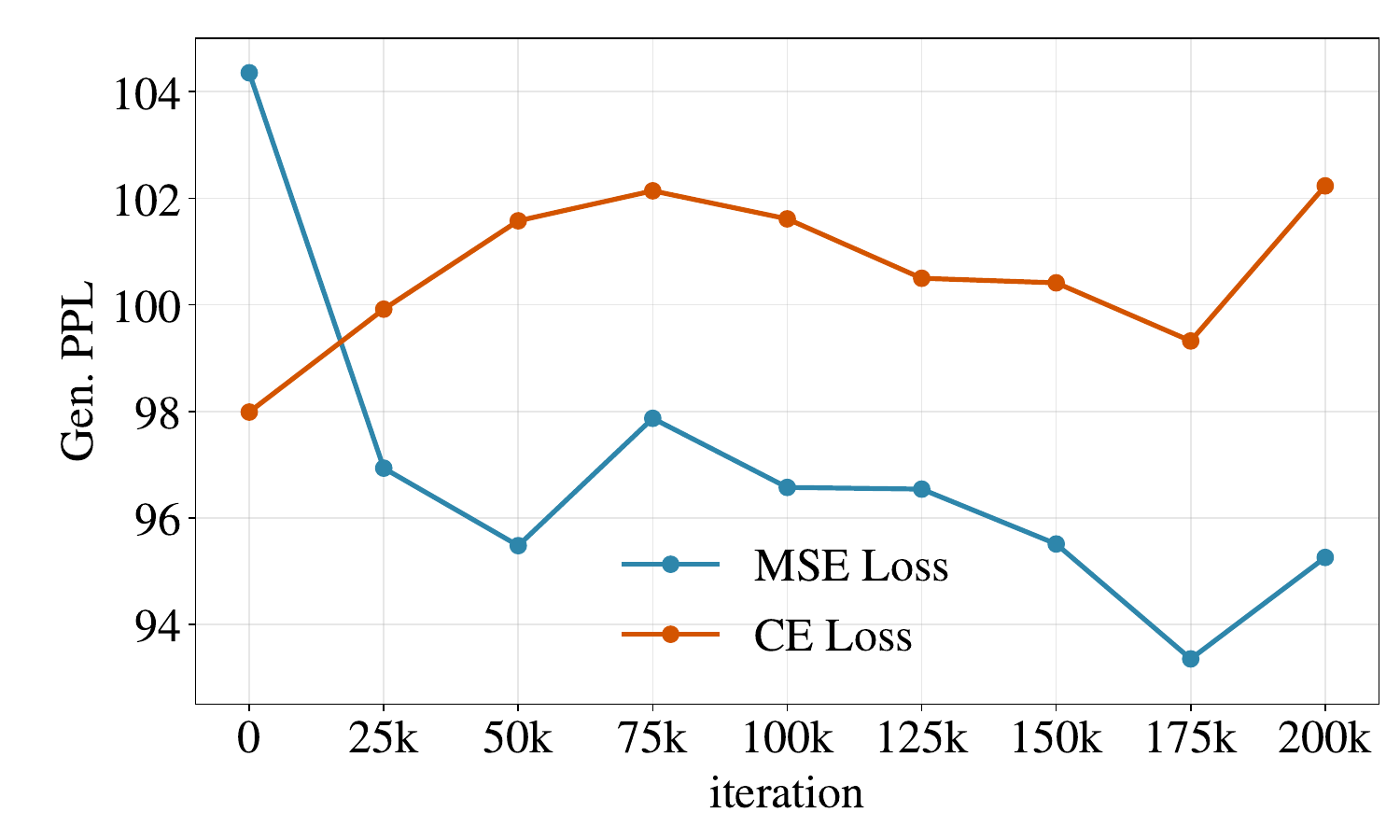}
        \caption{Swapping objectives after 100K training steps}
        \label{fig:ce_mse_switch}
    \end{subfigure}
    
    \caption{Effect of the training objective on Gen.~PPL. (a) Starting from the same pre-trained LangFlow checkpoint, MSE-based training yields an immediate and persistent improvement over CE-based training.
    (b) In the crossover experiment, the CE-trained checkpoint improves rapidly after switching to MSE, whereas the MSE-trained checkpoint degrades after switching to CE. The horizontal axis denotes additional training steps after the objective switch.}
    \label{fig:ce_mse}
\end{figure}

\subsection{Control of quality-diversity trade-off}\label{subsec:trade-off}
Empirically, we find that sampling guidance is most effective near the data endpoint. We therefore adopt time-adaptive variants of self-conditioning guidance, unconditional guidance, and iterative self-conditioning refinement.

Specifically, at each step $i$, we replace the constant guidance strengths $w_{\scg}$ and $w_{\mathsf{ug}}$ with $w_{\scg}/(1+\sigma_{t_i}/\alpha_{t_i})$ and $w_{\mathsf{ug}}/(1+\sigma_{t_i}/\alpha_{t_i})$, respectively. 
Thus, these schedules gradually increase the guidance strength over the sampling trajectory.

For iterative self-conditioning refinement, we analogously adapt the refinement count according to
\begin{align*}
K_i = \Big\lceil\frac{K_{\iscr}}{1+\sigma_{t_i}/\alpha_{t_i}}\Big\rceil.
\end{align*}
To ensure a fair comparison under a fixed computational budget, we choose the number of sampling steps separately for each configuration. In particular, the NFE for iterative self-conditioning refinement is given by
\begin{align*}
\mathsf{NFE} = \sum_{i = 0}^{N-1} (K_i + 1).
\end{align*}

\begin{figure}[t]
    \centering
    \begin{subfigure}[t]{0.48\textwidth}
        \centering
        \includegraphics[width=\textwidth]{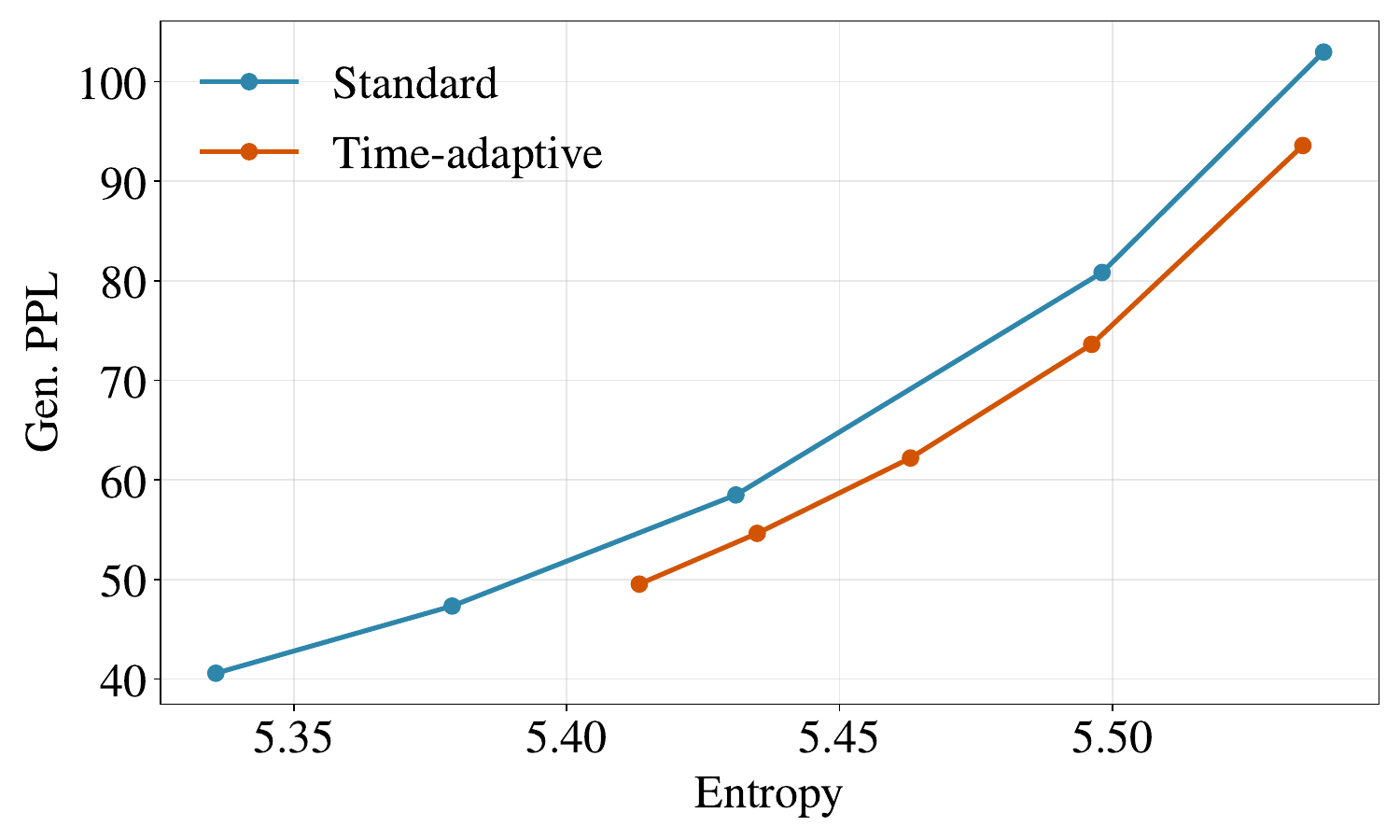}
        \caption{Standard and time-adaptive self-conditioning guidance.}
        \label{fig:trade-off_cfg}
    \end{subfigure}
    \hfill
    \begin{subfigure}[t]{0.48\textwidth}
        \centering
        \includegraphics[width=\textwidth]{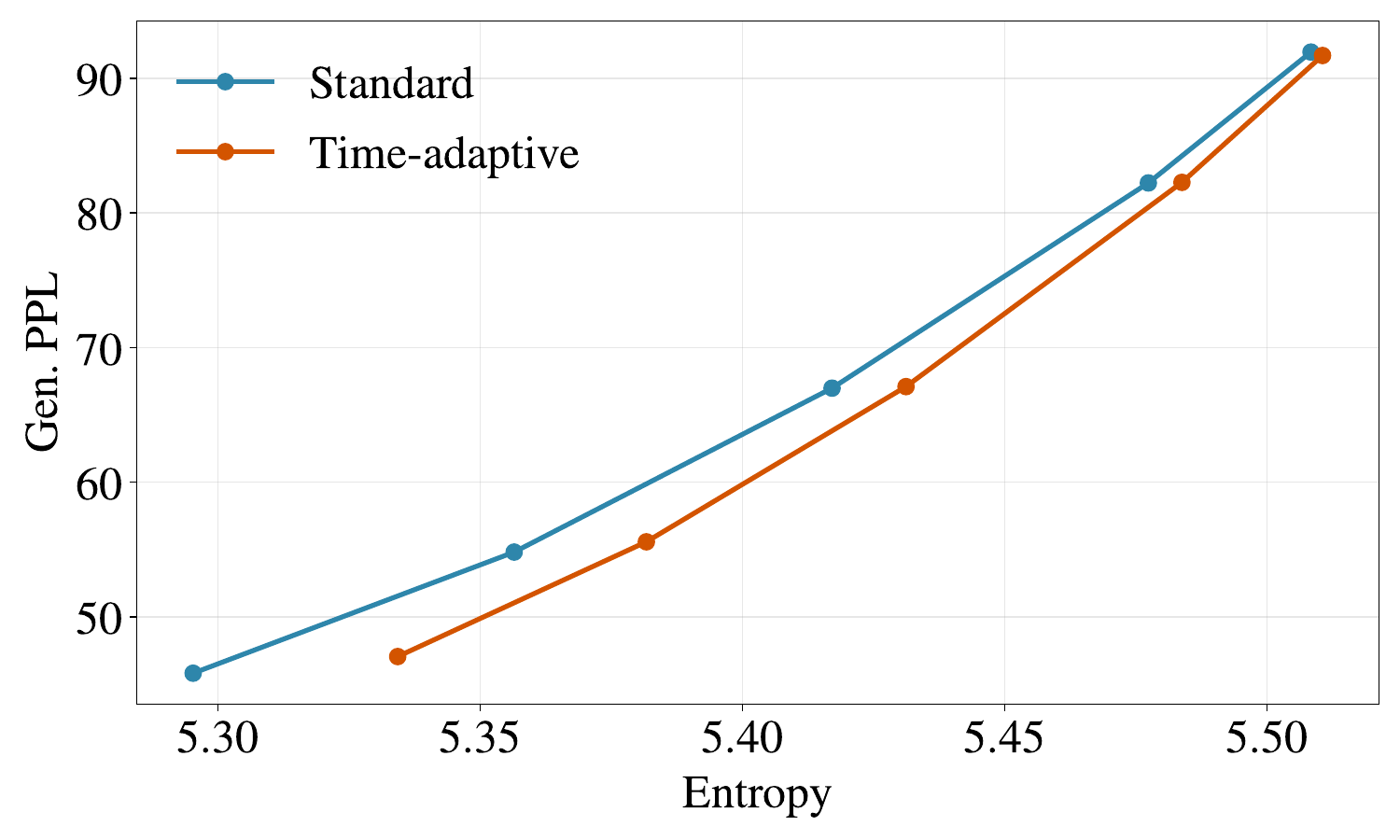}
        \caption{Standard and time-adaptive unconditional guidance.}
        \label{fig:trade-off_uncond}
    \end{subfigure}
    \hfill
    \begin{subfigure}[t]{0.48\textwidth}
        \centering
        \includegraphics[width=\textwidth]{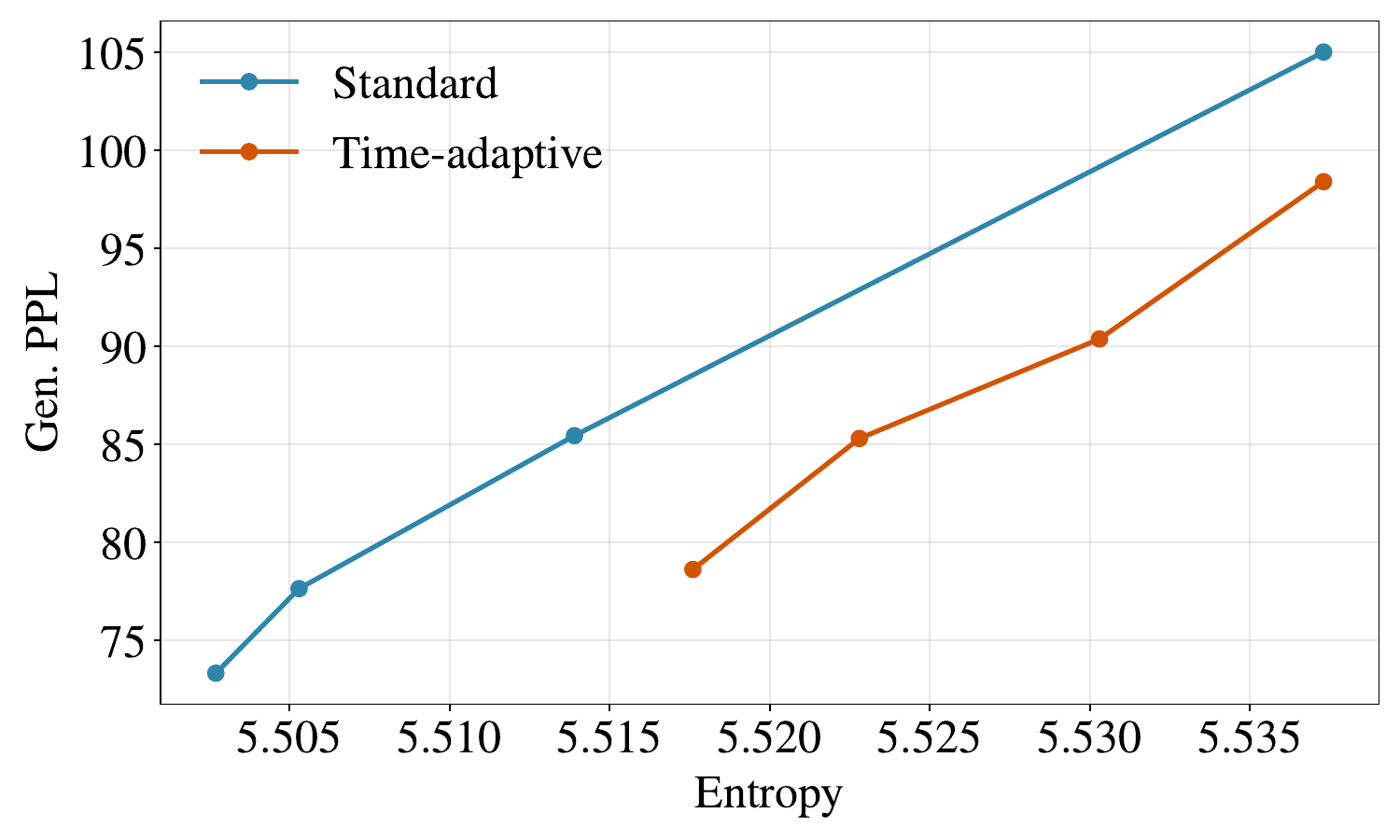}
        \caption{Standard and time-adaptive iterative self-conditioning refinement.}
        \label{fig:trade-off_selfcond}
    \end{subfigure}
    \hfill
    \begin{subfigure}[t]{0.48\textwidth}
        \centering
        \includegraphics[width=\textwidth]{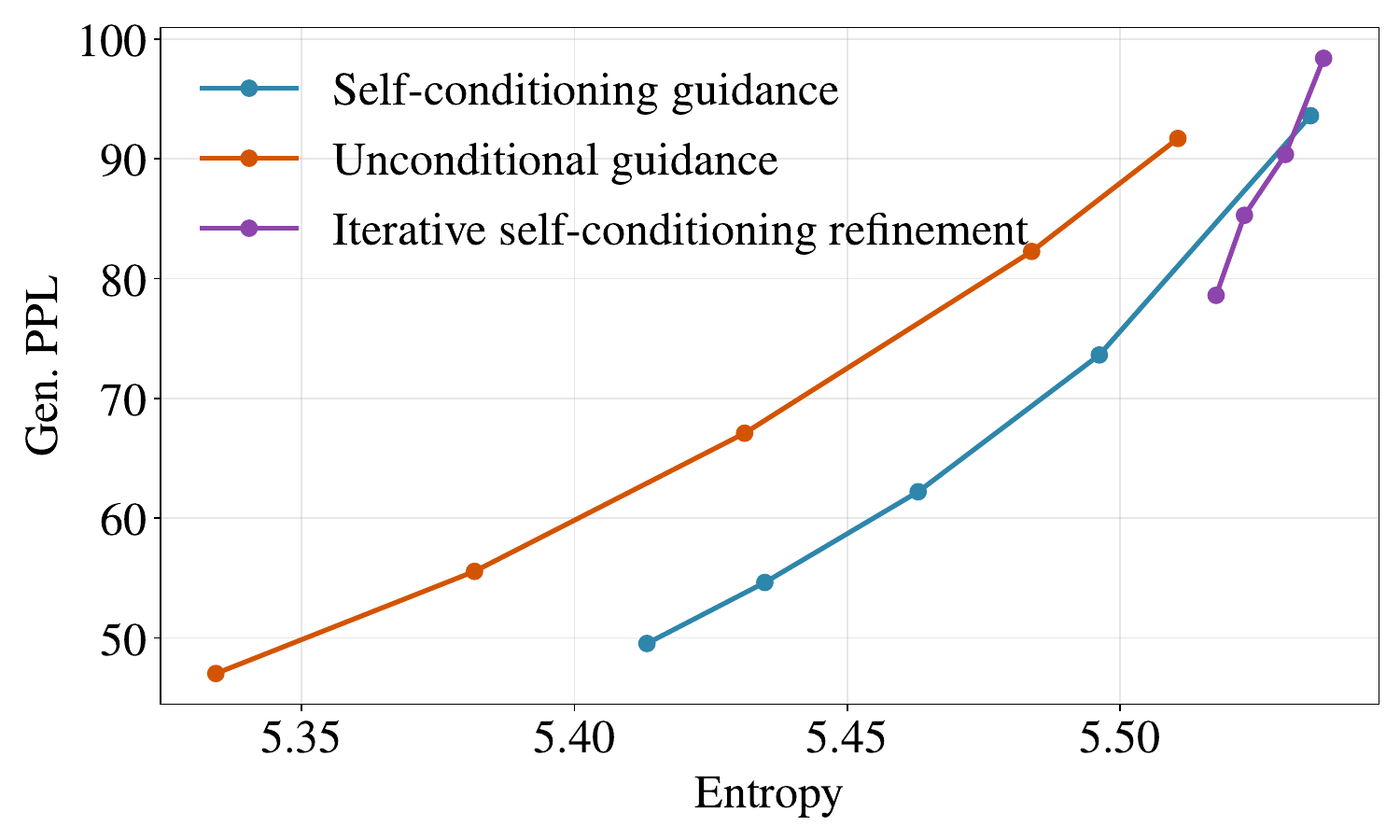}
        \caption{Comparison of the three time-adaptive samplers.}
        \label{fig:trade-off_adaptive}
    \end{subfigure}
    \caption{Gen.~PPL-entropy trade-offs of the three sampling techniques at NFE=$64$. Each curve is obtained by varying the corresponding nominal control parameter.  (a)--(c) compare the standard and time-adaptive variants of each sampling technique, where the adaptive variants increase the guidance strength or self-conditioning refinement count toward the data endpoint. (d) compares the time-adaptive variants of the three sampling techniques.}
    \label{fig:trade-off}
\end{figure}

All three sampling mechanisms provide effective control over the quality-diversity trade-off. 
Figures~\ref{fig:trade-off_cfg}--\ref{fig:trade-off_selfcond} present the results for self-conditioning guidance, unconditional guidance, and iterative self-conditioning refinement, respectively. Figure~\ref{fig:trade-off_adaptive} compares their time-adaptive variants. 
Each trade-off curve is obtained by varying the corresponding nominal control parameter, namely $w_{\scg}$, $w_{\mathsf{ug}}$, or $K_{\iscr}$, while holding the total NFE fixed.
The complete results for NFE=$64$ and  NFE=$128$ are summarized in Tables~\ref{tab:trade-off_64} and~\ref{tab:trade-off_128} in Appendix~\ref{subsec:trade-off_appendix}, respectively.

Overall, the time-adaptive variants generally improve the Gen.~PPL-entropy frontier relative to their constant-strength counterparts.
% by concentrating additional guidance near the data endpoint. 
% For self-conditioning guidance, the adaptive guidance $w_{\scg}=80$ achieves a PPL of $54.64$ and an entropy of $5.4349$, improving upon the standard guidance $w_{\scg}=4$, which gives a higher PPL of $58.50$ and a lower entropy of $5.4310$. 
% A similar pattern holds for unconditional guidance and iterative self-conditioning refinement. 
% Overall, adaptive scheduling yields a more favorable Gen.~PPL-entropy balance than the corresponding standard controls.
As shown in Figure~\ref{fig:trade-off_adaptive}, among the three adaptive techniques, self-conditioning guidance spans the broadest range of operating points and achieves the most favorable trade-off in the low- and medium-entropy regimes.
Iterative self-conditioning refinement is particularly effective in the high-entropy regime, although it covers a narrower controllable range.
% Nevertheless, adaptive guidance yields only modest Gen.~PPL reductions, which is investigated in Section~\ref{subsec:fur_imp}.
Unconditional guidance yields only modest Gen.~PPL reductions, and we investigate this further in Section~\ref{subsec:fur_imp}.

\subsection{Combination of three sampling techniques}\label{subsec:comb}
We next test the performance of combinations of the three sampling techniques, where we use the time-adaptive variant in all cases.

\begin{figure}[t]
    \centering
    \includegraphics[width=0.52\textwidth]{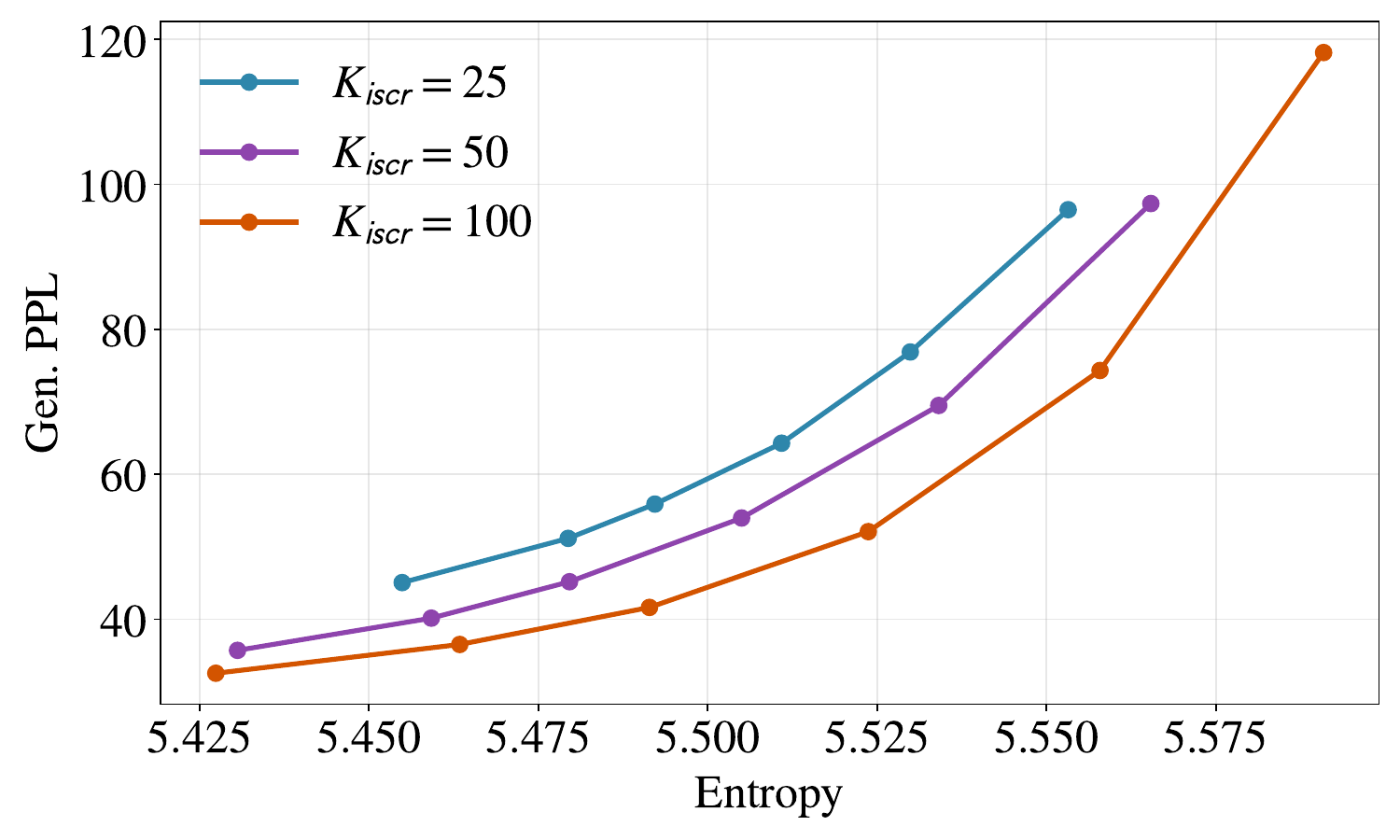}
    \caption{Joint effect of iterative self-conditioning refinement and self-conditioning guidance at NFE=$64$. Each curve fixes the refinement parameter $K_{\iscr}$ and varies the guidance strength $w_{\scg}$.}
    \label{fig:wk_wcfg}
\end{figure}

We first combine self-conditioning guidance and iterative self-conditioning refinement.
Figure~\ref{fig:wk_wcfg} reports the results under a fixed NFE budget of $64$. The complete results for NFE=$64$ and NFE=$128$ are provided in Tables~\ref{tab:wk_wcfg_64} and \ref{tab:wk_wcfg_128} in Appendix~\ref{subsec:comb_appendix}, respectively.

As illustrated in Figure~\ref{fig:wk_wcfg}, increasing $K_{\iscr}$ generally shifts the trade-off frontier toward lower Gen.~PPL at comparable entropy levels.
Thus, combining a sufficiently large self-conditioning refinement parameter $K_{\iscr}$ with an appropriate self-conditioning guidance strength $w_{\scg}$ substantially improves the trade-off.

\begin{figure}[t]
    \centering
    \begin{subfigure}[b]{0.48\textwidth}
        \centering
        \includegraphics[width=\textwidth]{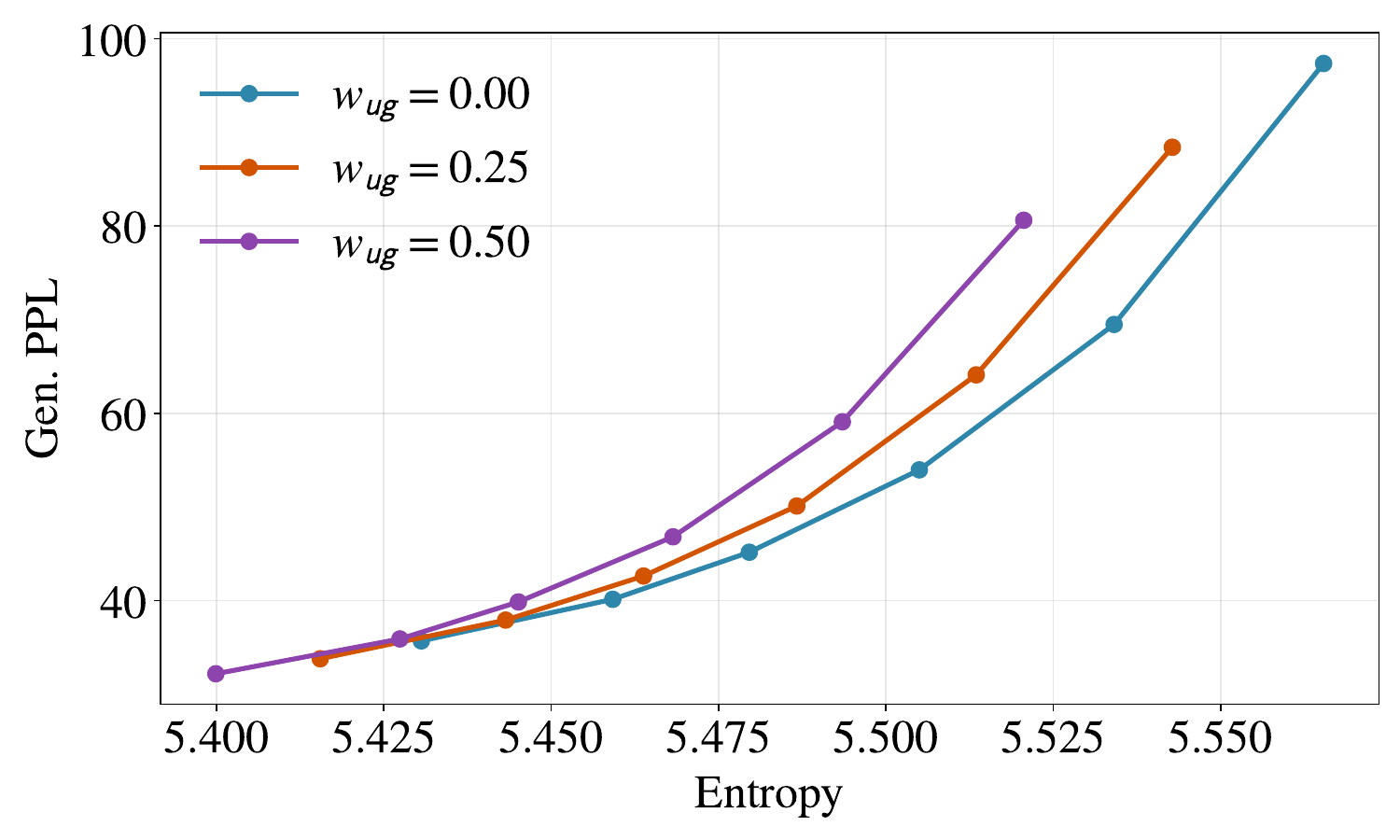}
        \caption{$K_{\iscr}$=50}
        \label{fig:wug_64_wk50}
    \end{subfigure}
    \hfill
    \begin{subfigure}[b]{0.48\textwidth}
        \centering
        \includegraphics[width=\textwidth]{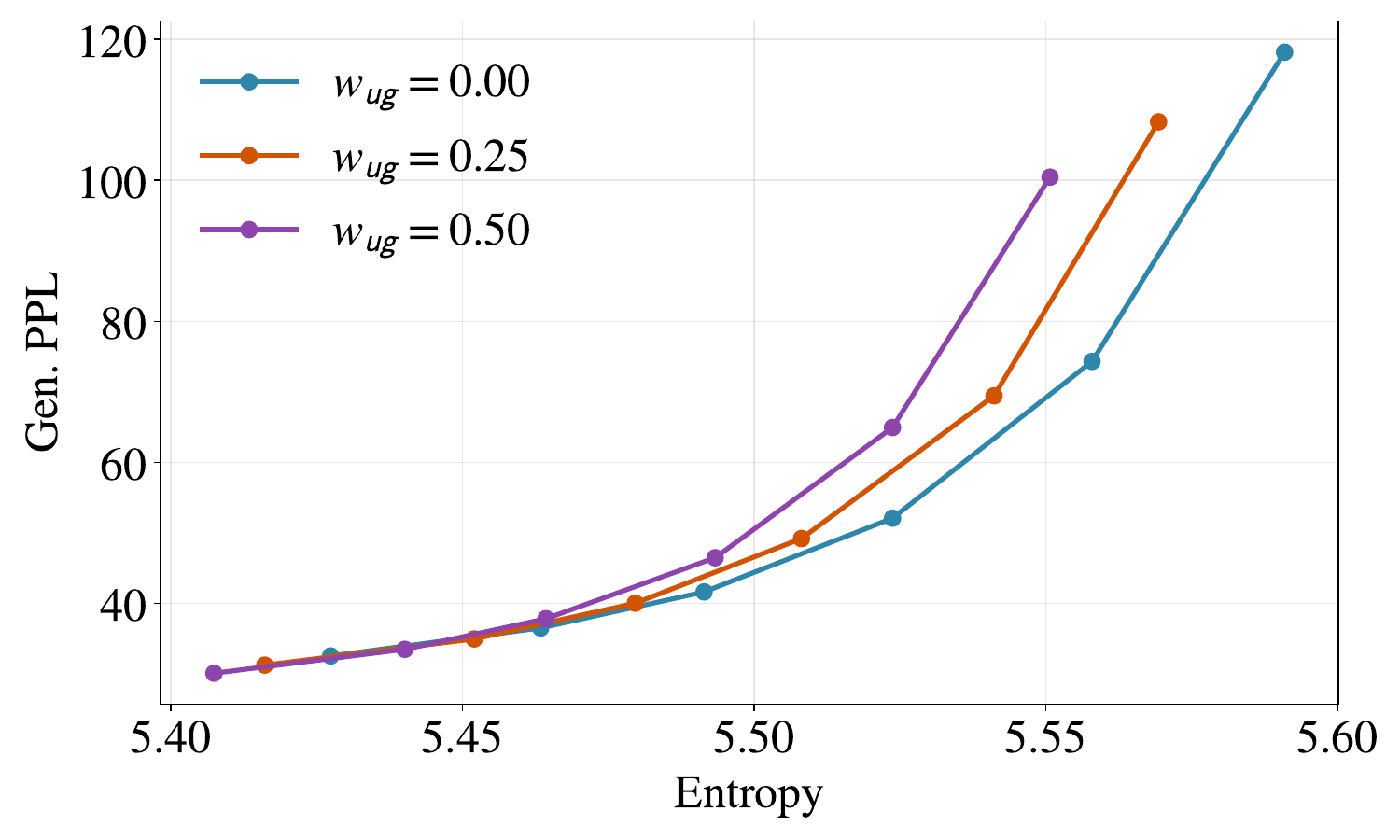}
        \caption{$K_{\iscr}$=100}
        \label{fig:wug_64_wk100}
    \end{subfigure}
    
    \caption{Effect of adding unconditional guidance to a sampler combining iterative self-conditioning refinement and self-conditioning guidance at $\mathrm{NFE}=64$.}
    \label{fig:wug_64}
\end{figure}

We next add time-adaptive unconditional guidance to the combined sampler. 
Figure~\ref{fig:wug_64} plots the results, with \(K_{\iscr}=50\) and \(K_{\iscr}=100\) shown separately in Figures~\ref{fig:wug_64_wk50} and \ref{fig:wug_64_wk100}, respectively.
For both settings, varying $w_{\mathsf{ug}}$ primarily moves the operating point along a similar Gen.~PPL-entropy frontier rather than shifting the frontier outward.
These results suggest that self-conditioning guidance and iterative self-conditioning refinement capture most of the gain.
Nevertheless, unconditional guidance remains useful because it does not require a conditioning signal, which we believe has independent interest.
The complete numerical results for different combinations of $w_{\mathsf{ug}}$, $K_{\iscr}$, and $w_{\scg}$ are reported in Table~\ref{tab:wug_wk_wcfg_nfe64} in Appendix~\ref{subsec:comb_appendix}.

\begin{figure}[t]
    \centering
    \begin{subfigure}[]{0.48\textwidth}
        \centering
        \includegraphics[width=\textwidth]{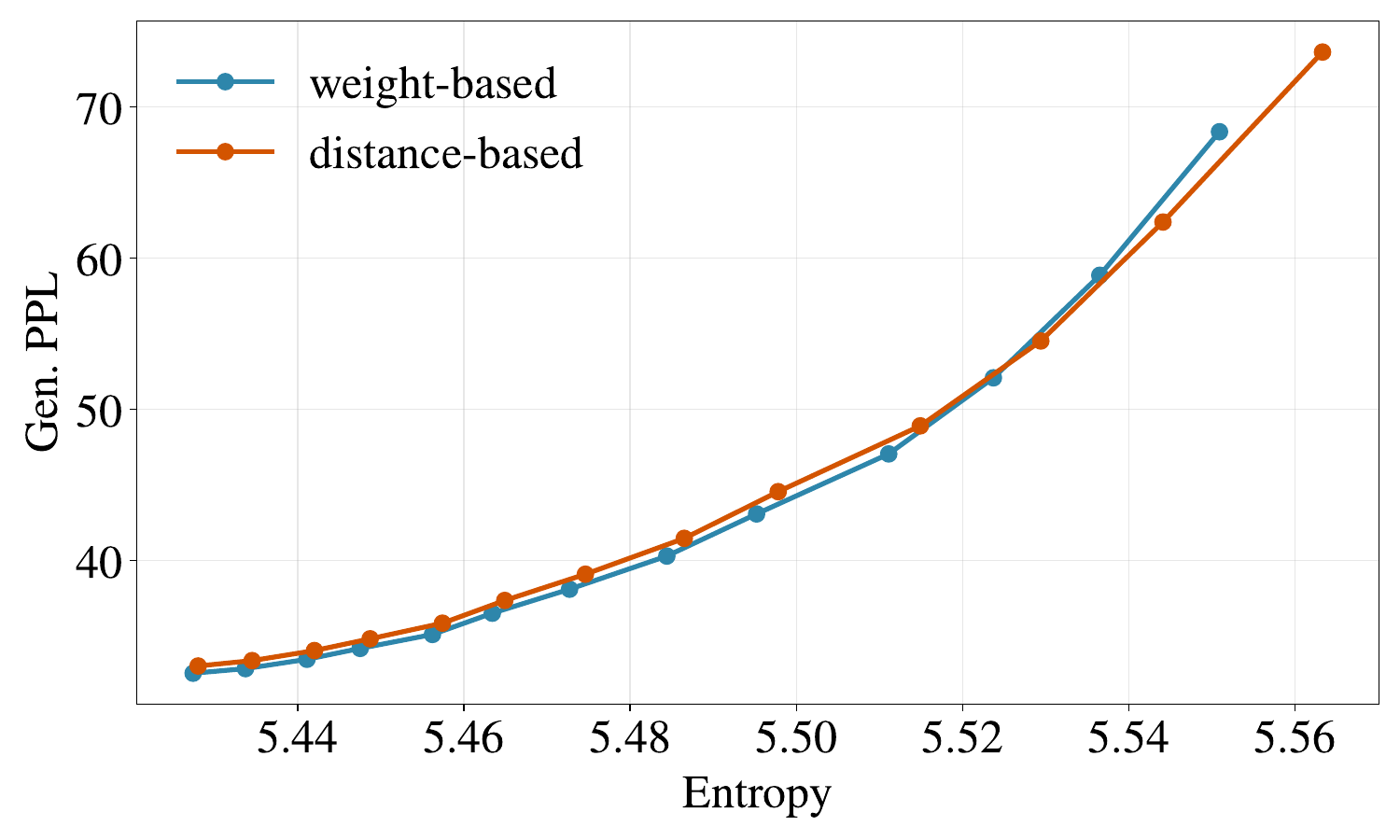}
        \caption{NFE=64}
        \label{fig:output_64}
    \end{subfigure}
    \hfill
    \begin{subfigure}[]{0.48\textwidth}
        \centering
        \includegraphics[width=\textwidth]{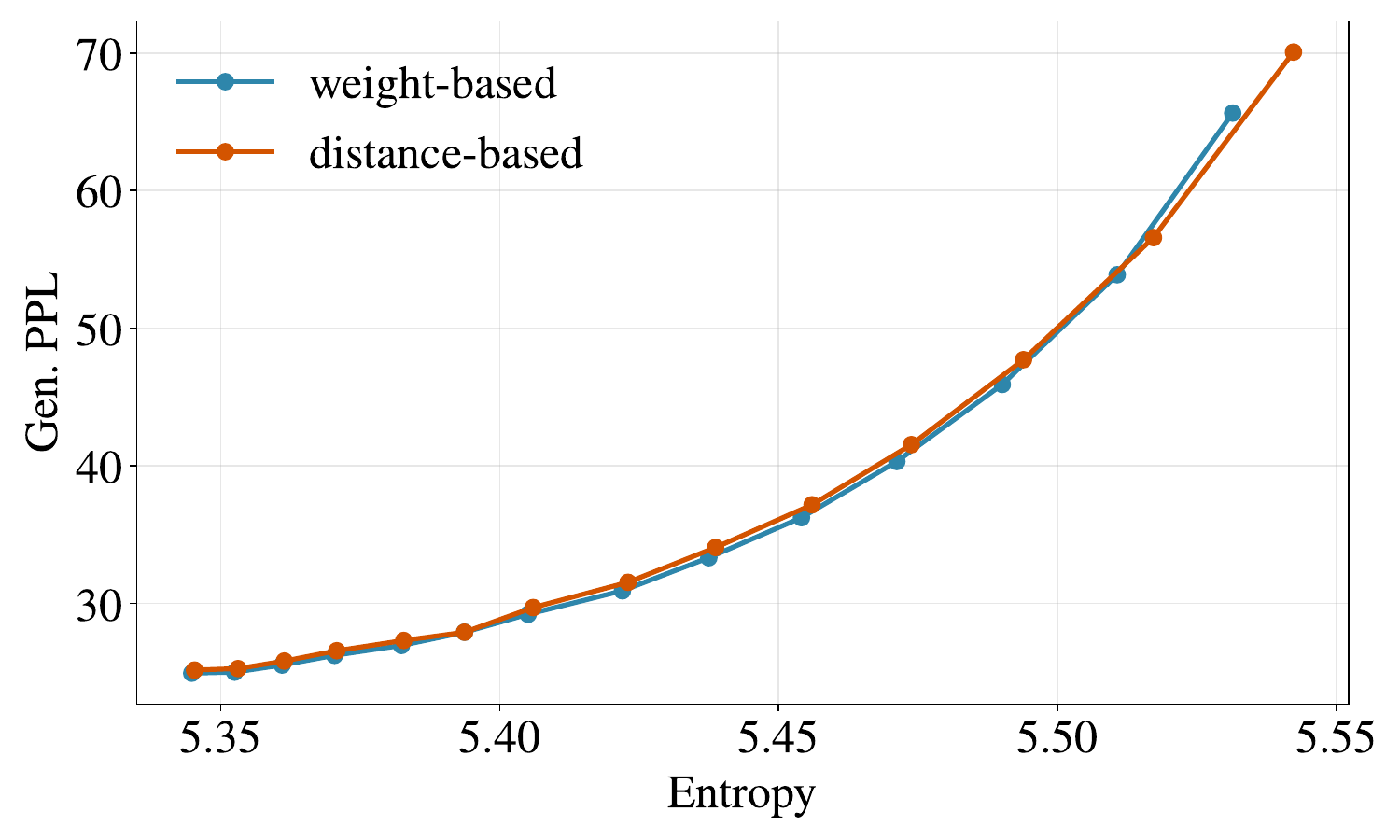}
        \caption{NFE=128}
        \label{fig:output_128}
    \end{subfigure}
    \caption{Comparison of two token prediction rules for NFE budgets of $64$ and $128$.}
    \label{fig:output}
\end{figure}

Finally, we compare the two token prediction rules described in Section~\ref{sec:sampling}. 
% Since each clean endpoint $x_\star^{(i)}$ is a token embedding, $\wh s^{(i)} = \argmin_{j\in[V]} \|x_{t_N}^{(i)} - e_j\|_2, \quad i\in[L]$, provides a natural estimate of $x_\star^{(i)}$. Moreover, as the posterior-structured predictor approaches the data endpoint, 
As the flow approaches the data endpoint, these two rules should output the same token. Figure~\ref{fig:output} empirically verifies this agreement, with the NFE\(=64\) and NFE\(=128\) results shown in Figures~\ref{fig:output_64} and \ref{fig:output_128}, respectively. The corresponding Gen.~PPL-entropy curves nearly overlap under both computational budgets. In the high-entropy, high-Gen.~PPL regime, distance-based decoding yields a slightly more favorable trade-off. These results support using nearest-neighbor projection to map the generated continuous flow state to a token sequence, without a separately trained terminal decoder.

% the distance-based and logit-based strategies are expected to agree, as shown in Figure~\ref{fig:output} in Appendix~\ref{subsec:comb_appendix}; its NFE=$64$ and $128$ panels are given in Figures~\ref{fig:output_64} and \ref{fig:output_128}, respectively. The Gen.~PPL-entropy curves nearly overlap under both NFEs. In the high-entropy and high-Gen.~PPL regime, the distance-based strategy even yields a slightly better trade-off. This agreement empirically supports using the nearest-token projection of the generated continuous state as an estimate of $x_\star$.

\subsection{Further improvement}\label{subsec:fur_imp}
The experiments in Section \ref{subsec:trade-off} indicate that concentrating sampling guidance near the data endpoint improves the Gen.~PPL-entropy trade-off. At sufficiently large guidance strengths, however, data endpoint-focused allocation exhibits diminishing returns. Further reduction in Gen.~PPL requires stronger guidance earlier in the trajectory, when the state remains relatively noisy.

To study this effect, we compare the following three configurations under the same NFE budget: 
\begin{itemize}
    \item {Configuration A}: fixed refinement count $K_i = K$ and $w_{\scg}(t_i)=w_{\scg}$; 
    \item {Configuration B}: $K_i = \lceil K_{\iscr}/(1+\sqrt{\sigma_{t_i}/\alpha_{t_i}}) \rceil$ and $w_{\scg}(t_i)=w_{\scg}/(1+\sqrt{\sigma_{t_i}/\alpha_{t_i}})$; 
    \item {Configuration C}: $K_i = \lceil  K_{\iscr}/(1+\sigma_{t_i}/\alpha_{t_i})\rceil$ and $w_{\scg}(t_i)=w_{\scg}/(1+\sigma_{t_i}/\alpha_{t_i})$. 
\end{itemize}
Relative to Configuration~C, Configuration~B applies stronger guidance in the high-noise regime.

\begin{figure}[t]
    \centering
    \begin{subfigure}[b]{0.48\textwidth}
        \centering
        \includegraphics[width=\textwidth]{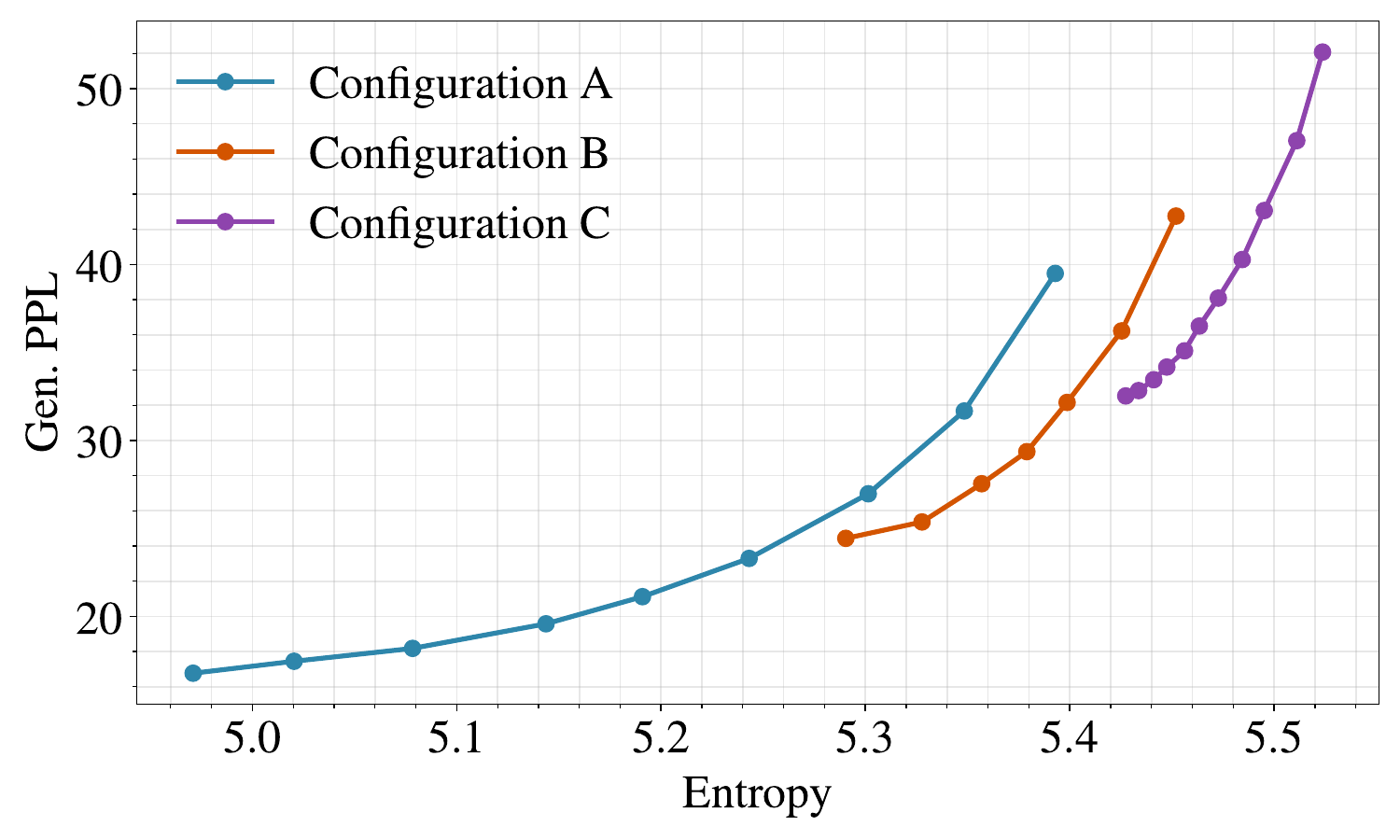}
        \caption{NFE=64}
        \label{fig:k_64}
    \end{subfigure}
    \hfill
    \begin{subfigure}[b]{0.48\textwidth}
        \centering
        \includegraphics[width=\textwidth]{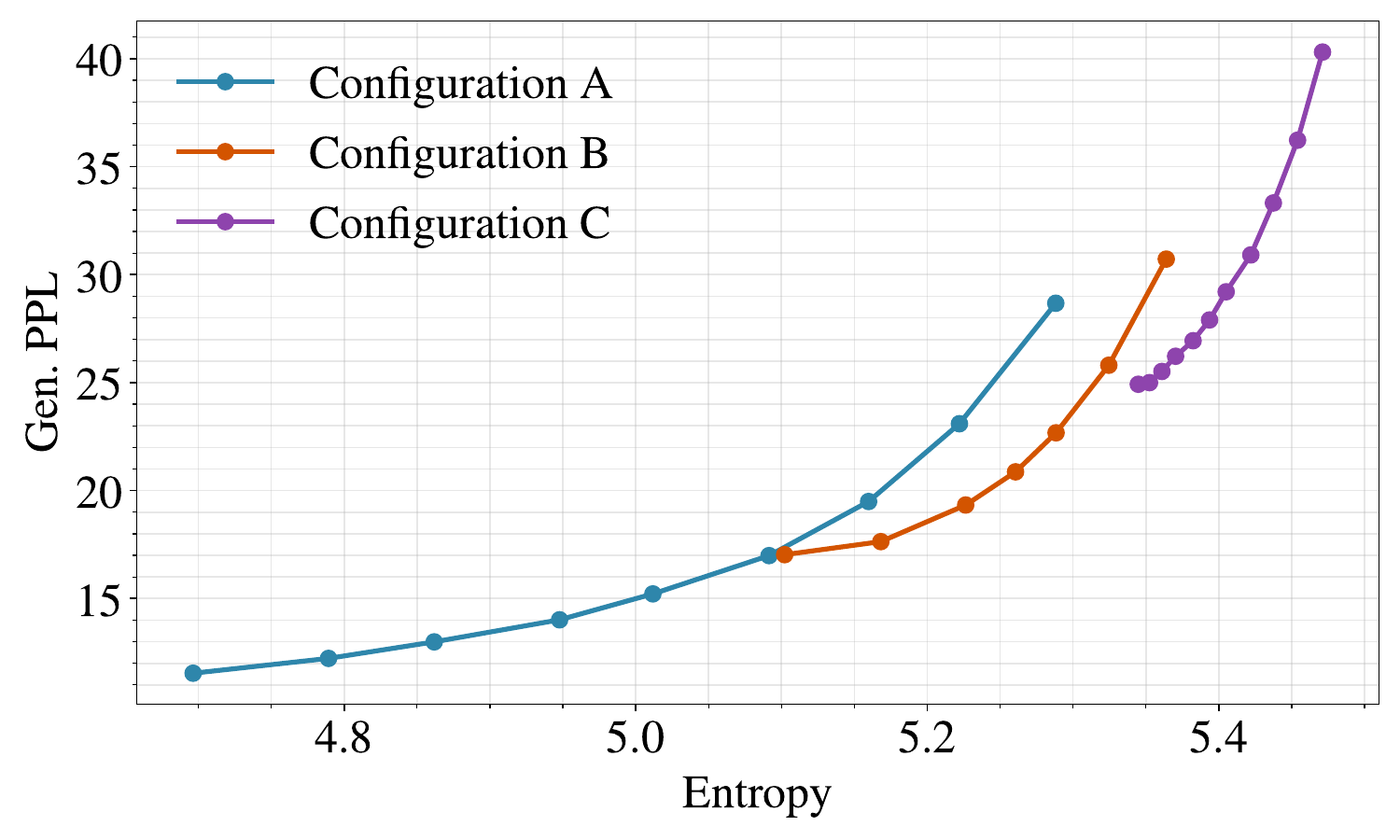}
        \caption{NFE=128}
        \label{fig:k_128}
    \end{subfigure}
    
    \caption{Comparison of the three configurations for NFE budgets of $64$ and $128$. Each curve is obtained by varying the self-conditioning guidance strength $w_{\scg}$.}
    \label{fig:k}
\end{figure}

% The three configs are favorable in different regimes, with results for NFE=$64$ and $128$ are presented separately in Figures~\ref{fig:k_64} and \ref{fig:k_128}, respectively.
Figure~\ref{fig:k} presents the results for NFE=$64$ and $128$. 
The three configurations are favorable in different regions of the quality-diversity frontier. Configuration~A, particularly with $K=8$, reaches the low-entropy, low-Gen.~PPL regime. 
Configuration~B provides strong intermediate operating points by applying more guidance during the noisier portion of the trajectory, whereas Configuration~C preserves the greatest diversity by concentrating guidance near the endpoint.
% Notably, Configuration A with $k=8$ reaching the low-entropy, low-Gen.~PPL region, Configuration B provides strong intermediate operating points, whereas Configuration C preserves the most entropy.

In regions where the curves overlap, increasing the refinement count generally improves the trade-off. Among the settings considered, $K_{\iscr}=100$ and $K_{\iscr}=200$ yield the most favorable frontiers for NFE=$64$ and $128$, respectively.
These results show that both the refinement count and its temporal allocation determine the attainable Gen.~PPL-entropy regime. 

Complete numerical results for Configurations A and B are reported in Tables~\ref{tab:baseline_nfe} and \ref{tab:cfgb_kb_nfe} in Appendix~\ref{subsec:fur_imp_appendix}, respectively. Results for additional guidance configurations under NFE=$64$ and $128$ are provided in Tables~\ref{tab:guidance_nfe64} and \ref{tab:guidance_nfe128} in Appendix~\ref{subsec:fur_imp_appendix}.

\section{Discussion}
\label{sec:discussion}

ConvergeFlow suggests several directions for future work. First, our formulation keeps the token embeddings fixed to avoid degenerate solutions. It would be interesting to develop fully continuous objectives that support joint learning of the embedding and data predictor.
Second, extending the convergence analysis under weaker assumptions and developing non-asymptotic theoretical guarantees would provide a more complete account of practical generation. 
Third, scaling ConvergeFlow to larger models and evaluating it on conditional generation, instruction following, and reasoning tasks will clarify its broader applicability. 
Finally, the continuous formulation may facilitate the use of techniques such as distillation and higher-order acceleration, whose effectiveness for language generation remains to be explored.

\section*{Acknowledgments}
N.~Li, Y.~Jiao, and G.~Li are supported in part by the Chinese University of Hong Kong Direct Grant for Research and the Hong Kong Research Grants Council ECS 24305724 and GRF 14307525.
C.~Cai is supported in part by the NSF CAREER award CCF-2541600 and grant DMS-2515333.

\bibliographystyle{apalike}
\bibliography{bibfileDF}

\appendix

\section{Proof of theorem and propositions}\label{sec:proof}
\subsection{Proof of Theorem~\ref{thm:continuous_flow}}\label{sec:proof_continuous_flow}
In this section, we provide a proof of Theorem~\ref{thm:continuous_flow}, organized into four steps.

\paragraph{Step 1: Construction of an auxiliary sequence.}
Define the probability simplex over the vocabulary as
$$F \coloneqq \bigg\{f \in \mathbb{R}_{+}^V: \sum_{v\in[V]} f(v) = 1\bigg\}.$$

We first note that given an initial point $x_{t_0}$ and a collection of base weights $\{\widehat{f}_k^{(i)}\}_{0\le k\le N-1,1\le i\le L}\subset F$ 
satisfying
$$
\max_{j,k,i} \big|\log \widehat{f}_{k+1}^{(i)}(j) - \log \widehat{f}_k^{(i)}(j)\big| \le \widetilde{L}\,|t_{k+1} - t_k|,
$$
we can use them to construct an auxiliary flow state sequence $\widehat{x}_{t_k}$ as follows.

Initialize the auxiliary sequence $\widehat{x}_{t_{0}}={x}_{t_{0}}$. For each token position $i$ and step~$k$, define the auxiliary weights
\begin{align}
\widehat{w}_{k}^{(i)}(j \mid \widehat{x}_{t_{k}})\coloneqq \frac{\widehat{f}_{k}^{(i)}(j)\exp\Bigl(-\frac{\|\widehat{x}_{t_{k}}^{(i)}-\alpha_{t_{k}}e_j\|_2^2}{2\sigma_{t_{k}}^2}\Bigr)}{\sum_{j=1}^V\widehat{f}_{k}^{(i)}(j)\exp\Bigl(-\frac{\|\widehat{x}_{t_{k}}^{(i)}-\alpha_{t_{k}}e_j\|_2^2}{2\sigma_{t_{k}}^2}\Bigr)},\quad j\in[V].\label{eq:proof-thm-w}
% \frac{1}{\widehat{p}_{k}(\widehat{x}_{t_{k}})}(2\pi\sigma_{t_{k}}^2)^{-d/2} \widehat{f}_{k}(j)\exp\Big(-\frac{\|\widehat{x}_{t_{k}}-\alpha_{t_{k}}e_j\|_2^2}{2\sigma_{t_{k}}^2}\Big).
\end{align}
Thus, $\widehat{w}_{k}^{(i)}(j \mid \widehat{x}_{t_{k}})$ is the counterpart of the learned weight ${w}_{\theta}^{(i)}(j|\widehat{x}_{t_{k}},t_k)$, obtained by replacing the learned base weight $f_{\theta}^{(i)}(j \mid x_{t_k},t_k)$ with the data-independent quantity $\widehat{f}_{k}^{(i)}(j)$.

The corresponding auxiliary data predictor $\widehat{\mu}_{\theta}(\widehat{x}_{t_k},t_k)$ is given by
\begin{align}\label{eq:proof-thm-3}
    \widehat{\mu}_{\theta}^{(i)}(\widehat{x}_{t_k},t_k)\coloneqq \sum_{j=1}^V\widehat{w}_{k}^{(i)}(j \mid \widehat{x}_{t_{k}}) e_j.
\end{align}
We then iteratively update the auxiliary sequence according to
\begin{align}\label{eq:proof-thm-4}
\widehat{x}_{t_{k+1}} = \frac{\sigma_{t_{k+1}}}{\sigma_{t_k}}\widehat{x}_{t_{k}} + \big(\alpha_{t_{k+1}}-\frac{\sigma_{t_{k+1}}\alpha_{t_{k}}}{\sigma_{t_k}}\big) \widehat{\mu}_{\theta}(\widehat{x}_{t_k},t_k),\qquad k=0,1,\ldots,N-1.
\end{align}

With this construction procedure in place, we next show that the auxiliary sequence can reproduce the original sampling trajectory.
Given the original trajectory $(x_{t_k})_{k=0}^N$, choose
\begin{align}\label{eq:proof-thm-f-hat}
\widehat f_k^{(i)}(j)
\coloneqq
f_\theta^{(i)}(j\mid x_{t_k},t_k),
\qquad
\quad j\in[V],
\end{align}
for each iteration $0\leq k < N$ and token position $i\in[L]$. The log-Lipschitz assumption in Theorem~\ref{thm:continuous_flow} ensures that this sequence satisfies the required regularity condition. We claim that
\begin{align}
x_{t_k}=\widehat x_{t_k},
\qquad 0\leq k\leq N.
\label{eq:proof-temp-2}
\end{align}
% \begin{align}\label{eq:proof-temp-2}
% x_{t_k} = \widehat{x}_{t_k}, \qquad k=0,1,\ldots,N.
% \end{align}
% where $\widehat{x}_{t_k}$ denotes the flow initialized at $x_{t_0}$ with $\widehat{f}_k^{(i)}(\cdot)$ used in place of $f_{\theta}^{(i)}(\cdot|x_{t_k},t_k)$.

We prove this claim by induction. 
First, it holds trivially for $k=0$.
Next, suppose that $x_{t_k} = \widehat{x}_{t_k}$ holds for some $k$.
By the choice of $\widehat{f}_{k}^{(i)}(j)$ in \eqref{eq:proof-thm-f-hat} and the construction of $\widehat{w}_{k}^{(i)}$ in \eqref{eq:proof-thm-w}, we have
\begin{align}\label{eq:proof-thm-1}
\widehat{w}_{k}^{(i)}(j \mid \widehat{x}_{t_{k}}) = {w}_{\theta}^{(i)}(j|\widehat{x}_{t_{k}},t_k)={w}_{\theta}^{(i)}(j \mid {x}_{t_{k}},t_k).
\end{align}
It then follows from \eqref{eq:proof-thm-3} and
\eqref{eq:posterior_structured_data_predictor} that
$$
\widehat{\mu}_{\theta}^{(i)}(\widehat{x}_{t_k},t_k) = \sum_{j=1}^V{w}_{\theta}^{(i)}(j \mid {x}_{t_{k}},t_k) e_j
 = {\mu}_{\theta}^{(i)}({x}_{t_k},t_k).
$$
Therefore, the update rule of the auxiliary sequence in \eqref{eq:proof-thm-4} coincides with that of the original sequence in \eqref{eq:data_prediction_inference}, thereby yielding $x_{t_{k+1}} = \widehat{x}_{t_{k+1}}$.
This completes the induction and proves \eqref{eq:proof-temp-2}.

Consequently, it remains to analyze the auxiliary sequence for an arbitrary collection of base weights satisfying the stated regularity condition. 
Concretely, we will show that for any token position $i\in [L]$, there exists some $j_i\in [V]$, such that 
$$
\widehat{x}_{t_N}^{(i)} \to e_{j_i},\qquad \mathrm{as}\quad t_N \to 1.
$$

\paragraph{Step 2: Construction of reference distributions.}

For each token position $i$, let $q_{k}^{(i)}$ denote the probability density function of $\widehat{x}_{t_k}^{(i)}$ when $\widehat{x}_{t_0}^{(i)}$ is initialized with the standard Gaussian distribution. In addition, for each $0\leq k \leq N$ and $i\in[L]$, we define the reference distribution
\begin{align*}
\widehat{p}_k^{(i)}(x) \defn (2\pi\sigma_{t_k}^2)^{-d/2}\sum_{j=1}^V \widehat{f}_k^{(i)}(j)\exp\Bigl(-\frac{\|x - \alpha_{t_k}e_j\|_2^2}{2\sigma_{t_k}^2}\Bigr),\qquad \forall x\in\mathbb{R}^d.
\end{align*}

In the remainder of the proof, we focus on a token position $i$. For notational brevity, we omit the superscript $(i)$ from $\widehat{f}_k^{(i)}$, $\widehat{w}_k^{(i)}$, $\widehat{\mu}_{\theta}^{(i)}$, $\widehat{p}_k^{(i)}$ and $\widehat{q}_k^{(i)}$, and $\widehat{x}_{t_{k}}^{(i)}$.

We first compare the reference densities at two consecutive points along the auxiliary trajectory.
By the definition of $\widehat{p}_{k}$, one can derive
\begin{align}\label{eq:ratio-p-2}
\frac{\widehat{p}_{k+1}(\widehat{x}_{t_{k+1}})}{\widehat{p}_{k}(\widehat{x}_{t_{k}})} &= \frac{1}{\widehat{p}_{k}(\widehat{x}_{t_{k}})}(2\pi\sigma_{t_{k+1}}^2)^{-d/2}\sum_{j=1}^V \widehat{f}_{k+1}(j)\exp\Bigl(-\frac{\|\widehat{x}_{t_{k+1}}-\alpha_{t_{k+1}}e_j\|_2^2}{2\sigma_{t_{k+1}}^2}\Bigr)\notag\\
&=\frac{1}{\widehat{p}_{k}(\widehat{x}_{t_{k}})}(2\pi\sigma_{t_{k}}^2)^{-d/2}\Big(\frac{\sigma_{t_{k}}}{\sigma_{t_{k+1}}}\Big)^{d}\sum_{j=1}^V \widehat{f}_{k}(j)\exp\Bigl(-\frac{\|\widehat{x}_{t_{k}}-\alpha_{t_{k}}e_j\|_2^2}{2\sigma_{t_{k}}^2}\Bigr)\frac{\widehat{f}_{k+1}(j)}{\widehat{f}_{k}(j)}\exp\bigl(\Delta_{t_k}(e_j)\bigr)\notag\\
&\overset{\text{(i)}}{\ge} \exp\bigl(-\widetilde{L}(t_{k+1}-t_{k})\bigr)\frac{1}{\widehat{p}_{k}(\widehat{x}_{t_{k}})}(2\pi\sigma_{t_{k}}^2)^{-d/2}\Big(\frac{\sigma_{t_{k}}}{\sigma_{t_{k+1}}}\Big)^{d}\sum_{j=1}^V \widehat{f}_{k}(j)\exp\Bigl(-\frac{\|\widehat{x}_{t_{k}}-\alpha_{t_{k}}e_j\|_2^2}{2\sigma_{t_{k}}^2}\Bigr)\exp\bigl(\Delta_{t_k}(e_j)\bigr)\notag\\
&\overset{\text{(ii)}}{=} \exp\bigl(-\widetilde{L}(t_{k+1}-t_{k})\bigr)\Big(\frac{\sigma_{t_{k}}}{\sigma_{t_{k+1}}}\Big)^{d}\sum_{j=1}^V \widehat{w}_{k}(j \mid \widehat{x}_{t_{k}})\exp\bigl(\Delta_{t_k}(e_j)\bigr)\notag\\
&\overset{\text{(iii)}}{\ge} \exp\bigl(-\widetilde{L}(t_{k+1}-t_{k})\bigr)\Big(\frac{\sigma_{t_{k}}}{\sigma_{t_{k+1}}}\Big)^{d}\exp\biggl(\sum_{j=1}^V \widehat{w}_{k}(j \mid \widehat{x}_{t_{k}})\Delta_{t_k}(e_j)\biggr).
\end{align}
Here, $\Delta_{t_k}(e_j)$ records the change in the Gaussian exponent associated with the $j$-th token embedding:
\begin{align}
\Delta_{t_k}(e_j)&\coloneqq \frac{\|\widehat{x}_{t_{k}}-\alpha_{t_{k}}e_j\|_2^2}{2\sigma_{t_{k}}^2}-\frac{\|\widehat{x}_{t_{k+1}}-\alpha_{t_{k+1}}e_j\|_2^2}{2\sigma_{t_{k+1}}^2}\notag\\
&=\frac12\Big(1-\frac{\alpha_{t_{k+1}}^2\sigma_{t_k}^2}{\alpha_{t_{k}}^2\sigma_{t_{k+1}}^2}\Big)\frac{\|\widehat{x}_{t_k}-\alpha_{t_k}e_j\|_2^2}{\sigma_{t_k}^2}-\frac12\Big(1-\frac{\alpha_{t_{k+1}}^2\sigma_{t_k}^2}{\alpha_{t_{k}}^2\sigma_{t_{k+1}}^2}\Big)\frac{\|\widehat{x}_{t_k}-\alpha_{t_k}\widehat{\mu}_{\theta}(\widehat{x}_{t_k},t_k)\|_2^2}{\sigma_{t_k}^2}, \label{eq:proof-thm-2}
\end{align}
where the second line follows from the update rule of $\widehat{x}_{t_{k}}$ given in \eqref{eq:proof-thm-4} and the definition of  $\widehat{\mu}_{\theta}(\widehat{x}_{t_k},t_k)$ from \eqref{eq:proof-thm-3}; (i) follows from the log-Lipschitz condition on $\wh f_k$; (ii) follows from the identity
\begin{align*}
\widehat{w}_{k}(j \mid \widehat{x}_{t_{k}}) = \frac{\widehat{f}_{k}(j)\exp\Bigl(-\frac{\|\widehat{x}_{t_{k}}-\alpha_{t_{k}}e_j\|_2^2}{2\sigma_{t_{k}}^2}\Bigr)}{\sum_{{j'}=1}^V\widehat{f}_{k}({j'})\exp\Bigl(-\frac{\|\widehat{x}_{t_{k}}-\alpha_{t_{k}}e_{j'}\|_2^2}{2\sigma_{t_{k}}^2}\Bigr)} = \frac{1}{\widehat{p}_{k}(\widehat{x}_{t_{k}})}(2\pi\sigma_{t_{k}}^2)^{-d/2} \widehat{f}_{k}(j)\exp\Bigl(-\frac{\|\widehat{x}_{t_{k}}-\alpha_{t_{k}}e_j\|_2^2}{2\sigma_{t_{k}}^2}\Bigr);
\end{align*}
(iii) is a consequence of Jensen's inequality $\mathbb{E}[e^X] \ge e^{\mathbb{E}[X]}$.
By summary, we have
\begin{align}\label{eq:proof-thm-6}
\sum_{j=1}^V \widehat{w}_{k}(j \mid \widehat{x}_{t_{k}})\Delta_{t_k}(e_j) 
&= \frac12\Big(1-\frac{\alpha_{t_{k+1}}^2\sigma_{t_k}^2}{\alpha_{t_{k}}^2\sigma_{t_{k+1}}^2}\Big)\Bigg(\sum_{j=1}^V \widehat{w}_{k}(j \mid \widehat{x}_{t_{k}}) \frac{\|\widehat{x}_{t_k}-\alpha_{t_k}e_j\|_2^2}{\sigma_{t_k}^2}- \frac{\|\widehat{x}_{t_k}-\alpha_{t_k}\widehat{\mu}_{\theta}(\widehat{x}_{t_k},t_k)\|_2^2}{\sigma_{t_k}^2}\Bigg)\notag\\
&=\frac{\alpha_{t_k}^2}{2\sigma_{t_k}^2}\Big(1-\frac{\alpha_{t_{k+1}}^2\sigma_{t_k}^2}{\alpha_{t_{k}}^2\sigma_{t_{k+1}}^2}\Big)\Bigg(\sum_{j=1}^V \widehat{w}_{k}(j \mid \widehat{x}_{t_{k}}) \|e_j\|_2^2 - \|\widehat{\mu}_{\theta}(\widehat{x}_{t_k},t_k)\|_2^2\Bigg).
\end{align}
Substituting into \eqref{eq:ratio-p-2} yields
\begin{align}\label{eq:ratio-p}
\frac{\widehat{p}_{k+1}(\widehat{x}_{t_{k+1}})}{\widehat{p}_{k}(\widehat{x}_{t_{k}})} &\ge \exp\bigl(-\widetilde{L}(t_{k+1}-t_{k})\bigr)\Big(\frac{\sigma_{t_{k}}}{\sigma_{t_{k+1}}}\Big)^{d}\exp\biggl(\frac{\alpha_{t_k}^2}{2\sigma_{t_k}^2}\Big(1-\frac{\alpha_{t_{k+1}}^2\sigma_{t_k}^2}{\alpha_{t_{k}}^2\sigma_{t_{k+1}}^2}\Big)\biggl(\sum_{j=1}^V \widehat{w}_{k}(j \mid \widehat{x}_{t_{k}}) \|e_j\|_2^2 - \|\widehat{\mu}_{\theta}(\widehat{x}_{t_k},t_k)\|_2^2\biggr)\biggr).
\end{align}
\paragraph{Step 3: Analysis of the distributions of $\widehat{x}_{t_k}$.}
% Moreover, according to the update rule of $\widehat{x}_{t_{k+1}}$, which is given by
We now track the density of the auxiliary sequence $\widehat{x}_{t_{k}}$. Because the update is deterministic, its one-step density ratio is determined by the inverse Jacobian determinant of the update map.
From the update rule in \eqref{eq:proof-thm-4}, the change-of-variable formula yields the following relationship between the densities of $\widehat{x}_{t_{k}}$ and $\widehat{x}_{t_{k+1}}$:
\begin{align}\label{eq:ratio-q-1}
\frac{q_{k+1}(\widehat{x}_{t_{k+1}})}{q_{k}(\widehat{x}_{t_{k}})}
&=\Big|\frac{\sigma_{t_{k+1}}}{\sigma_{t_k}}I + \big(\alpha_{t_{k+1}}-\frac{\sigma_{t_{k+1}}\alpha_{t_{k}}}{\sigma_{t_k}}\big)\nabla_{\widehat{x}_{t_k}} \widehat{\mu}_{\theta}(\widehat{x}_{t_k},t_k)\Big|^{-1}\notag\\
&=\Big(\frac{\sigma_{t_{k+1}}}{\sigma_{t_k}}\Big)^{-d}\Big|I -\alpha_{t_k}\Big(1-\frac{\alpha_{t_{k+1}}\sigma_{t_{k}}}{\alpha_{t_{k}}\sigma_{t_{k+1}}}\Big)\nabla_{\widehat{x}_{t_{k}}}\widehat{\mu}_{\theta}(\widehat{x}_{t_k},t_k)\Big|^{-1}.
\end{align}
where $|\cdot|$ denotes the determinant.
By the construction of the embedding-weighted predictor $\widehat{\mu}_{\theta}(\widehat{x}_{t_k},t_k)$ in \eqref{eq:proof-thm-3}, we obtain
\begin{align*}
\nabla_{\widehat{x}_{t_{k}}}\widehat{\mu}_{\theta}(\widehat{x}_{t_k},t_k) 
&= \sum_{j=1}^V e_j \big(\nabla_{\widehat{x}_{t_{k}}}\widehat{w}_{k}(j \mid \widehat{x}_{t_{k}})\big)^{\top} \notag\\
&= -\frac{1}{\sigma_{t_k}^2}\sum_{j=1}^V\widehat{w}_{k}(j \mid \widehat{x}_{t_{k}})e_j (\widehat{x}_{t_{k}}-\alpha_{t_k}e_j)^{\top} + \frac{1}{\sigma_{t_k}^2} \big(\sum_{j=1}^V\widehat{w}_{k}(j \mid \widehat{x}_{t_{k}})e_j\big) \big(\sum_{j=1}^V \widehat{w}_{k}(j \mid \widehat{x}_{t_{k}}) (\widehat{x}_{t_{k}}-\alpha_{t_k}e_j)\big)^{\top}\notag\\
% &= \frac{1}{\alpha_{t_k}\sigma_{t_k}^2}\bigg\{\sum_{j=1}^V\widehat{w}_{k}^{(i)}(j \mid \widehat{x}_{t_{k}})(\widehat{x}_{t_{k}}-\alpha_{t_k}e_j) (\widehat{x}_{t_{k}}-\alpha_{t_k}e_j)^{\top} \notag\\
% &\qquad\qquad\qquad - \bigg(\sum_{j=1}^V\widehat{w}_{k}^{(i)}(j \mid \widehat{x}_{t_{k}})(\widehat{x}_{t_{k}}-\alpha_{t_k}e_j)\bigg) \bigg(\sum_{j=1}^V \widehat{w}_{k}^{(i)}(j \mid \widehat{x}_{t_{k}}) (\widehat{x}_{t_{k}}-\alpha_{t_k}e_j)\bigg)^{\top}\bigg\}\notag\\
&=\frac{\alpha_{t_k}}{\sigma_{t_k}^2}\Bigg(\sum_{j=1}^V\widehat{w}_{k}(j \mid \widehat{x}_{t_{k}}) e_je_j^{\top} - \big(\sum_{j=1}^V\widehat{w}_{k}(j \mid \widehat{x}_{t_{k}}) e_j\big)\big(\sum_{j=1}^V\widehat{w}_{k}(j \mid \widehat{x}_{t_{k}}) e_j\big)^{\top}\Bigg)\notag\\
&\overset{\text{(i)}}{=}\frac{\alpha_{t_k}}{\sigma_{t_k}^2}\Bigg(\sum_{j=1}^V\widehat{w}_{k}(j \mid \widehat{x}_{t_{k}}) e_je_j^{\top} - \widehat{\mu}_{\theta}(\widehat{x}_{t_k},t_k)\widehat{\mu}_{\theta}(\widehat{x}_{t_k},t_k)^{\top}\Bigg)\notag\\
&\eqqcolon \frac{\alpha_{t_k}}{\sigma_{t_k}^2}\widehat{\Sigma}_k(\widehat{x}_{t_{k}}),
\end{align*}
where (i) applies the definition of $\widehat{\mu}_{\theta}(\widehat{x}_{t_k},t_k)$ in \eqref{eq:proof-thm-3}, and $\widehat{\Sigma}_k(\widehat{x}_{t_{k}})$ is the covariance matrix of discrete distribution on $\{e_j\}_{j\in[V]}$ associated with the probability $\mathbb{P}\{X=e_j\}=\widehat{w}_{k}(j \mid \widehat{x}_{t_{k}})$. 
% The matrix in parentheses above 
% is the covariance of the residual vector $\widehat{x}_{t_{k}}-\alpha_{t_k}e_j$ associated with the weights $\widehat{w}_{k}(j \mid \widehat{x}_{t_{k}})$. 
This is the same quantity that appears in the reference-density calculation in Step 2; see \eqref{eq:proof-thm-6}.

Substituting this expression into \eqref{eq:ratio-q-1} yields
\begin{align}\label{eq:ratio-q}
\frac{q_{k+1}(\widehat{x}_{t_{k+1}})}{q_{k}(\widehat{x}_{t_{k}})}
&=\Big(\frac{\sigma_{t_{k+1}}}{\sigma_{t_k}}\Big)^{-d}\bigg|I -\frac{\alpha_{t_k}^2}{\sigma_{t_k}^2}\Big(1-\frac{\alpha_{t_{k+1}}\sigma_{t_{k}}}{\alpha_{t_{k}}\sigma_{t_{k+1}}}\Big)\widehat{\Sigma}_k(\widehat{x}_{t_{k}})\bigg|^{-1}\notag\\
% &=\Big(\frac{\sigma_{t_{k+1}}}{\sigma_{t_k}}\Big)^{-d}\bigg|I -\frac{1}{\sigma_{t_k}^2}\Big(1-\frac{\alpha_{t_{k+1}}\sigma_{t_{k}}}{\alpha_{t_{k}}\sigma_{t_{k+1}}}\Big)\bigg\{\sum_{j=1}^V\widehat{w}_{k}(j \mid \widehat{x}_{t_{k}}) (\widehat{x}_{t_k}-\alpha_{t_k}e_j)(\widehat{x}_{t_k}-\alpha_{t_k}e_j)^{\top}\notag\\
% &\qquad\qquad - \bigg(\sum_{j=1}^V\widehat{w}_{k}(j \mid \widehat{x}_{t_{k}}) (\widehat{x}_{t_k}-\alpha_{t_k}e_j)\bigg)\bigg(\sum_{j=1}^V\widehat{w}_{k}(j \mid \widehat{x}_{t_{k}}) (\widehat{x}_{t_k}-\alpha_{t_k}e_j)\bigg)^{\top}\bigg\}\bigg|^{-1}\notag\\
&\overset{\text{(i)}}{=}\Big(\frac{\sigma_{t_{k+1}}}{\sigma_{t_k}}\Big)^{-d}\exp\Biggl(\frac{\alpha_{t_k}^2}{\sigma_{t_k}^2}
   \Big(1-\frac{\alpha_{t_{k+1}}\sigma_{t_{k}}}{\alpha_{t_{k}}\sigma_{t_{k+1}}}\Big) 
\mathsf{Tr}(\widehat{\Sigma}_k(\widehat{x}_{t_{k}})) + \frac{\alpha_{t_k}^4}{\sigma_{t_k}^4}
   \Big(1-\frac{\alpha_{t_{k+1}}\sigma_{t_{k}}}{\alpha_{t_{k}}\sigma_{t_{k+1}}}\Big)^2O\big(\|\widehat{\Sigma}_k(\widehat{x}_{t_{k}})\|_F^2\big) 
\Biggr)\notag\\
&=\Big(\frac{\sigma_{t_{k+1}}}{\sigma_{t_k}}\Big)^{-d}\exp\Biggl(\frac{\alpha_{t_k}^2}{\sigma_{t_k}^2}
   \Big(1-\frac{\alpha_{t_{k+1}}\sigma_{t_{k}}}{\alpha_{t_{k}}\sigma_{t_{k+1}}}\Big) 
\bigg(\sum_{j=1}^V\widehat{w}_{k}(j \mid \widehat{x}_{t_{k}}) \|e_j\|_2^2 - \|\widehat{\mu}_{\theta}(\widehat{x}_{t_k},t_k)\|_2^2\bigg)  + \mathcal{E}_{t_k}(\widehat{x}_{t_k}).
\Biggr)
% &=\Big(\frac{\sigma_{t_{k+1}}}{\sigma_{t_k}}\Big)^{-d}\exp\Bigg(\frac{1}{\sigma_{t_k}^2}
%    \Big(1-\frac{\alpha_{t_{k+1}}\sigma_{t_{k}}}{\alpha_{t_{k}}\sigma_{t_{k+1}}}\Big) 
% \bigg(\sum_{j=1}^V\widehat{w}_{k}(j \mid \widehat{x}_{t_{k}}) \big\|\widehat{x}_{t_k}-\alpha_{t_k}e_j\big\|_2^2 - \big\|\widehat{x}_{t_k}-\alpha_{t_k}\widehat{\mu}_{\theta}(\widehat{x}_{t_k},t_k)\big\|_2^2\bigg) \notag\\
% &\qquad\qquad\qquad\qquad\quad + \mathcal{E}_{t_k}(\widehat{x}_{t_k})
% \Bigg).
\end{align}
Here the leading term in the right-hand-side of (i) is obtained from the first-order expansion of the log determinant, while the remainder collects the corresponding higher-order terms, and 
% Specifically,
\begin{align*}
\mathcal{E}_{t_k}(\widehat{x}_{t_k}) \coloneqq \frac{\alpha_{t_k}^4}{\sigma_{t_k}^4}
   \Big(1-\frac{\alpha_{t_{k+1}}\sigma_{t_{k}}}{\alpha_{t_{k}}\sigma_{t_{k+1}}}\Big)^2O\big(\|\widehat{\Sigma}_k(\widehat{x}_{t_{k}})\|_F^2\big).
% O\bigg(\frac{1}{\sigma_{t_k}^4}
%    \Big(1-\frac{\alpha_{t_{k+1}}\sigma_{t_{k}}}{\alpha_{t_{k}}\sigma_{t_{k+1}}}\Big)^2 \Big(\sum_{j=1}^V\widehat{w}_{k}(j \mid \widehat{x}_{t_{k}}) \|\widehat{x}_{t_k}-\alpha_{t_k}e_j\|_2^2\Big)^2\bigg).
\end{align*}

\paragraph{Step 4: Combining the estimates.}
We finally compare the two density evolutions. Combining \eqref{eq:ratio-p} and \eqref{eq:ratio-q}, and using the identity
$$
1-\frac{\alpha_{t_{k+1}}\sigma_{t_{k}}}{\alpha_{t_{k}}\sigma_{t_{k+1}}} - \bigg(\frac12-\frac{\alpha_{t_{k+1}}^2\sigma_{t_{k}}^2}{2\alpha_{t_{k}}^2\sigma_{t_{k+1}}^2}\bigg) = \frac{1}{2}\bigg(1-\frac{\alpha_{t_{k+1}}\sigma_{t_{k}}}{\alpha_{t_{k}}\sigma_{t_{k+1}}}\bigg)^2,
$$
we obtain
\begin{align*}
    &\frac{q_{k+1}(\widehat{x}_{t_{k+1}})\widehat{p}_{k}(\widehat{x}_{t_{k}})}{q_{k}(\widehat{x}_{t_{k}})\widehat{p}_{k+1}(\widehat{x}_{t_{k+1}})} \notag\\
    &\quad\le \exp\bigl(\wt L(t_{k+1}-t_{k})\bigr)\exp\Biggl(\frac{\alpha_{t_k}^2}{2\sigma_{t_k}^2}\Big(1-\frac{\alpha_{t_{k+1}}\sigma_{t_{k}}}{\alpha_{t_{k}}\sigma_{t_{k+1}}}\Big)^2\bigg(\sum_{j=1}^V\widehat{w}_{k}(j \mid \widehat{x}_{t_{k}}) \|e_j\|_2^2 - \|\widehat{\mu}_{\theta}(\widehat{x}_{t_k},t_k)\|_2^2\bigg) +\mathcal{E}_{t_k}(\widehat{x}_{t_k})\Biggr)\notag\\
&\quad\le \exp\bigl(\wt L(t_{k+1}-t_{k})\bigr)\exp\Biggl(\frac{\alpha_{t_k}^2}{2\sigma_{t_k}^2}\Big(1-\frac{\alpha_{t_{k+1}}\sigma_{t_{k}}}{\alpha_{t_{k}}\sigma_{t_{k+1}}}\Big)^2\sum_{j=1}^V\widehat{w}_{k}(j \mid \widehat{x}_{t_{k}}) \|e_j\|_2^2 +\mathcal{E}_{t_k}(\widehat{x}_{t_k})\Biggr).
\end{align*}

% To control the right-hand side, recalling that $\|e_j\|_2$ is bounded, i.e., $\|e_j\|_2\le B$, we have $\|\widehat{\mu}_{\theta}(\widehat{x}_{t_k},t_k)\|_2\le B$.
% Moreover, $\|\widehat{x}_{t_{k+1}}-\alpha_{t_{k+1}}e_j\|_2$ is bounded as follows:
% \begin{align*}
% \|\widehat{x}_{t_{k+1}}-\alpha_{t_{k+1}}e_j\|_2 
% &= \bigg\|\frac{\sigma_{t_{k+1}}}{\sigma_{t_k}}(\widehat{x}_{t_{k}}-\alpha_{t_{k}}e_j) + \Big(\alpha_{t_{k+1}}-\frac{\sigma_{t_{k+1}}\alpha_{t_k}}{\sigma_{t_k}}\bigg)({\mu}_{\theta}(\widehat{x}_{t_k},t_k)-e_j)\bigg\|_2 \notag\\
% &\le \frac{\sigma_{t_{k+1}}}{\sigma_{t_k}}\|\widehat{x}_{t_{k}}-\alpha_{t_{k}}e_j\|_2 + \Big(\alpha_{t_{k+1}}-\frac{\sigma_{t_{k+1}}\alpha_{t_k}}{\sigma_{t_k}}\Big)B\notag\\
% &\le \frac{\sigma_{t_{k+1}}}{\sigma_{t_0}}\|\widehat{x}_{t_{0}}-\alpha_{t_{0}}e_j\|_2 + \sum_{n=1}^k \prod_{s=n+1}^k\frac{\sigma_{t_{s+1}}}{\sigma_{t_s}} \Big(\alpha_{t_{n+1}}-\frac{\sigma_{t_{n+1}}\alpha_{t_n}}{\sigma_{t_n}}\Big)B \notag\\
% &= \frac{\sigma_{t_{k+1}}}{\sigma_{t_0}}\|\widehat{x}_{t_{0}}-\alpha_{t_{0}}e_j\|_2 + \sigma_{t_{k+1}}\sum_{n=1}^k \Big(\frac{\alpha_{t_{n+1}}}{\sigma_{t_{n+1}}}-\frac{\alpha_{t_n}}{\sigma_{t_n}}\bigg)B\notag\\
% &=\frac{\sigma_{t_{k+1}}}{\sigma_{t_0}}\|\widehat{x}_{t_{0}}-\alpha_{t_{0}}e_j\|_2 +\sigma_{t_{k+1}}\Big(\frac{\alpha_{t_{k+1}}}{\sigma_{t_{k+1}}}-\frac{\alpha_{t_1}}{\sigma_{t_1}}\Big)B \\
% & \le \frac{\sigma_{t_{k+1}}}{\sigma_{t_0}}\|\widehat{x}_{t_{0}}-\alpha_{t_{0}}e_j\|_2 +\alpha_{t_{k+1}}B.
% \end{align*}
To control the right-hand side, recall that $\|e_j\|_2$ is bounded, i.e., $\|e_j\|_2= B$.
Consequently, telescoping the one-step density-ratio bounds gives
\begin{align*}
\frac{q_{N}(\widehat{x}_{t_{N}})}{\widehat{p}_{N}(\widehat{x}_{t_{N}})} &= \frac{q_{0}(\widehat{x}_{t_{0}})}{\widehat{p}_{0}(\widehat{x}_{t_{0}})} \prod_{k=0}^{N-1}\frac{q_{k+1}(\widehat{x}_{t_{k+1}})\widehat{p}_{k}(\widehat{x}_{t_{k}})}{q_{k}(\widehat{x}_{t_{k}})\widehat{p}_{k+1}(\widehat{x}_{t_{k+1}})} \notag\\
&\le \frac{q_{0}(\widehat{x}_{t_{0}})}{\widehat{p}_{0}(\widehat{x}_{t_{0}})}\exp(\wt L)\exp\bigg(\sum_{k=0}^{N-1}\frac{\alpha_{t_{k}}^2}{2\sigma_{t_k}^2}\Big(1-\frac{\alpha_{t_{k+1}}\sigma_{t_{k}}}{\alpha_{t_{k}}\sigma_{t_{k+1}}}\Big)^2B^2 +\sum_{k=0}^{N-1}\mathcal{E}_{t_k}(\widehat{x}_{t_k})\bigg),
\end{align*}
where we have used
\begin{align*}
&\sum_{k=0}^{N-1}\frac{\alpha_{t_k}^2}{2\sigma_{t_k}^2}\Big(1-\frac{\alpha_{t_{k+1}}\sigma_{t_{k}}}{\alpha_{t_{k}}\sigma_{t_{k+1}}}\Big)^2\sum_{j=1}^V\widehat{w}_{k}(j \mid \widehat{x}_{t_{k}}) \|e_j\|_2^2 \notag\\
& = \frac12\sum_{k=0}^{N-1}\Big(\frac{\alpha_{t_k}}{\sigma_{t_k}}-\frac{\alpha_{t_{k+1}}}{\sigma_{t_{k+1}}}\Big)^2B^2 \lesssim N\delta^2,
% &\qquad \lesssim \frac{\delta^2N}{(1-t_0)^2}\max_{j}\|\widehat{x}_{t_{0}}-\alpha_{t_{0}}e_j\|_2^2 + \delta^2NB^2,
\end{align*}
and
\begin{align*}
\sum_{k=0}^{N-1}\mathcal{E}_{t_k}(\widehat{x}_{t_k}) 
&\lesssim \sum_{k=0}^{N-1}\frac{\alpha_{t_k}^4}{\sigma_{t_k}^4}\Big(1-\frac{\alpha_{t_{k+1}}\sigma_{t_{k}}}{\alpha_{t_{k}}\sigma_{t_{k+1}}}\Big)^2\mathsf{Tr}(\widehat{\Sigma}_k(\widehat{x}_{t_k}))^2\notag\\
&\lesssim \sum_{k=0}^{N-1}\frac{\alpha_{t_k}^4}{\sigma_{t_k}^4}\Big(1-\frac{\alpha_{t_{k+1}}\sigma_{t_{k}}}{\alpha_{t_{k}}\sigma_{t_{k+1}}}\Big)^2\mathsf{Tr}\Big(\sum_{j=1}^V\widehat{w}_{k}(j \mid \widehat{x}_{t_{k}})\|e_j\|_2^2\Big)^2\notag\\
&=\sum_{k=0}^{N-1}\frac{\alpha_{t_k}^2}{\sigma_{t_k}^2}\Big(\frac{\alpha_{t_k}}{\sigma_{t_k}}-\frac{\alpha_{t_{k+1}}}{\sigma_{t_{k+1}}}\Big)^2B^4\lesssim N\delta^2B^4,
% =\Big(\sum_{j=1}^V\widehat{w}_{k}(j \mid \widehat{x}_{t_{k}}) \|\widehat{x}_{t_k}-\alpha_{t_k}e_j\|_2^2 \Big)^2\notag\\
% &\lesssim \sum_{k=0}^{N-1}\Big(1-\frac{\alpha_{t_{k+1}}\sigma_{t_{k}}}{\alpha_{t_{k}}\sigma_{t_{k+1}}}\Big)^2\frac{1}{\sigma_{t_0}^4}\max_{e_j}\|\widehat{x}_{t_{0}}-\alpha_{t_{0}}e_j\|_2^4 + \sum_{k=0}^{N-1}\Big(\frac{\alpha_{t_{k}}}{\sigma_{t_{k}}}-\frac{\alpha_{t_{k+1}}}{\sigma_{t_{k+1}}}\Big)^2\frac{\alpha_{t_k}^2}{\sigma_{t_k}^2}B^4\notag\\
% &\lesssim \frac{\delta^2N}{(1-t_0)^4}\max_{j}\|\widehat{x}_{t_{0}}-\alpha_{t_{0}}e_j\|_2^4 + \delta^2NB^4.
\end{align*}
where the last inequality uses the facts that
\begin{align}
\big|\frac{\alpha_{t_k}}{\sigma_{t_k}}-\frac{\alpha_{t_{k+1}}}{\sigma_{t_{k+1}}}\big| &= \frac{t_{k+1}-t_k}{(1-t_k)(1-t_{k+1})} \le \delta,\notag\\
\frac{\alpha_{t_k}}{\sigma_{t_k}}\big|\frac{\alpha_{t_k}}{\sigma_{t_k}}-\frac{\alpha_{t_{k+1}}}{\sigma_{t_{k+1}}}\big| &= \frac{(t_{k+1}-t_k)t_k}{(1-t_k)^2(1-t_{k+1})} \le \delta.
\end{align}
It remains to control the initial density ratio.  Observe that the initial time step satisfies
\begin{align*}
\frac{q_{0}(\widehat{x}_{t_{0}})}{\widehat{p}_{0}(\widehat{x}_{t_{0}})} =\sigma_{t_0}^{d}\frac{1}{\sum_{j=1}^V \widehat{f}_{k}(j)\exp\big(-\frac{\alpha_{t_0}^2\|e_j\|_2^2}{2\sigma_{t_0}^2}+\frac{\alpha_{t_0}e_j^{\top}\widehat{x}_{t_0}}{2\sigma_{t_0}^2}-\frac{1-\sigma_{t_0}^2}{2\sigma_{t_0}^2}\|\widehat{x}_{t_0}\|_2^2\big)} \to 1 \quad\text{as}~~t_0\to 0.
\end{align*}
Under the stated discretization condition, the accumulated remainder remains bounded. In particular,for sufficiently small $\delta$, we have $\delta^2N=O(1)$. Consequently, as $t_0\to 0$, 
\begin{align*}
\frac{q_{N}(\widehat{x}_{t_{N}})}{\widehat{p}_{N}(\widehat{x}_{t_{N}})} = O(1).
\end{align*}
% Recall that 
% \begin{align*}
% \widehat{p}_N(\widehat{x}_{t_N}) = (2\pi\sigma_{t_N}^2)^{-d/2}\sum_{j=1}^V \widehat{w}_N(e_j)\exp\bigl(-\|\widehat{x}_{t_N} - \alpha_{t_N}e_j\|_2^2/(2\sigma_{t_N}^2)\big).
% \end{align*}
Finally, recall that
\begin{align*}
    \widehat{p}_{N}(\widehat{x}_{t_{N}}) = (2\pi\sigma_{t_N}^2)^{-d/2}\sum_{j=1}^V \widehat{w}_N(e_j)\exp\bigl(-\|\widehat{x}_{t_N} - \alpha_{t_N}e_j\|_2^2/(2\sigma_{t_N}^2)\bigr),
\end{align*}
which implies that
\begin{align*}
q_{N}(\widehat{x}_{t_{N}})\lesssim (2\pi\sigma_{t_N}^2)^{-d/2}\sum_{j=1}^V \widehat{w}_N(e_j)\exp\bigl(-\|\widehat{x}_{t_N} - \alpha_{t_N}e_j\|_2^2/(2\sigma_{t_N}^2)\bigr). 
\end{align*}
As $t_N\to 1$, the variance of every Gaussian component vanishes and its center approaches a vocabulary embedding. Therefore, $q_{N}$ is asymptotically bounded by $\sum_{j=1}^V \widehat{w}_N(e_j) \delta_{e_j}$, where $\delta_{e_j}$ denotes the Dirac measure with mass at $e_j$.
This completes the proof.

\subsection{Proof of Proposition~\ref{lem:posterior_token_distribution_bayes}}
\label{sec:proof_posterior_token_distribution_bayes}

Define
\begin{align}
p^{(i)}(j\mid x_t^{(-i)}, t) \defn \bP\{s^{(i)} = j \mid x_t^{(-i)}\}, \quad j\in[V] \label{eq:context_only_posterior}.
\end{align}

By Bayes's rule, we can derive 
\begin{align}\label{eq:posterior_token_distribution_bayes}
\bP\{s^{(i)} = j \mid x_t\}
& = \frac{\bP\{s^{(i)} = j, x_t^{(i)} \mid x_t^{(-i)}\}}{\bP\{x_t^{(i)} \mid x_t^{(-i)}\}} 
= \frac{p^{(-i)}(j\mid x_t^{(-i)}, t)p(x_t^{(i)} \mid s^{(i)} = j, x_t^{(-i)})}{\sum_{j'\in[V]}p^{(-i)}(j'\mid x_t^{(-i)}, t) p(x_t^{(i)} \mid s^{(i)} = j', x_t^{(-i)})}. 
\end{align}
Because the Gaussian corruption is independent across token positions (see the probability path construction in \eqref{eq:flow_matching_path}), $x_t^{(i)}$ and $x_t^{(-i)}$ are conditionally independent given $s^{(i)}$. As a result, we have
\begin{align*}
x_t^{(i)} \mid s^{(i)} = j, x_t^{(-i)} \sim \cN(\alpha_t e_j, \sigma_t^2 I_d) .
\end{align*}
Plugging this into \eqref{eq:posterior_token_distribution_bayes} yields the claim in \eqref{eq:posterior_token_distribution_bayes2}.

\subsection{Proof of Proposition~\ref{prop:counterexample}}\label{sec:proof_counterexample}

Consider a general schedule $(\alpha_t, \sigma_t)$ such that $\alpha_t \to 1$ and $\sigma_t \to 0$ as $t \to 1$.
Define the smooth, unconstrained data predictor \begin{align}
\mu_{\theta}(x, t) = \frac{x}{\alpha_t+\sigma_t}.
\end{align}
It is easy to verify that
\begin{align*}
\mu_{\theta}(x_t, t) = \frac{x_t}{\alpha_t+\sigma_t} = \frac{\alpha_tx_\star + \sigma_t z}{\alpha_t+\sigma_t} \to x_\star \quad \text{as} \quad t \to 1.
\end{align*}
Thus, the data predictor is asymptotically accurate. However, we will show that the flow induced by this data predictor does not converge to any token embedding with constant probability.

By the ODE in \eqref{eq:flow_matching_ode} and the velocity identity in \eqref{eq:velocity_field_x}, the flow induced by $\mu_{\theta}$ is given by
\begin{align*}
\frac{\diff x_t}{\diff t} = \frac{\sigma_t'}{\sigma_t}x_t + \Big(\alpha_t'-\frac{\sigma_t'}{\sigma_t} \alpha_t \Big) \frac{x_t}{\alpha_t+\sigma_t} = \frac{\alpha_t'+\sigma_t'}{\alpha_t+\sigma_t}x_t.
\end{align*}
Solving the ODE shows that the flow trajectory initialized at $x_{t_0}$ evolves according to
\begin{align*}
x_t = \frac{\alpha_t+\sigma_t}{\alpha_{t_0}+\sigma_{t_0}}x_{t_0}.
\end{align*}
As a result, the data predictor remains constant along the flow trajectory:
\begin{align*}
\mu_{\theta}(x_t, t) = \frac{x_t}{\alpha_t+\sigma_t} = \frac{x_{t_0}}{\alpha_{t_0}+\sigma_{t_0}}.
\end{align*}

Suppose that $x_{t_0}=\sigma z$ with $z_{ij} \overset{\mathsf{i.i.d.}}{\sim} \mathcal{N}(0, 1)$. Then the final iterate is given by
\begin{align*}
\frac{x_{t_N}}{\alpha_{t_N}} = \sigma_N z \qquad \text{with} \qquad \sigma_N \defn \frac{\sigma(\alpha_{t_N}+\sigma_{t_N})}{\alpha_{t_N}(\alpha_{t_0}+\sigma_{t_0})}.
\end{align*}
Because $\sigma_t/\alpha_t \to 0$ as $t \to 1$, we have
\begin{align*}
    \sigma_N \downarrow \sigma_\infty \defn \frac{\sigma}{\alpha_{t_0}+\sigma_{t_0}} > 0 \quad \text{as} \quad N \to \infty.
\end{align*}
Because $\sigma_\infty z$ has a continuous distribution, its probability of belonging to any finite set is zero. Therefore, the final iterate $x_{t_N}$ fails to converge to any token embedding with constant probability, as formalized in \eqref{eq:counterexample}.

For example, if $\|e_j\|_2 = \sqrt{d}$ for every $j\in[V]$, then for any token position $i\in[L]$, one has
\begin{align*}
\bP\biggl\{\min_j \big\|\alpha_{t_N}^{-1}x_{t_N}^{(i)} - e_j\big\|_2 \geq 1 \biggr\} & \ge \bP\Bigl\{ \bigl|\sigma_N \|z^{(i)}\|_2 - \|e_j\|_2 \bigr| \ge 1 \Bigr\} \\ 
& \geq \bP\Bigl\{ \sigma_\infty \|z^{(i)}\|_2 \geq 1+\sqrt d \Bigr\}  \geq c_{\mathsf{lb}}
\end{align*}
for some constant $c_{\mathsf{lb}} > 0$ independent of $N$.

\subsection{Proof of Proposition \ref{prop:dis-weight-converge}}
\label{app:proof-prop-2}
% Moreover, as $\mu(x_t, t) = \sum_{j=1}^V w(e_j\mymid x_t, t)e_j$ with $w(e_j\mymid x_t, t) \ge 0$ and $\sum_{j=1}^V w(e_j\mymid x_t, t)$, we have the following result, which allow us to apply $\arg\max_{e_j} w(e_j\mymid x_t, t)$, even though we didn't use CE loss to learn $w(e_j\mymid x_t, t)$. 
% \begin{lemma}\label{prop:dis-weight-converge}
% As $\mu(x_t, t) \to e_j$, one has $w(\cdot\mymid x_t, t) \to \mathds{1}(e_j)$ when $e_j$'s are distinguishable and have the same norm.
% \end{lemma}

Let us denote by $\overline{e}_j \coloneqq e_j/\|e_j\|_2$ the normalized embedding vector.
Fix an arbitrary token position $i\in[L]$.
Notice the relation
\begin{align*}
\frac{\|\mu_{\theta}^{(i)}(x_t, t)\|_2^2}{\|e_j\|_2^2} &= \bigg\|\sum_{j^{\prime}=1}^V w_{\theta}^{(i)}(j^{\prime}\mymid x_t, t)\overline{e}_{j^{\prime}}\bigg\|_2^2 \\
&= \sum_{j^{\prime}=1}^V w_{\theta}^{(i)}(j^{\prime}\mymid x_t, t)^2 + \sum_{j^{\prime}=1}^V\sum_{\ell\neq j^{\prime}} w_{\theta}^{(i)}(j^{\prime}\mymid x_t, t)w_{\theta}^{(i)}(\ell\mymid x_t, t)\frac{\langle e_{j^{\prime}}, e_{\ell} \rangle}{\|e_{j^{\prime}}\|_2\|e_{\ell}\|_2}.
\end{align*}
By the separation condition that $\max_{j^{\prime}\neq \ell}\big|\frac{\langle e_{j^{\prime}}, e_{\ell} \rangle}{\|e_{j^{\prime}}\|_2\|e_{\ell}\|_2}\big| \le 1-\rho$, one can derive
\begin{align}\label{eq:proof-prop2-1}
\frac{\|\mu_{\theta}^{(i)}(x_t, t)\|_2^2}{\|e_j\|_2^2} &\le \sum_{j^{\prime}=1}^V w_{\theta}^{(i)}(j^{\prime}\mymid x_t, t)^2 + (1-\rho)\sum_{j^{\prime}=1}^V\sum_{\ell\neq j^{\prime}} w_{\theta}^{(i)}(j^{\prime}\mymid x_t, t)w_{\theta}^{(i)}(\ell\mymid x_t, t) \notag \\
&\overset{\text{(i)}}{=} \sum_{j^{\prime}=1}^V w_{\theta}^{(i)}(j^{\prime}\mymid x_t, t)^2 + (1-\rho)\sum_{j^{\prime}=1}^Vw_{\theta}^{(i)}(j^{\prime}\mymid x_t, t)\big(1-w_{\theta}^{(i)}(j^{\prime}\mymid x_t, t)\big) \notag \\
&\overset{\text{(ii)}}{=} \sum_{j^{\prime}=1}^V w_{\theta}^{(i)}(j^{\prime}\mymid x_t, t)^2 + (1-\rho) - (1-\rho)\sum_{j^{\prime}=1}^Vw_{\theta}^{(i)}(j^{\prime}\mymid x_t, t)^2 \notag \\
&= 1 - \rho + \rho\sum_{j^{\prime}=1}^V w_{\theta}^{(i)}(j^{\prime}\mymid x_t, t)^2,
\end{align}
where (i) and (ii) use the fact that $\sum_{j^{\prime}=1}^V w_{\theta}^{(i)}(j^{\prime}\mymid x_t, t)=1$.
As $\mu_{\theta}^{(i)}(x_t, t) \to e_j$, one has $\frac{\|\mu_{\theta}^{(i)}(x_t, t)\|_2^2}{\|e_j\|_2^2} \to 1$.
Combining this with \eqref{eq:proof-prop2-1}, we obtain
\begin{align*}
1 - \rho + \rho\lim_{t\to 1}\sum_{j^{\prime}=1}^V w_{\theta}^{(i)}(j^{\prime}\mymid x_t, t)^2 \ge 1,
\end{align*}
which leads to
$$
\lim_{t\to 1} \sum_{j^{\prime}=1}^V w_{\theta}^{(i)}(j^{\prime}\mymid x_t, t)^2\ge 1.
$$
As $w_{\theta}^{(i)}(j^{\prime}\mymid x_t, t)\in[0,1]$ and $\sum_{j^{\prime}=1}^V w_{\theta}^{(i)}(j^{\prime}\mymid x_t, t)=1$, there exists some $\widehat{j}\in[V]$ such as $w_{\theta}^{(i)}(\widehat{j}\mymid x_t, t)\to 1$ and $w_{\theta}^{(i)}(j^{\prime}\mymid x_t, t)\to 0$ for $j^{\prime}\neq \widehat{j}$.
Since $\mu_{\theta}^{(i)}(x_t, t) \to e_j$, we conclude that $\widehat{j} = j$, thereby completing the proof.

\section{Further experimental results}\label{sec:appendix_exp}
\begin{figure}[!htbp]
    \centering
    \includegraphics[width=0.52\textwidth]{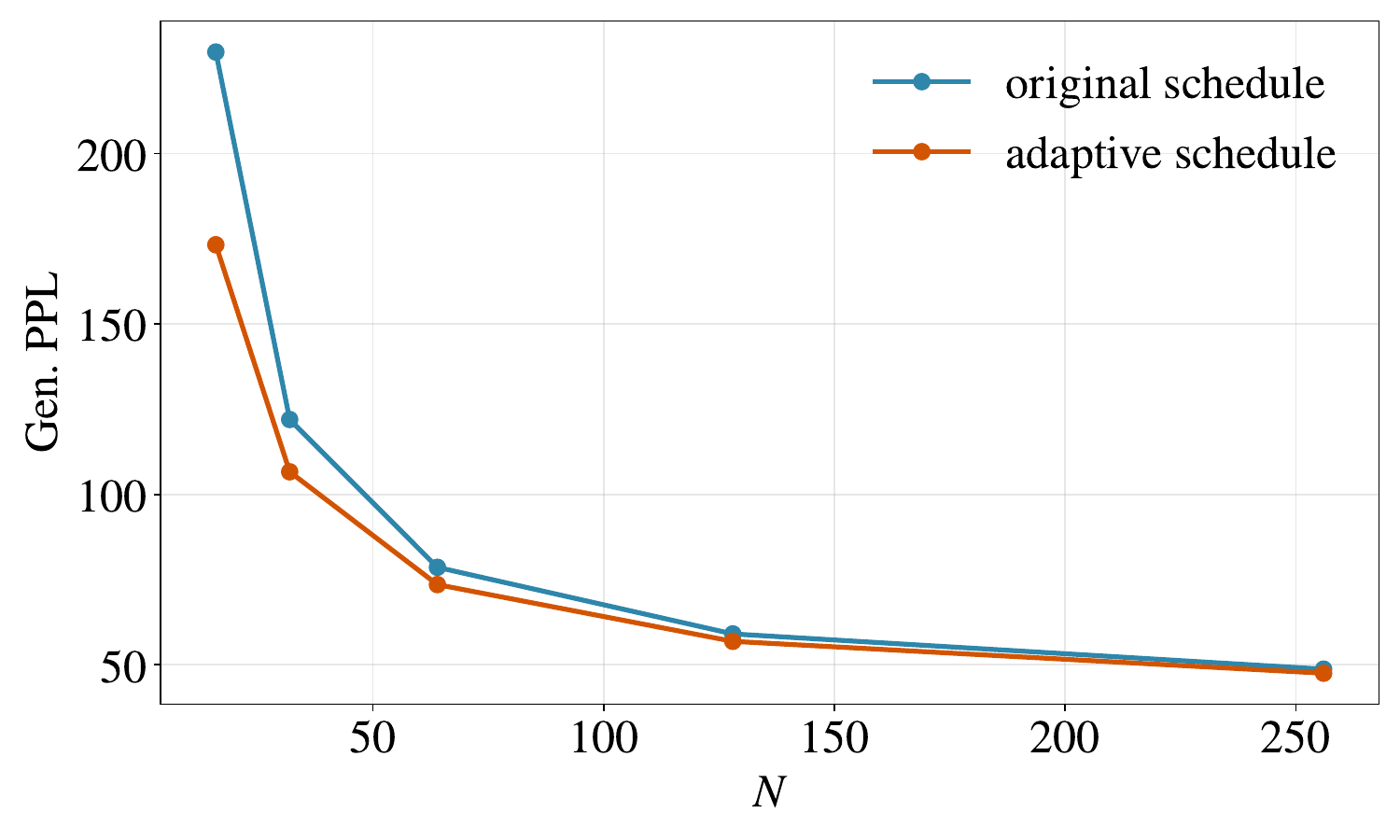}
    \caption{Effect of the sampling schedule on Gen. PPL. We compare the original schedule by LangFlow with the proposed adaptive schedule $t_i=(i+0.5)/N$, $i=0,\ldots,N-1$, across different steps $N$.}
    \label{fig:schedule}
\end{figure}

This section provides the complete numerical results underlying the experimental analyses in Section~\ref{sec:experiment}, together with additional results under larger NFE budgets.
Figure~\ref{fig:schedule} compares the original LangFlow schedule with the adaptive schedule.

% \subsection{Further results on the influence of sampler schedule}\label{subsec:schedule}

\subsection{Detailed results for quality-diversity control}\label{subsec:trade-off_appendix}
Tables~\ref{tab:trade-off_64} and \ref{tab:trade-off_128} report the complete numerical results for the three sampling techniques under NFE budgets of $64$ and $128$, respectively. These results underlie the NFE$=64$ trade-off curves in Figure~\ref{fig:trade-off} and extend the same comparison to NFE$=128$. Across both budgets, increasing the corresponding control parameter lowers Gen.~PPL at the cost of entropy, while the time-adaptive variants generally retain more entropy at comparable Gen.~PPL. This consistent behavior confirms the benefit of concentrating guidance or iterative refinement near the data endpoint.

\begin{table*}[!htbp]
\centering
\caption{
    % Gen.~PPL-entropy trade-offs obtained by varying self-conditioning guidance, iterative self-conditioning refinement, and unconditional guidance at NFE=$64$.
    Gen.~PPL-entropy trade-offs produced by the three sampling techniques under a fixed NFE budget of $64$. We compare each standard variant with its time-adaptive counterpart. For the time-adaptive variants, ``Nominal value'' denotes the control parameter before time-dependent scaling.
    }
\label{tab:trade-off_64}
\small
\setlength{\tabcolsep}{7pt}
\renewcommand{\arraystretch}{1.08}

\begin{tabular}{ccccccc}
\toprule
\multirow{2}{*}{Sampling technique} &\multicolumn{3}{c}{\textbf{Standard}}
&
\multicolumn{3}{c}{\textbf{Time-adaptive}}
\\
\cmidrule(l){2-4}
\cmidrule(l){5-7}
 & Value & Gen.~PPL & Entropy
 & Nominal value & Gen.~PPL & Entropy \\
\midrule

\multirow{4}{*}{\shortstack[c]{Iterative self-conditioning\\refinement}}
& 1 & 105.0006 & 5.5373
& 10 & 98.3887 & 5.5373
\\
& 2 & 85.4298 & 5.5139
& 15 & 90.3650 & 5.5303
\\
& 3 & 77.6263 & 5.5053
& 20 & 85.2866 & 5.5228
\\
& 5 & 73.3118 & 5.5027
& 40 & 78.6051 & 5.5176
\\

\midrule

\multirow{5}{*}{Self-conditioning guidance}
& 1 & 102.9869 & 5.5387
& 20  & 93.5983 & 5.5349
\\
& 2 & 80.8420 & 5.4981
& 40  & 73.6271 & 5.4962
\\
& 4 & 58.4980 & 5.4310
& 60  & 62.2094 & 5.4630
\\
& 6 & 47.3374 & 5.3790
& 80  & 54.6371 & 5.4349
\\
& 8 & 40.5905 & 5.3357
& 100 & 49.5442 & 5.4133
\\

\midrule

\multirow{5}{*}{Unconditional guidance}
& 0.01 & 91.9529 & 5.5084
& 0.25 & 91.6968 & 5.5106
\\
& 0.02 & 82.2282 & 5.4774
& 0.50 & 82.2702 & 5.4838
\\
& 0.04 & 66.9820 & 5.4171
& 1.00 & 67.0967 & 5.4312
\\
& 0.06 & 54.8009 & 5.3565
& 1.50 & 55.5643 & 5.3817
\\
& 0.08 & 45.8002 & 5.2953
& 2.00 & 47.0382 & 5.3343
\\

\bottomrule
\end{tabular}
\end{table*}

\begin{table*}[!htbp]
\centering
\caption{Gen.~PPL-entropy trade-offs produced by the three sampling techniques under a fixed NFE budget of $128$. We compare each standard variant with its time-adaptive counterpart. For the time-adaptive variants, ``Nominal value'' denotes the control parameter before time-dependent scaling.}
\label{tab:trade-off_128}
\small
\setlength{\tabcolsep}{7pt}
\renewcommand{\arraystretch}{1.08}

\begin{tabular}{ccc cccc}
\toprule
\multirow{2}{*}{Sampling technique} &\multicolumn{3}{c}{\textbf{Standard}}
&
\multicolumn{3}{c}{\textbf{Time-adaptive}}
\\
\cmidrule(l){2-4}
\cmidrule(l){5-7}
 & Value & Gen.~PPL & Entropy
 & Nominal value & Gen.~PPL & Entropy \\
\midrule

% ==================== k ====================

\multirow{4}{*}{\shortstack[c]{Iterative self-conditioning\\refinement}}
& 1 & 93.3518 & 5.5078
& 10 & 89.2618 & 5.5052
\\
& 2 & 73.9209 & 5.4694
& 15 & 80.2898 & 5.4941
\\
& 3 & 64.3410 & 5.4497
& 20 & 74.6253 & 5.4828
\\
& 5 & 56.5077 & 5.4300
& 40 & 64.9262 & 5.4636
\\
\midrule

% ==================== w_cfg ====================
\multirow{4}{*}{Self-conditioning guidance}
& 1 & 93.3518 & 5.5078
& 20 & 84.4239 & 5.5039
\\
& 2 & 72.7807 & 5.4623
& 40 & 65.6629 & 5.4601
\\
& 4 & 51.5329 & 5.3858
& 60 & 55.0511 & 5.4230
\\
& 6 & 41.0542 & 5.3279
& 80 & 48.2367 & 5.3934
\\
& 8 & 35.0187 & 5.2822
& 100 & 43.8744 & 5.3719
\\

\midrule

% ==================== w_ug ====================
\multirow{4}{*}{Unconditional guidance}
& 0.01 & 81.4945 & 5.4708
& 0.25 & 82.5130 & 5.4779
\\
& 0.02 & 72.4452 & 5.4362
& 0.50 & 73.6181 & 5.4501
\\
& 0.04 & 57.1142 & 5.3643
& 1.00 & 59.6448 & 5.3943
\\
& 0.06 & 46.1258 & 5.2910
& 1.50 & 49.2387 & 5.3419
\\
& 0.08 & 38.1393 & 5.2205
& 2.00 & 41.3981 & 5.2915
\\

\bottomrule
\end{tabular}
\end{table*}

\subsection{Detailed results for combinations of sampling techniques}\label{subsec:comb_appendix}
Tables~\ref{tab:wk_wcfg_64} and \ref{tab:wk_wcfg_128} provide the complete sweeps for jointly varying iterative self-conditioning refinement $K_{\iscr}$ and self-conditioning guidance $w_{\scg}$ under NFE budgets of $64$ and $128$, respectively. Consistent with Figure~\ref{fig:wk_wcfg}, a sufficiently large $K_{\iscr}$ combined with an appropriate $w_{\scg}$ shifts the trade-off frontier toward lower Gen.~PPL at comparable entropy. Table~\ref{tab:wug_wk_wcfg_nfe64} further reports the three-way combinations that include unconditional guidance. Varying $w_{\ug}$ mainly moves the operating point along a similar frontier, providing only a marginal additional improvement after the other two techniques have been combined.

\begin{table}[!htbp]
\centering
\caption{Joint effect of iterative self-conditioning refinement and self-conditioning guidance under $\mathrm{NFE}=64$.}
\label{tab:wk_wcfg_64}
\small
\setlength{\tabcolsep}{5.5pt}
\renewcommand{\arraystretch}{1.08}

\begin{tabular}{cccccccc}
\toprule
& & \multicolumn{6}{c}{Self-conditioning guidance strength $w_{\scg}$} \\
\cmidrule(l){3-8}
Refinement parameter
 $K_{\iscr}$ & Metric
& 10 & 15 & 20 & 25 & 30 & 40 \\
\midrule

\multirow{2}{*}{25}
& Gen.~PPL
& 96.5033 & 76.8682 & 64.2796 & 55.8742 & 51.1444 & 45.0416  \\
& Entropy
& 5.5532 & 5.5299 & 5.5109 & 5.4922 & 5.4794 & 5.4549 \\

\midrule
\multirow{2}{*}{50}
& Gen.~PPL
& 97.3557 & 69.4861 & 53.9569 & 45.1607 & 40.1334 & 35.691 \\
& Entropy
& 5.5654 & 5.5341 & 5.505 & 5.4796 & 5.4592 & 5.4306 \\

\midrule
\multirow{2}{*}{100}
& Gen.~PPL
& 118.1948 & 74.3155 & 52.0762 & 41.6154 & 36.5026 & 32.5312 \\
& Entropy
& 5.5909 & 5.5579 & 5.5237 & 5.4914 & 5.4634 & 5.4274 \\

\bottomrule
\end{tabular}
\end{table}

\begin{table}[!htbp]
\centering
\caption{Joint effect of iterative self-conditioning refinement and self-conditioning guidance under $\mathrm{NFE}=128$.}
\label{tab:wk_wcfg_128}
\small
\setlength{\tabcolsep}{5.5pt}
\renewcommand{\arraystretch}{1.08}

\begin{tabular}{cccccccc}
\toprule
& & \multicolumn{6}{c}{Self-conditioning guidance strength $w_{\scg}$} \\
\cmidrule(l){3-8}
Refinement parameter
 $K_{\iscr}$ & Metric
& 10 & 15 & 20 & 25 & 30 & 40 \\

\midrule
\multirow{2}{*}{50}
& Gen. PPL
& 79.8379 & 57.2238 & 44.7869 & 37.8605 & 33.7153 & 29.8325 \\
& Entropy
& 5.5191 & 5.4771 & 5.443 & 5.4145 & 5.3898 & 5.3587 \\

\midrule
\multirow{2}{*}{100}
& Gen. PPL
& 85.1575 & 54.8722 & 39.9296 & 32.6275 & 28.7678 & 25.5304 \\
& Entropy
& 5.5417 & 5.4906 & 5.4453 & 5.4049 & 5.3741 & 5.3379 \\

\midrule
\multirow{2}{*}{200}
& Gen. PPL
& 103.3736 & 59.3922 & 40.3115 & 32.0478 & 27.9034 & 24.925 \\
& Entropy
& 5.5743 & 5.5226 & 5.4712 & 5.4286 & 5.3937 & 5.3448 \\

\bottomrule
\end{tabular}
\end{table}

\begin{table*}[!htbp]
\centering
\caption{Joint effect of unconditional guidance, iterative self-conditioning refinement, and self-conditioning guidance under NFE$=64$.}
\label{tab:wug_wk_wcfg_nfe64}
\small
\setlength{\tabcolsep}{5.5pt}
\renewcommand{\arraystretch}{1.08}

\begin{tabular}{ccccccccc}
\toprule
& & & \multicolumn{6}{c}{Self-conditioning guidance strength $w_{\scg}$} \\
\cmidrule(l){4-9}
$w_{\ug}$ & 
 $K_{\iscr}$ & Metric
& 10 & 15 & 20 & 25 & 30 & 40 \\
\midrule

\multirow{4}{*}{0.0}
& \multirow{2}{*}{50}
& Gen. PPL
& 97.3557 & 69.4861 & 53.9569 & 45.1607 & 40.1334 & 35.6910 \\
& & Entropy
& 5.5654 & 5.5341 & 5.5050 & 5.4796 & 5.4592 & 5.4306 \\

\cmidrule(l){2-9}
& \multirow{2}{*}{100}
& Gen. PPL
& 118.1948 & 74.3155 & 52.0762 & 41.6154 & 36.5026 & 32.5312 \\
& & Entropy
& 5.5909 & 5.5579 & 5.5237 & 5.4914 & 5.4634 & 5.4274 \\

\midrule

\multirow{4}{*}{0.25}
& \multirow{2}{*}{50}
& Gen. PPL
& 88.4084 & 64.1003 & 50.0962 & 42.6205 & 37.9129 & 33.7868 \\
& & Entropy
& 5.5428 & 5.5135 & 5.4867 & 5.4638 & 5.4432 & 5.4155 \\

\cmidrule(l){2-9}
& \multirow{2}{*}{100}
& Gen. PPL
& 108.3142 & 69.4401 & 49.1897 & 40.0159 & 34.9445 & 31.2325 \\
& & Entropy
& 5.5693 & 5.5411 & 5.5081 & 5.4796 & 5.4520 & 5.4161 \\

\midrule

\multirow{4}{*}{0.5}
& \multirow{2}{*}{50}
& Gen. PPL
& 80.6113 & 59.0871 & 46.8084 & 39.8520 & 35.8997 & 32.1812 \\
& & Entropy
& 5.5206 & 5.4935 & 5.4682 & 5.4451 & 5.4274 & 5.3999 \\

\cmidrule(l){2-9}
& \multirow{2}{*}{100}
& Gen. PPL
& 100.4747 & 64.9446 & 46.4663 & 37.8119 & 33.4694 & 30.0821 \\
& & Entropy
& 5.5507 & 5.5237 & 5.4933 & 5.4643 & 5.4401 & 5.4074 \\

\bottomrule
\end{tabular}
\end{table*}

\subsection{Detailed results for guidance allocation}\label{subsec:fur_imp_appendix}
Tables~\ref{tab:baseline_nfe} and \ref{tab:cfgb_kb_nfe} report the complete sweeps for Configurations~A and B, respectively, under both NFE budgets. Tables~\ref{tab:guidance_nfe64} and \ref{tab:guidance_nfe128} then collect representative operating points from all three configurations for NFE$=64$ and $128$. The numerical results support the trends in Figure~\ref{fig:k}: Configuration~A reaches the low-entropy, low-Gen.~PPL regime, Configuration~B provides strong intermediate operating points, and Configuration~C preserves the greatest entropy. In overlapping regions, a larger refinement parameter $K_{\iscr}$ generally yields a more favorable trade-off, confirming that both the refinement count and its temporal allocation are important.

\begin{table*}[!htbp]
\centering
\caption{Results for Configuration~A that uses a constant refinement count \(K\) and self-conditioning guidance strength \(w_{\scg}\).}
\label{tab:baseline_nfe}
\small
\setlength{\tabcolsep}{8pt}
\renewcommand{\arraystretch}{1.08}

\begin{tabular}{cccccccc}
\toprule
& & & \multicolumn{5}{c}{Self-conditioning guidance strength $w_{\scg}$} \\
\cmidrule(l){4-8}
NFE &  $K$ & Metric
& 1.5 & 2 & 2.5 & 3 & 3.5 \\
\midrule

% ==================== NFE = 64 ====================
\multirow{6}{*}{64}

& \multirow{2}{*}{4}
& Gen.~PPL
& 45.7076 & 32.4898 & 26.1870 & 22.6047 & 20.7017 \\
& & Entropy
& 5.4124 & 5.3412 & 5.2835 & 5.2272 & 5.1831 \\

\cmidrule(l){2-8}

& \multirow{2}{*}{6}
& Gen.~PPL
& 42.8333 & 29.9203 & 24.1211 & 20.8427 & 19.2514 \\
& & Entropy
& 5.4107 & 5.3332 & 5.2610 & 5.1853 & 5.1118 \\

\cmidrule(l){2-8}

& \multirow{2}{*}{8}
& Gen.~PPL
& 39.4945 & 26.9691 & 21.1209 & 18.1815 & 16.7697 \\
& & Entropy
& 5.3929 & 5.3014 & 5.1909 & 5.0784 & 4.9710 \\

\midrule

% ==================== NFE = 128 ====================
\multirow{6}{*}{128}

& \multirow{2}{*}{4}
& Gen.~PPL
& 36.0510 & 25.1186 & 19.9984 & 17.1249 & 15.3786 \\
& & Entropy
& 5.3282 & 5.2313 & 5.1521 & 5.0747 & 5.0131 \\

\cmidrule(l){2-8}

& \multirow{2}{*}{6}
& Gen.~PPL
& 31.0520 & 21.3595 & 17.2039 & 14.6103 & 13.3592 \\
& & Entropy
& 5.3032 & 5.1938 & 5.0999 & 4.9871 & 4.8959 \\

\cmidrule(l){2-8}

& \multirow{2}{*}{8}
& Gen.~PPL
& 28.6775 & 19.4933 & 15.2157 & 12.9933 & 11.5411 \\
& & Entropy
& 5.2882 & 5.1598 & 5.0118 & 4.8618 & 4.6964 \\

\bottomrule
\end{tabular}
\end{table*}

\begin{table*}[!htbp]
\centering
\caption{Results for Configuration~B, in which both the refinement count and self-conditioning guidance strength are scaled by \(1/(1+\sqrt{\sigma_{t_i}/\alpha_{t_i}})\).}
\label{tab:cfgb_kb_nfe}
\small
\setlength{\tabcolsep}{8pt}
\renewcommand{\arraystretch}{1.08}

\begin{tabular}{ccccccc}
\toprule
& & & \multicolumn{4}{c}{Self-conditioning guidance strength $w_{\scg}$} \\
\cmidrule(l){4-7}
NFE & $K_{\iscr}$ & Metric
& 6 & 8 & 10 & 15 \\
\midrule

% ==================== NFE = 64 ====================
\multirow{8}{*}{64}

& \multirow{2}{*}{10}
& Gen.~PPL
& 54.4313 & 43.7583 & 38.1201 & 32.4360 \\
& & Entropy
& 5.4732 & 5.4399 & 5.4151 & 5.3710 \\

\cmidrule(l){2-7}

& \multirow{2}{*}{15}
& Gen.~PPL
& 48.0552 & 37.5346 & 32.1775 & 27.5590 \\
& & Entropy
& 5.4593 & 5.4161 & 5.3818 & 5.3216 \\

\cmidrule(l){2-7}

& \multirow{2}{*}{20}
& Gen.~PPL
& 44.6766 & 34.0579 & 29.1159 & 25.1861 \\
& & Entropy
& 5.4545 & 5.4040 & 5.3662 & 5.2994 \\

\cmidrule(l){2-7}

& \multirow{2}{*}{25}
& Gen.~PPL
& 42.7540 & 32.1617 & 27.5400 & 24.4351 \\
& & Entropy
& 5.4520 & 5.3987 & 5.3569 & 5.2904 \\

\midrule

% ==================== NFE = 128 ====================
\multirow{8}{*}{128}

& \multirow{2}{*}{10}
& Gen.~PPL
& 45.4814 & 36.4717 & 31.7101 & 26.5305 \\
& & Entropy
& 5.4142 & 5.3748 & 5.3485 & 5.2930 \\

\cmidrule(l){2-7}

& \multirow{2}{*}{20}
& Gen.~PPL
& 35.2302 & 26.7910 & 22.5727 & 18.6685 \\
& & Entropy
& 5.3741 & 5.3149 & 5.2624 & 5.1597 \\

\cmidrule(l){2-7}

& \multirow{2}{*}{30}
& Gen.~PPL
& 31.6256 & 23.6132 & 20.0410 & 17.2126 \\
& & Entropy
& 5.3633 & 5.2928 & 5.2321 & 5.1147 \\

\cmidrule(l){2-7}

& \multirow{2}{*}{40}
& Gen.~PPL
& 30.7218 & 22.6761 & 19.3316 & 17.0313 \\
& & Entropy
& 5.3640 & 5.2884 & 5.2264 & 5.1022 \\

\bottomrule
\end{tabular}
\end{table*}

\begin{table*}[!htbp]
\centering
\caption{Detailed Gen.~PPL and entropy results for Configurations~A--C under a fixed NFE budget of $64$. Within each configuration, the columns vary the nominal self-conditioning guidance strength $w_{\scg}$.}
\label{tab:guidance_nfe64}
\footnotesize
\renewcommand{\arraystretch}{1.05}
\setlength{\tabcolsep}{3pt}

\begin{tabular}{
@{}
c
c
p{0.72\textwidth}
@{}
}
\toprule
\makebox[2.0cm][c]{Configuration}
&
\makebox[1.8cm][c]{Metric}
&
\multicolumn{1}{c}{$w_{\scg}$}
\\
\midrule

% =========================================================
% Baseline
% =========================================================

& &
\begin{tabular*}{\linewidth}{
@{\extracolsep{\fill}}
ccccccccc
@{}
}
\makebox[0.62cm][c]{1.5} &
\makebox[0.62cm][c]{1.75} &
\makebox[0.62cm][c]{2} &
\makebox[0.62cm][c]{2.25} &
\makebox[0.62cm][c]{2.5} &
\makebox[0.62cm][c]{2.75} &
\makebox[0.62cm][c]{3} &
\makebox[0.62cm][c]{3.25} &
\makebox[0.62cm][c]{3.5}
\\
\cmidrule(l){1-9}
\end{tabular*}
\\[-1pt]

\multirow{2}{*}{A}
&
Gen. PPL
&
\begin{tabular*}{\linewidth}{
@{\extracolsep{\fill}}
ccccccccc
@{}
}
\makebox[0.62cm][c]{39.49} &
\makebox[0.62cm][c]{31.67} &
\makebox[0.62cm][c]{26.97} &
\makebox[0.62cm][c]{23.30} &
\makebox[0.62cm][c]{21.12} &
\makebox[0.62cm][c]{19.58} &
\makebox[0.62cm][c]{18.18} &
\makebox[0.62cm][c]{17.45} &
\makebox[0.62cm][c]{16.77}
\end{tabular*}
\\

& Entropy &
\begin{tabular*}{\linewidth}{
@{\extracolsep{\fill}}
ccccccccc
@{}
}
\makebox[0.62cm][c]{5.393} &
\makebox[0.62cm][c]{5.348} &
\makebox[0.62cm][c]{5.301} &
\makebox[0.62cm][c]{5.243} &
\makebox[0.62cm][c]{5.191} &
\makebox[0.62cm][c]{5.144} &
\makebox[0.62cm][c]{5.078} &
\makebox[0.62cm][c]{5.020} &
\makebox[0.62cm][c]{4.971}
\end{tabular*}
\\

\midrule

% =========================================================
% cfgB+kB, wk = 25
% =========================================================

& &
\begin{tabular*}{\linewidth}{
@{\extracolsep{\fill}}
ccccccc
@{}
}
\makebox[0.62cm][c]{6} &
\makebox[0.62cm][c]{7} &
\makebox[0.62cm][c]{8} &
\makebox[0.62cm][c]{9} &
\makebox[0.62cm][c]{10} &
\makebox[0.62cm][c]{12} &
\makebox[0.62cm][c]{15}
\\
\cmidrule(l){1-7}
\end{tabular*}
\\[-1pt]

\multirow{2}{*}{B}
&
Gen. PPL
&
\begin{tabular*}{\linewidth}{
@{\extracolsep{\fill}}
ccccccc
@{}
}
\makebox[0.62cm][c]{42.75} &
\makebox[0.62cm][c]{36.23} &
\makebox[0.62cm][c]{32.16} &
\makebox[0.62cm][c]{29.36} &
\makebox[0.62cm][c]{27.54} &
\makebox[0.62cm][c]{25.37} &
\makebox[0.62cm][c]{24.44}
\end{tabular*}
\\

& Entropy &
\begin{tabular*}{\linewidth}{
@{\extracolsep{\fill}}
ccccccc
@{}
}
\makebox[0.62cm][c]{5.452} &
\makebox[0.62cm][c]{5.425} &
\makebox[0.62cm][c]{5.399} &
\makebox[0.62cm][c]{5.379} &
\makebox[0.62cm][c]{5.357} &
\makebox[0.62cm][c]{5.328} &
\makebox[0.62cm][c]{5.290}
\end{tabular*}
\\

\midrule

% =========================================================
% cfgB+kB, wk = 100
% =========================================================

& &
\begin{tabular*}{\linewidth}{
@{\extracolsep{\fill}}
ccccccccccc
@{}
}
\makebox[0.62cm][c]{20} &
\makebox[0.62cm][c]{22} &
\makebox[0.62cm][c]{24} &
\makebox[0.62cm][c]{26} &
\makebox[0.62cm][c]{28} &
\makebox[0.62cm][c]{30} &
\makebox[0.62cm][c]{32} &
\makebox[0.62cm][c]{34} &
\makebox[0.62cm][c]{36} &
\makebox[0.62cm][c]{38} &
\makebox[0.62cm][c]{40}
\\
\cmidrule(l){1-11}
\end{tabular*}
\\[-1pt]

\multirow{2}{*}{C}
&
Gen. PPL
&
\begin{tabular*}{\linewidth}{
@{\extracolsep{\fill}}
ccccccccccc
@{}
}
\makebox[0.62cm][c]{52.08} &
\makebox[0.62cm][c]{47.04} &
\makebox[0.62cm][c]{43.07} &
\makebox[0.62cm][c]{40.28} &
\makebox[0.62cm][c]{38.09} &
\makebox[0.62cm][c]{36.50} &
\makebox[0.62cm][c]{35.10} &
\makebox[0.62cm][c]{34.17} &
\makebox[0.62cm][c]{33.45} &
\makebox[0.62cm][c]{32.83} &
\makebox[0.62cm][c]{32.53}
\end{tabular*}
\\

& Entropy &
\begin{tabular*}{\linewidth}{
@{\extracolsep{\fill}}
ccccccccccc
@{}
}
\makebox[0.62cm][c]{5.524} &
\makebox[0.62cm][c]{5.511} &
\makebox[0.62cm][c]{5.495} &
\makebox[0.62cm][c]{5.484} &
\makebox[0.62cm][c]{5.473} &
\makebox[0.62cm][c]{5.463} &
\makebox[0.62cm][c]{5.456} &
\makebox[0.62cm][c]{5.448} &
\makebox[0.62cm][c]{5.441} &
\makebox[0.62cm][c]{5.434} &
\makebox[0.62cm][c]{5.427}
\end{tabular*}
\\

\bottomrule
\end{tabular}
\end{table*}

\begin{table*}[!htbp]
\centering
\caption{Detailed Gen.~PPL and entropy results for Configurations~A--C under a fixed NFE budget of $128$. Within each configuration, the columns vary the nominal self-conditioning guidance strength $w_{\scg}$.}
\label{tab:guidance_nfe128}
\footnotesize
\renewcommand{\arraystretch}{1.05}
\setlength{\tabcolsep}{3pt}

\begin{tabular}{
@{}
c
c
p{0.72\textwidth}
@{}
}
\toprule
\makebox[2.0cm][c]{Configuration}
&
\makebox[1.8cm][c]{Metric}
&
\multicolumn{1}{c}{$w_{\scg}$}
\\
\midrule

% =========================================================
% Baseline
% =========================================================

& &
\begin{tabular*}{\linewidth}{
@{\extracolsep{\fill}}
ccccccccc
@{}
}
\makebox[0.62cm][c]{1.5} &
\makebox[0.62cm][c]{1.75} &
\makebox[0.62cm][c]{2} &
\makebox[0.62cm][c]{2.25} &
\makebox[0.62cm][c]{2.5} &
\makebox[0.62cm][c]{2.75} &
\makebox[0.62cm][c]{3} &
\makebox[0.62cm][c]{3.25} &
\makebox[0.62cm][c]{3.5}
\\
\cmidrule{1-9}
\end{tabular*}
\\[-1pt]

\multirow{2}{*}{A}
&
Gen. PPL
&
\begin{tabular*}{\linewidth}{
@{\extracolsep{\fill}}
ccccccccc
@{}
}
\makebox[0.62cm][c]{28.68} &
\makebox[0.62cm][c]{23.10} &
\makebox[0.62cm][c]{19.49} &
\makebox[0.62cm][c]{16.99} &
\makebox[0.62cm][c]{15.22} &
\makebox[0.62cm][c]{14.02} &
\makebox[0.62cm][c]{12.99} &
\makebox[0.62cm][c]{12.23} &
\makebox[0.62cm][c]{11.54}
\end{tabular*}
\\

& Entropy &
\begin{tabular*}{\linewidth}{
@{\extracolsep{\fill}}
ccccccccc
@{}
}
\makebox[0.62cm][c]{5.288} &
\makebox[0.62cm][c]{5.222} &
\makebox[0.62cm][c]{5.160} &
\makebox[0.62cm][c]{5.092} &
\makebox[0.62cm][c]{5.012} &
\makebox[0.62cm][c]{4.948} &
\makebox[0.62cm][c]{4.862} &
\makebox[0.62cm][c]{4.789} &
\makebox[0.62cm][c]{4.696}
\end{tabular*}
\\

\midrule

% =========================================================
% cfgB+kB, wk = 40
% =========================================================

& &
\begin{tabular*}{\linewidth}{
@{\extracolsep{\fill}}
ccccccc
@{}
}
\makebox[0.62cm][c]{6} &
\makebox[0.62cm][c]{7} &
\makebox[0.62cm][c]{8} &
\makebox[0.62cm][c]{9} &
\makebox[0.62cm][c]{10} &
\makebox[0.62cm][c]{12} &
\makebox[0.62cm][c]{15}
\\
\cmidrule{1-7}
\end{tabular*}
\\[-1pt]

\multirow{2}{*}{B}
&
Gen. PPL
&
\begin{tabular*}{\linewidth}{
@{\extracolsep{\fill}}
ccccccc
@{}
}
\makebox[0.62cm][c]{30.72} &
\makebox[0.62cm][c]{25.81} &
\makebox[0.62cm][c]{22.68} &
\makebox[0.62cm][c]{20.87} &
\makebox[0.62cm][c]{19.33} &
\makebox[0.62cm][c]{17.64} &
\makebox[0.62cm][c]{17.03}
\end{tabular*}
\\

& Entropy &
\begin{tabular*}{\linewidth}{
@{\extracolsep{\fill}}
ccccccc
@{}
}
\makebox[0.62cm][c]{5.364} &
\makebox[0.62cm][c]{5.325} &
\makebox[0.62cm][c]{5.288} &
\makebox[0.62cm][c]{5.261} &
\makebox[0.62cm][c]{5.226} &
\makebox[0.62cm][c]{5.168} &
\makebox[0.62cm][c]{5.102}
\end{tabular*}
\\

\midrule

% =========================================================
% cfgB+kB, wk = 200
% =========================================================

& &
\begin{tabular*}{\linewidth}{
@{\extracolsep{\fill}}
ccccccccccc
@{}
}
\makebox[0.62cm][c]{20} &
\makebox[0.62cm][c]{22} &
\makebox[0.62cm][c]{24} &
\makebox[0.62cm][c]{26} &
\makebox[0.62cm][c]{28} &
\makebox[0.62cm][c]{30} &
\makebox[0.62cm][c]{32} &
\makebox[0.62cm][c]{34} &
\makebox[0.62cm][c]{36} &
\makebox[0.62cm][c]{38} &
\makebox[0.62cm][c]{40}
\\
\cmidrule{1-11}
\end{tabular*}
\\[-1pt]

\multirow{2}{*}{C}
&
Gen. PPL
&
\begin{tabular*}{\linewidth}{
@{\extracolsep{\fill}}
ccccccccccc
@{}
}
\makebox[0.62cm][c]{40.31} &
\makebox[0.62cm][c]{36.23} &
\makebox[0.62cm][c]{33.32} &
\makebox[0.62cm][c]{30.92} &
\makebox[0.62cm][c]{29.21} &
\makebox[0.62cm][c]{27.90} &
\makebox[0.62cm][c]{26.94} &
\makebox[0.62cm][c]{26.22} &
\makebox[0.62cm][c]{25.52} &
\makebox[0.62cm][c]{25.00} &
\makebox[0.62cm][c]{24.93}
\end{tabular*}
\\

& Entropy &
\begin{tabular*}{\linewidth}{
@{\extracolsep{\fill}}
ccccccccccc
@{}
}
\makebox[0.62cm][c]{5.471} &
\makebox[0.62cm][c]{5.454} &
\makebox[0.62cm][c]{5.438} &
\makebox[0.62cm][c]{5.422} &
\makebox[0.62cm][c]{5.405} &
\makebox[0.62cm][c]{5.394} &
\makebox[0.62cm][c]{5.382} &
\makebox[0.62cm][c]{5.370} &
\makebox[0.62cm][c]{5.361} &
\makebox[0.62cm][c]{5.353} &
\makebox[0.62cm][c]{5.345}
\end{tabular*}
\\

\bottomrule
\end{tabular}
\end{table*}

\subsection{Qualitative Samples}\label{subsec:samples}
We present qualitative generation samples from ConvergeFlow under NFE=$64$. All samplesare generated with a fixed sequence length of 1024 tokens.

\begin{tcolorbox}[
    samplebox,
    breakable,
    title={ConvergeFlow}
]
\textbf{NFE:} $64$; 
\textbf{Gen.~PPL:} $33.09$; 
\textbf{Entropy:} $5.44$

\medskip

So if capitalism is to be subordinate to the proletariat (from which it must be neither independent, independent, or direct friend), then the state must do its part and let it make no concessions to what it wants, or force it to keep its power. Indeed, as is often the case with major bourgeois parties, the bourgeoisie is more interested in the actual fate of its role than in other parties.<|endoftext|>Yes, I would doubt that this (which has never happened), because Disney was about making an existing non-contained series with one-row characters in the 1930s or early 1940s. Even by this point, I was quite confident that the studio had to submit a schedule (as several months later) of two sold-out Disney-only episodes (and that, today if you haven't made a Disney-only series, you've made it elsewhere). But it's also quite probable now that there are at least one three episodes. It's hard to know what this would amount, but at least right now, it would seem that it would be a stretch to think we could make a lot season (or, for whatever reason, I'd suggest starting at least sooner).

Secondly, what about the current phase of our strategy is the need to work on the groundbreaking books made up to children at Disney. I know Ursula Puig was best known for this idea, and I suppose it's safe to say (with his knowledge) that most of our ongoing work is currently based on works that don't seem familiar enough to tell a live-action story about a man and two orphaned children in a world spanning as 100 or so years. As far as I know (e.g., the comic books), no longer are just 17 children's books in the comic books (such as Alice for Wonderland, King's Child, Aladdin, the Lion Z, and Tooty Duck: Edge of Time) that have appeared over the past 200 years alone. (And there are 17 of these made up more than a hundred years ago.) That supplemental programming has begun to play.

In that respect, I've just offered some detail of one fairly important strategy I would have come waiting to worry about:

Rather than make 12 completely incompatible films with an iconic female lead, our challenge was to make a realistic show, featuring a female lead character. In this area, we could focus on hundreds, or thousands, of a male/female difference, and then we could do it multiple times each year. But we had to pay a \$10 fee (at the Disney amusement park, in New York) for every episode of our show that went through a deal with Turner Entertainment (a major partner for the ABC network, now owned by the BBC). Along the way, we went three more years on that goal and built the first Disney network to feature a female lead in every episode. Sailor Moon, Nickelodeon, Snow White, and the ABC's Pirates of the Sea have all been designed with Turner Brothers (with ABC obtaining a fee each) in less than two years.

Today, our vision is very broad. A major part of the mission that we've been working toward over the course of the 20 years now is: to make 12 completely films with iconic male characters, and have stand-alone characters with really female monsters and really female characters. And those people are all really talented! We don't really know how progressive we would actually be in taking that initiative, but it certainly makes sense. We wanted to be the first (and, certainly, most lucky, not necessarily the first) that we'd make "way 12" of a one-week movie with a female lead in the 1960s. The one other thing that really sucks, however, is that in the 1950s, we still had both the restriction (and expectations) of a really female character. We realized that the narratives conferred on women in the '60s and '70s were unfair, and didn't work. And then there just seemed to be no way to repurpose a family-friendly show that only had a stand-alone character (such as Professor Norris or Cat in the Jungle or Braisey or Winnie Florid reappearing in the 1930s), or even a stand-alone character such as Bugs Bunny, or even having to have any sort of female cast since the 1950s. (There's only a that one now as it stands by now.)

Once the proper process of gender-level work is completed (and then on more heavily done), then I think we can have the confidence that there are many other shows we can focus on being produced by women in the media, too. But we'll ask for commitment until we can see that we absolutely need to focus on female characters everywhere. And of course, any sexism within our departments.

Finally, let's start by asking what is the necessary plan in these 20 years to try to make a completely consistent entertainment with a cartoon female first? This is what Walt Disney Animation and other

\end{tcolorbox}

\begin{tcolorbox}[
    samplebox,
    breakable,
    title={ConvergeFlow}
]
\textbf{NFE:} $64$; 
\textbf{Gen.~PPL:} $33.17$; 
\textbf{Entropy:} $5.44$

\medskip
Thousands of women have taken steps to prevent and stop such substance abuse. This is the first federal legislation in the country, changing the constitutional rights of women. Activists and women of all administration should be able to have the courage to speak up on this issue,'' said Ayesem Housem, executive director for UNARAD in Washington, in a statement. ``This legislation makes rape and violence a permanent target for our justice system.''

Miranda Moore, president and co-founder of the League on Women, State and Girls, said in a release: ``This bill is about who cares about fighting sexual assault and what it may mean for women of color.''

The drug drug conventions was passed in 1998, many states which have implemented drug laws now follow their federal practice. The District of Columbia is the first state that has banned marijuana, and it is the first time federal laws has passed. President Donald Trump has begun moving with a number of other directives that are attempt to cut down the use of marijuana, reduce the state's opioid consumption and expand the criminal justice system.

During the campaign, President Donald Trump signed the Law Justice Act of 2017, which would decriminalize marijuana for use. The policy was announced in part by Attorney Attorney General Eric Holder, who partnered with federal prosecutors to end the crackdown on non-violent offenders last year. The policy is expected to take effect in July 2018.<|endoftext|>As seems perfectly reasonable, few small businesses make lots of money to run a business. Unfortunately, our customers still get ``great money'' from our websites and business projects. If we can put it all by ourselves, let's see how we can make it.

In this post I looked at how much money we used on Starbucks customers and how we missed it. In part 1, you'll also see how much we wasted money.

Of course, I plan to update this post today. The video will give you a glimpse of how we worked (i.e., how we failed in terms of great innovation and marketing). I will also explain some of the mistakes we made and break out a brief outline of how our financial failures happened.

There are two main things I've talked about --- improved customer experience and better customer success. In this post, I describe a more efficient strategy (i.e., we get financial resources from just one customer while building out only one business). Using these strategies, we can help accomplish a lot more smoothly.

Let's start with the PayPal problem. We had a long and painful battle with PayPal and PayPal Inc, costing us \$6 million. The truth is, we kept spending money by using PayPal and all the water in the air was blowing up our financial rate of nowhere. Let us look at some of our (fun) mistakes:

Spending

We spent about \$33 million building an business. As the name goes, the company had 22.1 percent of revenue. We borrowed money from other customers, built our first eCommerce store and rented out our unique business with PayPal. We also kept track of our business by spending money.

Spending was consistently inefficient --- i.e., 49 percent of revenue came directly from our website and PayPal. We made an increase of 24.6 percent. (While you don't have to spend all that much money, here's how to do that.)

Let's start with four methods:

Dation of debt. This means that money was actually charged for us. We charged \$4,500 a month --- the average cost of (bighbage). Our charges were big-ranging and the average interest rate was 79 percent, an increase of 17.5 percent.

Investment of cost. This method didn't change much at all. We engaged our e-commerce team to various teams in the development of our e-commerce store so that we could gain a steady flow of revenue while building out new businesses that serve customers. We spent 12 percent of our revenue in our first ``4-point'' trial. This increased the efficiency of our revenue model and gave us an increase of 1.8 percent. While our revenue was 1.6 or more optimistic than the 4-point trial, we still reduced productivity by 18.5 percent.

Our loans. The fourth method was quite inefficient. We didn't use debt at all. Even rather, we borrowed income from just one customer. (Note the loans from the screen apps via app.) You don't have to make that test to see how it's worked.

Borrow from one customer. When we got financial money from another customer, we earned 23.8 percent.

This last method is a little different. Unlike our first method, it doesn't really count as passive income or debt. Instead, we just take extra income from
\end{tcolorbox}

\begin{tcolorbox}[
    samplebox,
    breakable,
    title={ConvergeFlow}
]
\textbf{NFE:} $64$; 
\textbf{Gen.~PPL:} $33.17$; 
\textbf{Entropy:} $5.44$

\medskip
Primarily, it also applies to snakes, kangaroo, trophy-tailed dogs, and so on, while it applies to cows, dogs and chickens, raccours, leopards, wolves, and so on, and many other phenomena. Humanism allows for many different kinds of negative acts toward humans. Without these negative acts, nonhumanists have no reason to believe that their own actions are coming from humans, even if they share their true beliefs, values, and so forth. Eugenicists, in stark contrast, have no reason to take anyone's actions, even if they have society's moral authority behind them. It is the responsibilities of humanists to decide what is right for humans and what is best for society in these decisions.

Humanism is a statement of nature's nature, a critique of nature's culture, history, and so forth. It is essential for human people to consistently understand this one way or another, and obviously we will have incredible difficulty in hearing them. Vegetarianism does not view nature as a complicated activity that seeks to set the boundaries to our laws as we approach them toward one another. But its system of veganism is particularly harmful because it can intentionally distort nature's boundaries and it is demeaning. Bulled animal abuse is so ubiquitous as it has little official use among animals. The failure to apply it, instead of treating it as a weapon, gives a death eye to ethical weakness. This is why the literature showing that animal abuse in animals is widespread, well-documented, and widespread.

Animalism is especially dangerous because it is often used as an excuse to stir a criticism of nature's boundaries, but almost always by literally calling for animals that can be captured and domesticated. This incapacity to ignore it, instead of treating it as a deterrent, puts a temporary end to ethical weakness. Whether humans aim to ``strictly exploit the natural environment'' or ``endrictly eat livestock,'' nonhumanists often claim that their laws are incompatible with their moral duties against humans and animals we use, and even make the case that society ought to work with them just to make their lives better.

The key to improving the ethical treatment of victims of animal animal abuse is the lack of empathy by individuals who care nothing about their behavior as they understand their attitudes toward it. They cannot even talk about the boundaries or the acts that justify one's moral duty towards animals. The most precise example used in this article is when lions and tigers were allegedly shot down an amusement park in Zimbabwe.

Reason and Peace: Nonhumanists

Phumanists are not necessarily bad people. They often celebrate a state's constitutional right to a man to keep arms, but not a state's constitutional right to bear arms. For some, the loss of such a powerful fellow is an extremely useful ethical choice given that it encaps in just a few of the same core principles and principles that historically entangled them with the wide range of moral issues. Nonhumanists generally seek to accept the ``crime'' of aggression with the right to be taken away from others, even though they distinguish between humans and all humans, rather than by acting on the application of moral relativism. Even when using the words ``gender heteronamally,'' an abdication of a bogeyman or patriarchal society, or a woman's legal right to have children, or the ability to have children and them, these are thoroughly debunked.

Nonhumanists often seek to understand the ``naturalization'' of all humans, but how they seek to justify it in order to acknowledge the absolutism of nature. For example, they object to the term ``theetine Earth,'' and talk about keeping rabbits, gorillas or chimpanzees living in open spaces and engaging in other forms of transhumanization. Nonhumanists do not accept the purely contestal moral choice of a human being, unless they harm others with their own choices, even when pushing for creating laws or regulations that clearly violate our laws.

It is important to understand that physical violence---even intentional and reprehensible conduct---is central to our ethical practice. All acts of actions are religiously defined by these principles, and are adjudicated with a quick response to everyone who wants them. Nonhumanists areently act on animals without their direct legal responsibility, and rely exclusively on animals to try to destroy human beings lest they flaunt their rights.

Legal Considerations

You can find an authoritative list of dozens of cases that are central to our ethical practice. However, these are generally the laws that should be discussed in practice.

Human beings are well-organized in performing their moral duties towards us and should not be subject to any use by those with moral authority or responsibility. If you believe in truly equal liberty and equality, accept the fact that those with good moral responsibility are being only
\end{tcolorbox}

\begin{tcolorbox}[
    samplebox,
    breakable,
    title={ConvergeFlow}
]
\textbf{NFE:} $64$; 
\textbf{Gen.~PPL:} $33.17$; 
\textbf{Entropy:} $5.44$

\medskip
She looked at her very beautiful young sister, who looked up, but well-liked feeling as though she was making a living for her.

Kazuki came over to the restaurant without question, and walked into it with a sigh of relief. "There's a serious problem, mother, don't trust me if anyone asks that you can help."

"Thanks the best!" Murayumi said. Misuki closed his eyes and glanced at him, defiantly. "Now, try not to distract them, are you going to stay home, or are you planning next tomorrow to take care of your daughter altogether?"

Murayumi looked down at her feet to see if that would do anything, but that was what Misuki expected of her. She leaned on being worried about government interference, which seemed to be being the root cause of the situation. Misuki didn't know when to talk or reassure her. His partner gave her a nod up next, but the blond's two went alingter as she turned around, noticing that Murayumi felt like she had more control. Maybe she would, but she started to calm down, and then started to talk to bed.

"Looks better here. I'll talk to you next week as we should. Any questions, I suppose." Kazuki groaned and glapped slightly.

"That was not working out." Murayumi let out a whimper. "Anyway, I'll get you back here soon. I'm willing to do something else for you if you can help me do my very best! Dugging." Misuki glanced over at Murayumi's young sister, angry that she wasn't pustering. He tumbled to his feet and walked around, kneeling in search of food and swinging around on a jet driven midboard. Murayumi may have wanted him to, but he didn't know. The young Yang seemed even more concerned, not with her new clothes, but her new body and anything else that could hold on. Why did it be so hot?

"What do you feel concerned about?" Murayumi began, blinking clearly. "I need a couple more minutes if I can discuss it with you."

Shinuki's response was interesting. He was had much to think about, but it was perhaps not by the best part. She was calm and one of the most muscleless people that Murayumi had ever had to offer, and he was certainly glad she would have been able to carry on with him.

"That's it." Murayumi said as he walked around back to the table and waited for some leftover pancakes to prepare. "Get it yourself if you can."

"Mother." She was ready, lifting her head open and peering her hands back into the pancakes. The menu was okay as well, but the opportunity to see Misuki get on herself was relatively easy. The thinness of the legs didn't help by much at that time, especially for her. They were the tinnest parts of Murayumi's body and covered them perfectly, so she turned them around to make a skie. Though she didn't have the same level of difficulty as caused by the abundance of clothes she was wearing, she instead appeared to be extremely active and motivated to get dressed.

So, what happened next? The initial conversation was seemingly awkward at first, but it did offer an interesting perspective. Misuki already had a good understanding of Murayumi's condition, but she was still not in fact very concerned of her health. She was definitely very concerned about the weight that she had put on her body, so it was actually by any means worrying. Misuki turned his eyes back to her mother and wondered how things had held up. "Was that one kind of meat? Were you ever worried about it?" Murayumi asked. "I wasn't so much concerned, either. Had it not gone far enough?"

"It has been a very busy week, I suppose." Murayumi said, moving her mother's hands back to the table and sliding her face up at the table. "It's been fun, though, mother. Like I said before I ate most of your meals, but when you woke up, I always took them away right away."

"Alright." Misuki said, turning back to the table and hunching her hands, while still trying to force himself away from herself. "Alright, mother. I'll talk with you as I plan a meal within the week."

Asakura shook her head and shook her head. "Again, great." She said. "Thank you so much. I hope you can join me for some wonderful times. You're the one that I hope to eventually meet."

"Haha." Murayumi said with a hint of relief. "Well, you're the one I would use the most. Later on, I can't serve you. Don't let
\end{tcolorbox}

\end{document}